\documentclass[twoside,11pt]{article}
\usepackage{jmlr2e}
\usepackage{amsmath,amssymb}
\usepackage{graphicx}
\usepackage{color}
\usepackage{mathrsfs}
\usepackage{bm}
\usepackage{multirow}
\usepackage{booktabs}
\usepackage{url}
\usepackage[left]{lineno}
\newcommand{\Part}[3]{ \frac{ \partial^{#3} #1 }{ \partial #2^{#3} } }
\newcommand{\V}[1]{\bm{#1} } 

\newcommand{\Ave}[1]{\left\langle {#1} \right\rangle} 
 
\newcommand{\sgn}[1]{{\rm sgn}\left({#1} \right)}

\newcommand{\mR}{\mathbb{R}}
\newcommand{\mN}{\mathbb{N}}
\newcommand{\lb}{\left(}
\newcommand{\rb}{\right)}
\newcommand{\lbb}{\left\{}
\newcommand{\rbb}{\right\}}
\newcommand{\lsb}{ \left[ }
\newcommand{\rsb}{ \right] }

\newcommand{\Req}[1]{(\ref{eq:#1})}
\newcommand{\BReq}[1]{(\ref{eq:#1})}
\newcommand{\NReq}[1]{(\ref{eq:#1})}
\newcommand{\Reqs}[2]{(\ref{eq:#1}),(\ref{eq:#2})}
\newcommand{\Reqss}[2]{(\ref{eq:#1})--(\ref{eq:#2})}
\newcommand{\Rfig}[1]{Figure \ref{fig:#1}}
\newcommand{\Rfigs}[2]{Figures \ref{fig:#1} and \ref{fig:#2}}
\newcommand{\Lfig}[1]{\label{fig:#1}}
\newcommand{\Leq}[1]{\label{eq:#1}}
\newcommand{\Rsec}[1]{Section\ \ref{sec:#1}}
\newcommand{\Rsecs}[2]{Sections\ \ref{sec:#1} and \ref{sec:#2}}
\newcommand{\Lsec}[1]{\label{sec:#1}}
\newcommand{\be}{\begin{eqnarray}}
\newcommand{\ee}{\end{eqnarray}}
\newcommand{\nbe}{\begin{eqnarray*}}
\newcommand{\nee}{\end{eqnarray*}}
\newcommand{\ba}{\begin{array}}
\newcommand{\ea}{\end{array}}
\newcommand{\no}{\nonumber}

\newcommand{\subbe}{\begin{subequations}}
\newcommand{\subee}{\end{subequations}}
\newcommand{\bs}{\backslash}
\newcommand{\mc}[1]{\mathcal{#1}}
\DeclareMathOperator*{\argmin}{arg\,min}

\newcommand{\MSE}{{\rm MSE}}

\newcommand{\BPC}[1]{\widetilde{#1}}

\newcommand{\Cov}[3]{{\rm Cov}^{#3}\left(#1,#2\right)}

\begin{document}
\title{Semi-Analytic Resampling in Lasso}
\author{	\name Tomoyuki Obuchi \email obuchi@c.titech.ac.jp \\
		\AND
		\name Yoshiyuki Kabashima \email kaba@c.titech.ac.jp \\
       		\addr Department of Mathematical and Computing Science\\
      			Tokyo Institute of Technology\\
       	2-12-1, Ookayama, Meguro-ku, Tokyo, Japan
       }
\editor{***}
\maketitle

\begin{abstract}
An approximate method for conducting resampling in Lasso, the $\ell_1$ penalized linear regression, in a semi-analytic manner is developed, whereby the average over the resampled datasets is directly computed without repeated numerical sampling, thus enabling an inference free of the statistical fluctuations due to sampling finiteness, as well as a significant reduction of computational time. The proposed method is based on a message passing type algorithm, and its fast convergence is guaranteed by the state evolution analysis, when covariates are provided as zero-mean independently and identically distributed Gaussian random variables. It is employed to implement bootstrapped Lasso (Bolasso) and stability selection, both of which are variable selection methods using resampling in conjunction with Lasso, and resolves their disadvantage regarding computational cost. To examine approximation accuracy and efficiency, numerical experiments were carried out using simulated datasets. Moreover, an application to a real-world dataset, the wine quality dataset, is presented. To process such real-world datasets, an objective criterion for determining the relevance of selected variables is also introduced by the addition of noise variables and resampling.
MATLAB codes implementing the proposed method are distributed in~\citep{obuchi:AMPR}.
\end{abstract}

\begin{keywords}
bootstrap method, Lasso, variable selection, message passing algorithm, replica method
\end{keywords}

\section{Introduction} \Lsec{Introduction}
Variable selection is an important problem in statistics, signal processing, and machine learning. A desire for useful techniques of variable selection has recently been growing, as cumulated advances in measurement and information technologies have started to steadily produce a large amount of high-dimensional data in science and engineering. A naive method for selecting relevant variables requires solving discrete optimization problems. This involves a serious computational difficulty as the dimensionality of the variables increases, even in the simplest case of linear models~\citep{natarajan1995sparse}. Hence, certain relaxations or approximations are required for handling such large high-dimensional datasets.

A great deal of progress has been made with regard to relaxation techniques by introducing an $\ell_1$ penalty~\citep{tibshirani1996regression,meinshausen2004consistent,banerjee2006convex,friedman2008sparse}. Its model consistency, that is, whether the estimated model converges to the true model in the large size limit of data, has been extensively studied, particularly in the case of linear models~\citep{knight2000asymptotics,zhao2006model,yuan2007non,wainwright2009sharp,meinshausen2009lasso}. These studies show that a naive usage of the $\ell_1$ penalty affects model consistency in realistic settings: The resultant estimator cannot completely reject variables not present in the true model if there exist non-trivial correlations between covariates, although the normal consistency in the $\ell_2$ sense is retained. This fact motivated the development of further techniques for recovering model consistency in the variable selection context, particularly in the last decade~\citep{zou2006adaptive,bach2008bolasso,meinshausen2010stability,javanmard2014confidence,javanmard2015biasing}. Among them, in this study, the focus is on resampling-based methods. 

Resampling is a versatile idea applicable to a wide range of problems in statistical modeling and is broadly used in machine learning algorithms; some examples can be found in boosting and bagging~\citep{Book:Hastie:09}. In statistics, Efron's bootstrap method is a pioneering example in which resampling is efficiently and systematically used~\citep{efron1994introduction}. In the case of the $\ell_1$-penalized linear regression or Lasso~\citep{tibshirani1996regression}, through a fine evaluation of each variable's probability to be selected (positive probability) in a fairly general setting, Bach showed that it is possible to perform statistically consistent variable selection by utilizing the bootstrap method~\citep{bach2008bolasso}; the associated algorithm is called {\it bootstrapped Lasso} (Bolasso). A similar but more efficient algorithm, called {\it stability selection} (SS), was also implemented by randomizing the penalty coefficient in Lasso~\citep{meinshausen2010stability}. All these examples demonstrate that resampling is powerful and versatile.

A common disadvantage of such resampling approaches is their computational cost. For example, in the Bolasso case, the $\ell_1$-penalized linear regression should be recursively solved according to the number of resampled datasets, which should be sufficiently large so that the positive probability of variables may be estimated. This multiple computational cost precludes the application of such resampling techniques to large datasets. The aim of this study is to avoid this problem by developing an approximation whereby the resampling is conducted in a semi-analytic manner, and to implement it in Lasso. 

Such an approximation was already obtained by \citet{malzahn2003approximate}, where a general framework of semi-analytic resampling was developed using the replica method from statistical mechanics, and was demonstrated in Gaussian process regression. We follow their idea and pursue how it works in Lasso with certain resampling manners.

The remaining of the paper is organized as follows. In the next section, the general framework for semi-analytic resampling using the replica method is reviewed. Concrete formulas in the Lasso case are also provided. After taking the average with respect to the resampling, there remains an intractable statistical model. Handling this model is another key issue, and certain approximations (expected to be exact under certain conditions) should be adopted. In this study, Gaussian approximation is used in conjunction with the so-called cavity method, providing a message passing type algorithm. Its dynamical behavior is analyzed by the so-called state evolution, which guarantees that the algorithm convergence is free from the model dimensionality and the dataset size when covariates are provided as zero-mean independently and identically distributed (i.i.d.) Gaussian random variables. They are explained in \Rsec{Handling}. 
In \Rsec{Numerical}, numerical experiments are carried out on both simulated and real-world datasets to examine the accuracy and efficiency of the proposed semi-analytic method. For processing real-world datasets, an objective criterion for determining the relevance of selected variables is also proposed, based on the addition of noise variables and resampling. The last section concludes the paper.

\section{Formulation for semi-analytic resampling} \Lsec{Formulation}
Let us start from preparing notations and definitions. We denote by $D$ a given dataset, which usually consists of a set of inputs $\V{x}_{\mu}$ and outputs $y_{\mu}$ as $D=\{(\V{x}_{\mu},y_{\mu})\}_{\mu=1}^{M}$, and denote our basic statistics by $\hat{\V{\beta}}(D)=(\hat{\beta}_1(D),\cdots,\hat{\beta}_{N}(D))^{\top}$. They are interpreted as an estimator of true parameters in the true model that is inferred. Based on the Bayesian inference framework, it is assumed that the basic estimator can be expressed as an average over a certain posterior distribution $P(\V{\beta}|\V{\lambda},D)$, that is,
\be
\hat{\V{\beta}}(\V{\lambda},D)
=\int d\V{\beta}\,\V{\beta} P(\V{\beta}|\V{\lambda},D)
\equiv \Ave{\V{\beta}},
\Leq{posterior average}
\ee
where $\Ave{\cdot}$ represents the average over the posterior distribution which is supposed to be constructed as follows:
\be
P(\V{\beta}|\V{\lambda},D)=\frac{1}{Z(\V{\lambda},D)}P_0(\V{\beta}|\V{\lambda})P(D|\V{\beta}).
\Leq{posterior}
\ee
Here, $P_0(\V{\beta}|\V{\lambda})$ is the prior distribution characterized by the parameters $\V{\lambda}$, and $P(D|\V{\beta})$ represents the likelihood. The normalization constant or the partition function can be explicitly expressed by
\be
Z(\V{\lambda},D)=\int d\V{\beta}\,P_0(\V{\beta}|\V{\lambda})P(D|\V{\beta}).
\Leq{Z-integral}
\ee

Considering the construction  of a subsample by resampling from the full data $D$ of size $M$ with replacement, let an indicator vector $\V{c}=(c_1,\cdots,c_{M})^{\top}$ specify any such subsample; each $c_{\mu}$ is a non-negative integer counting the number of occurrences of the $\mu$-th datapoint of $D$ in the subsample. A subsample identified by $\V{c}$ is denoted by $D_{\V{c}}$. Specifying a resampling defines the distribution $P(\V{c})$ of $\V{c}$, and the question is the behavior of the basic estimators and their functions with respect to the average over $P(\V{c})$. In some resampling techniques, additional randomization is introduced in the prior distribution~\citep{meinshausen2010stability}. Hence, an average over the parameters $\V{\lambda}$ is further considered, and the distribution $P(\V{\lambda})$ of $\V{\lambda}$ is introduced. For notational simplicity, the average over these distributions is denoted by square brackets with appropriate subscripts as follows:
\nbe
\lsb (\cdots) \rsb_{\V{c},\V{\lambda}}
\equiv
\sum_{\V{c}}\int d\V{\lambda} (\cdots) 
P(\V{c})P(\V{\lambda}).
\nee
The average of this type is called {\em configurational average} throughout this paper.

\subsection{General theory of semi-analytic resampling}\Lsec{General theory}
The purpose of resampling is to obtain the distribution of the basic estimator
\be
P(\beta_i)=\left [\delta \left (\beta_i-\hat{\beta}_i(\V{\lambda},D_c)  \right )\right ]_{\V{c},\V{\lambda}}, 
\label{boot_strapp_dist}
\ee
which is reduced to computing the moments $\lsb \hat{\beta}_i^{r}(\V{\lambda},D_{\V{c}}) \rsb_{\V{c},\V{\lambda}}$ of arbitrary degree $\forall{r}\in \mN$. We thus show that these moments can be evaluated in the following manner.

By definition, the moment $\lsb \hat{\beta}_i^r(\V{\lambda},D_{\V{c}}) \rsb_{\V{c},\V{\lambda}}$ with $r\in \mN$ can be written as
\be
&&
\lsb \hat{\beta}_i^{r}(\V{\lambda},D_{\V{c}}) \rsb_{\V{c},\V{\lambda}}
=
\lsb 
\frac{1}{Z^r(\V{\lambda},D_{\V{c}})} 
\int 
\lbb
\prod_{a=1}^{r} d\V{\beta}^{a}\, \beta^a_{i} P_0(\V{\beta}^a|\V{\lambda}) 
P(D_{\V{c}}|\V{\beta}^a) 
\rbb
\rsb_{\V{c},\V{\lambda}}
\Leq{no replica} \\ &&
=
\lim_{n\to 0}
\lsb
Z^{n-r}(\V{\lambda},D_{\V{c}})
\int 
\lbb 
\prod_{a=1}^{r} d\V{\beta}^{a}\, \beta^a_{i} P_0(\V{\beta}^a|\V{\lambda}) 
P(D_{\V{c}}|\V{\beta}^a) 
\rbb
\rsb_{\V{c},\V{\lambda}}
\Leq{replica-real} \\ &&
\doteq
\lim_{n\to 0}
\lsb 
\int \lbb  \prod_{b=1}^{n} d\V{\beta}^{b} \rbb
\lbb \prod_{a=1}^{r}  \beta^a_{i} \rbb
\lbb 
\prod_{b=1}^{n} 
P_0(\V{\beta}^b|\V{\lambda}) 
P(D_{\V{c}}|\V{\beta}^b) 
\rbb
\rsb_{\V{c},\V{\lambda}}.
\Leq{replica}
\ee
The evaluation of \Req{no replica} is technically difficult owing to the existence of $Z^r(\V{\lambda},D_{\V{c}})$ in the denominator: The negative power of the partition function does not allow the analytical evaluation of the average over $\V{c}$ and $\V{\lambda}$, even if $P(\V{c})$ and $P(\V{\lambda})$ take reasonable forms. To overcome this difficulty,  an auxiliary parameter $n$ is introduced in \Req{replica-real}. The exponent $n$ is assumed to be a positive integer larger than $r$ in \Req{replica}; thus, the power of the partition function can be expanded in the integral form \NReq{Z-integral}. Integral variables $\{ \V{\beta}^1,\V{\beta}^2,\ldots,\V{\beta}^n \}$ are termed {\em replicas} since they are regarded as constituting $n$ copies of the original system. In \Req{replica}, the configurational average can be analytically computed under appropriate approximations. This yields an expression as a function of $n$ that is analytically continuable from $\mN$ to $\mR$. The $n \to 0$ limit is taken by employing the analytically continued expression after all computations. The evaluation technique based on these procedures is often called the {\em replica method}~\citep{mezard1987spin,nishimori2001statistical,dotsenko2005introduction}. Evidently, this technique is not justified in a strict sense, but it is known that the replica method gives correct results for many problems. Justification of the replica method is known to be difficult~\citep{talagrand2003spin}, and here we leave it as a future work and just employ the method for our purpose. In the present problem, it is far from trivial to obtain an analytically continuable expression, and below we concentrate on its derivation.

Accordingly, the configurational average of any power of the basic estimator can be evaluated from the following quantities:
\nbe
&&
\Xi \lb \lbb \V{\beta}^{a}\rbb_{a=1}^{n} \left. \right| D  \rb
\equiv
\int 
\lbb  \prod_{a=1}^{n} d\V{\beta}^{a} \rbb 
\lsb 
\prod_{a=1}^{n} 
P_0(\V{\beta}^a|\V{\lambda}) 
P(D_{\V{c}}|\V{\beta}^a) 
\rsb_{\V{c},\V{\lambda}}
, \\ &&
P \lb \lbb \V{\beta}^{a}\rbb_{a=1}^{n} \left. \right| D  \rb
\equiv
\frac{1}{\Xi \lb \lbb \V{\beta}^{a}\rbb_{a=1}^{n} \left. \right| D  \rb}
\lsb 
\prod_{a=1}^{n} 
P_0(\V{\beta}^a|\V{\lambda}) 
P(D_{\V{c}}|\V{\beta}^a) 
\rsb_{\V{c},\V{\lambda}},
\nee
where the former is called replicated partition function and the latter is called replicated Boltzmann distribution. The average over the replicated Boltzmann distribution is denoted by
\be
\overline{\lb \cdots \rb}\equiv \int \lbb \prod_{a=1}^{n}d\V{\beta}^{a} \rbb \lb \cdots \rb P \lb \lbb \V{\beta}^{a}\rbb_{a=1}^{n} \left. \right| D  \rb.
\Leq{total ave}
\ee
It follows from \Reqss{no replica}{replica} that this average converges to the desired total average over the posterior and the configurational variables $\V{c}$ and $\V{\lambda}$ as $n\to 0$. The next step is to compute this average. The nontrivial technical difficulties are in the integrations with respect to replicas $\{\V{\beta}^{a}\}_{a=1}^{n}$ and in deriving a functional form that is analytically continuable with respect to $n$. These will be handled after considering the specific case of Lasso.

\subsection{Specifics to Lasso with independent sampling} \Lsec{Specifics to}
The above general theory is now rephrased in the context of Lasso. Given a set of covariates $\V{x}_{\mu}\in \mR^{N}$ and responses $y_{\mu}\in \mR$, $D=\{ (\V{x}_{\mu},y_{\mu}) \}_{\mu=1}^{M}$, the usual estimator by Lasso is
\nbe
\hat{\V{\beta}}(\lambda,D)
=\argmin_{\V{\beta}}
\lbb 
\frac{1}{2}\sum_{\mu=1}^{M}\lb y_{\mu}-\sum_{i}x_{\mu i}\beta_i\rb^2+\lambda ||\V{\beta}||_1
\rbb.
\nee
In contrast to this, considering Bolasso and SS, given the resampled data $D_{\V{c}}$ and the randomized penalty coefficients $\V{\lambda}=(\lambda_1,\cdots,\lambda_N)^{\top}$, the following estimator is introduced:
\be
\hat{\V{\beta}}(\V{\lambda},D_{\V{c}})
=\argmin_{\V{\beta}}
\lbb 
\frac{1}{2}\sum_{\mu=1}^{M}c_{\mu}\lb y_{\mu}-\sum_{i}x_{\mu i}\beta_i\rb^2
+\sum_{i=1}^{N}\lambda_i |\beta_i|
\rbb.
\Leq{Lasso-random}
\ee
For representing this in terms of the posterior average, the following quantities are introduced:
\nbe
&&
\mathcal{H}(\V{\beta}|\V{\lambda},D_{\V{c}})
=
\frac{1}{2}
\sum_{\mu=1}^{M}c_{\mu}\lb y_{\mu}-\sum_{i}x_{\mu i}\beta_i\rb^2
+\sum_{i=1}^{N}\lambda_i |\beta_i|,
\\ &&
Z_{\gamma}(\V{\lambda},D_{\V{c}})=\int d\V{\beta}\,e^{-\gamma \mathcal{H}(\V{\beta}|\V{\lambda},D_{\V{c}})},
\\ &&
P_{\gamma}(\V{\beta}|\V{\lambda},D_{\V{c}})=\frac{1}{Z_{\gamma}}
e^{-\gamma \mathcal{H}(\V{\beta}|\V{\lambda},D_{\V{c}})},
\nee
where these quantities are called (in the order they appear) Hamiltonian, partition function, and Boltzmann distribution, in accordance with physics terminology. As $\gamma \to \infty$, the Boltzmann distribution converges to a pointwise measure at $\hat{\V{\beta}}(\V{\lambda},D_{\V{c}})$ and thereby becomes the desired posterior distribution, allowing the identification of the Boltzmann distribution with the posterior distribution; thus, the average over the Boltzmann distribution is thereafter denoted by  $\Ave{\cdots}$ introduced in \Req{posterior average}\footnote{It is assumed that the order of the integration and the $\gamma \to \infty$ limit may be changed.}. The prior distribution $P_{0}(\V{\beta}|\V{\lambda})$ and the likelihood $P(D|\V{\beta})$ in \Req{posterior} correspond to the factors $e^{-\gamma \sum_{i} \lambda_i|\beta_i|}$ and $e^{-\frac{\gamma}{2}\sum_{\mu=1}^{M}c_{\mu}\lb y_{\mu}-\sum_{i}x_{\mu i}\beta_i\rb^2}$ in the Boltzmann distribution, respectively.

Considering Bolasso and SS, we draw the subsample $\V{c}$ of a fixed size $m$ in an unbiased manner. Each sample comes out with a probability of $M^{-m}m!/(c_1!\cdots c_M!)$, which has a weak dependency among $\{ c_{\mu} \}_{\mu}$. However, this dependency is not essential for large $M$ and $m$; thus, it is ignored. Using Stirling's formula $m!\approx (\frac{m}{e})^m$ in conjunction with the relation $m=\sum_{\mu=1}^M c_\mu $, $c_{\mu}$ is approximately regarded as an i.i.d. variable from a Poisson distribution of mean $\tau=m/M$, namely,
\nbe
P(\V{c}) 
=\prod_{\mu=1}^{M}\frac{\tau^{c_{\mu}} }{c_{\mu}!}e^{-\tau}
=\prod_{\mu=1}^{M}P(c_{\mu})
.
\nee
It is also natural to require that $P(\V{\lambda})$ is factorized into a batch of identical distributions as follows:
\nbe
P(\V{\lambda})=\prod_{i=1}^{N}P(\lambda_i).
\nee
At this point, the explicit form of $P(\lambda_i)$ is not specified. These factorized natures allow us to express the replicated Boltzmann distribution as
\be
P_{\gamma}(\{ \V{\beta}^{a} \}_{a=1}^{n}|D)
\propto
\lsb 
e^{
-\gamma \sum_{a=1}^{n} \mathcal{H}(\V{\beta}^{a}|\V{\lambda},D_{\V{c}})
}
\rsb_{\V{c},\V{\lambda}}
=
\prod_{\mu=1}^{M} \Phi_{\mu}\lb \{ \V{\beta}_i \}_{i=1}^{N} \rb
\prod_{i=1}^{N} \Psi \lb \V{\beta}_i  \rb
 \Leq{replicatedBD},
\ee
where
\nbe
&&
\Phi_{\mu}\lb \{ \V{\beta}_i \}_{i=1}^{N} \rb
=\lsb e^{-\frac{1}{2}\gamma c\sum_{a}(y_{\mu}-\sum_i x_{\mu i}\beta^{a}_i )^2}  \rsb_{c}
, \\ &&
\Psi \lb \V{\beta}_i  \rb
=
\lsb e^{-\gamma \lambda \sum_{a=1}^{n} |\beta_i^a|}  \rsb_{\lambda},
\nee
and the vector notation $\V{\beta}_i=(\beta^{a}_{i})_a$ was introduced for later convenience. $\Phi_{\mu}$ is hereafter called $\mu$-th potential function. 

To proceed further, an approximation should be introduced to make the right-hand side of \Req{replicatedBD} tractable. Although there are several ways for that, we here make an approximation based on the cavity method from statistical physics. The details are in the next section.

\section{Handling the replicated system} \Lsec{Handling}
We below work with the cavity method, or the belief propagation (BP) in computer science, and use a Gaussian approximation. This treatment is essentially identical to that used in deriving the known approximate message passing (AMP)~\citep{kabashima2003cdma,donoho2009message}, and can be justified if each covariate is i.i.d. from Gaussian distributions~\citep{bayati2011dynamics,barbier2017mutual}. For treating nontrivial correlations between covariates, more sophisticated approximations~\citep{opper2001adaptive,opper2001tractable,opper2005expectation,kabashima2014signal,cakmak2014s,cespedes2014expectation,rangan2016vector,ma2017orthogonal,takeuchi2017rigorous} are required. Such extensions are left as future work.

\BReq{replicatedBD} implies that the replicated Boltzmann distribution naturally has a factor graph structure. Hence, by the cavity method, \Req{replicatedBD} can be ``approximately'' decomposed into two messages as follows:
\be
&&
\BPC{\phi}_{\mu \to i}(\V{\beta}_{i})=\frac{1}{Z_{\mu \to i}}\int \prod_{j(\neq i)}d\V{\beta}_{j}
~\Phi_{\mu}\lb \{ \V{\beta}_i \}_{i=1}^{N} \rb
\prod_{j(\neq i)} \phi_{ j \to \mu }(\V{\beta}_j), 
\Leq{BP1}
\\ &&
\phi_{i \to \mu}(\V{\beta}_{i})=
\frac{1}{Z_{i \to \mu}}
\Psi \lb \V{\beta}_i  \rb
\prod_{\nu (\neq \mu)} \BPC{\phi}_{ \nu \to i }(\V{\beta}_i), 
\Leq{BP2}
\ee
where $Z_{\mu \to i},Z_{i \to \mu}$ are normalization factors that are not relevant and will be discarded below.  The average of \Req{total ave} can be computed by employing this set of equations.

\subsection{Gaussian approximation on cavity method}\Lsec{Gaussian approximation}
A crucial observation for assessing \Reqs{BP1}{BP2} is that the residual $R^a_{\mu}=y_{\mu}-\sum_{j}x_{\mu j}\beta_{j}^a$ appearing in $\Phi_{\mu}$ has a sum of a large number of random variables; thus, the central limit theorem justifies treating it as a Gaussian variable with the appropriate mean and variance. This consideration leads to the following decomposition in \Req{BP1}
\nbe
R^a_{\mu}+x_{\mu i}\beta_{i}^a 
\equiv 
y_{\mu}-\sum_{j(\neq i)}x_{\mu j}\beta_{j}^a
\approx
y_{\mu}-\sum_{j(\neq i)}x_{\mu j}\overline{\beta}_{j}^{\bs \mu}
+Z^a_{\mu i}
\equiv
r_{\mu i}
+Z^a_{\mu i},
\nee
where $Z^a_{\mu i}$ is a zero-mean Gaussian variable whose covariance is equivalent to that of $\sum_{j(\neq i)}x_{\mu j}\beta_{j}^a$, and $\overline{\lb \cdots \rb}^{\bs \mu}$ is the average over $\prod_{j} \phi_{j \to \mu}(\V{\beta}_{j})$ or the replicated Boltzmann distribution without $\mu$-th potential function. This shows that it is difficult to consider the correlation between $\beta_i$ and $\beta_j$ with different $i,j$ in the present framework. This could be overcome even in the cavity method framework~\citep{opper2001adaptive,opper2001tractable}; however, it is beyond the scope of this study.

The so-called replica symmetry is now assumed in the messages. This symmetry implies that any permutation of the replicas $\{\beta_{i}^{1},\cdots,\beta_i^{n}\}$ yields an identical message, which in turn implies, by De Finetti's theorem~\citep{hewitt1955symmetric}, that the messages can be expressed as
\subbe
\nbe
&&
\phi_{i \to \mu}(\V{\beta}_{i})
=\int d\mathcal{F}\rho_{i \to \mu}(\mathcal{F})\prod_{a=1}^{n}\mathcal{F}(\beta_{i}^a),
\\ &&
\BPC{\phi}_{\mu \to i}(\V{\beta}_{i})
=\int d\mathcal{G}\rho_{\mu \to i}(\mathcal{G})\prod_{a=1}^{n}\mathcal{G}(\beta_{i}^a),
\nee
\subee
where $d\mathcal{F} \rho_{i \to \mu}(\mathcal{F})$ and $d\mathcal{G} \rho_{i \to \mu}(\mathcal{G})$ are probability measures over the probability distribution functions $\mathcal{F}$ and $\mathcal{G}$, respectively. From this form, two different variances naturally emerge: 
\subbe
\nbe
&&
V_{i}^{\bs \mu}\equiv \overline{\lb \beta_i^a \rb^2 - \beta_i^a \beta_i^b }^{\bs \mu}
=
\int d \V{\beta}_i
\lb \lb \beta_i^a \rb^2 - \beta_i^a \beta_i^b \rb
\int d\mathcal{F} \rho_{i \to \mu}(\mathcal{F})
\prod_{c=1}^{n}\mathcal{F}(\beta_{i}^c)
\no \\ &&
=\int d\mathcal{F} \rho_{i \to \mu}(\mathcal{F})\lbb \int \mathcal{F}(\beta)\beta^2 d\beta- \lb \int \mathcal{F}(\beta)\beta d\beta \rb^2 \rbb,
\\ &&
W_{i}^{\bs \mu}
\equiv
\overline{\beta_i^a \beta_i^b}^{\bs \mu} - \overline{\beta_i^a}^{\bs \mu}\overline{\beta_i^b}^{\bs \mu}
\no \\ &&
=
\int d\mathcal{F} \rho_{i \to \mu}(\mathcal{F}) \lb \int \mathcal{F}(\beta)\beta d\beta \rb^2
-
\lb \int d\mathcal{F} \rho_{i \to \mu}(\mathcal{F})  \int \mathcal{F}(\beta)\beta d\beta \rb^2,
\nee
\subee
where it was assumed that $a\neq b$. Below, the $\mu$-dependence of $V_i^{\bs \mu}$ and $W_i^{\bs \mu}$ is ignored, as it is small, and let
\subbe
\nbe
&&
V_{i}^{\bs \mu}\approx V_i=\overline{\lb \beta_i^a \rb^2 - \beta_i^a \beta_i^b }, 
\\ &&
W_{i}^{\bs \mu}\approx W_i=\overline{\beta_i^a \beta_i^b} - \overline{\beta_i^a}\overline{\beta_i^b}.
\nee
\subee
It can be implied that $V_i$ describes the average of the variance inside a fixed resampled dataset, whereas $W_i$ represents the inter-sample variance. Using $V_i$ and $W_i$, the covariance of $\V{\beta}_i$ is generally written as 
\nbe
\Cov{\beta_{i}^a}{\beta_{i}^b}{\bs \mu}
\equiv 
\overline{\beta_i^a \beta_i^b}^{\bs \mu} - \overline{\beta_i^a}^{\bs \mu}\overline{\beta_i^b}^{\bs \mu}
=W_i^{\bs \mu}+V_i^{\bs \mu}\delta_{ab}
\approx W_i+V_i\delta_{ab}.
\nee
Using this relation, the covariance of $\{ Z_{\mu i}^{a} \}_a$ can be expressed as
\nbe
&&
\Cov{Z_{\mu i}^{a}}{Z_{\mu i}^{b}}{\bs \mu}
=
\sum_{j,k(\neq i)}x_{\mu i}x_{\mu j}\Cov{\beta_{j}^{a}}{\beta_{k}^{b}}{\bs \mu}
\approx
\sum_{j}x_{\mu j}^2\Cov{\beta_{j}^{a}}{\beta_{j}^{b}}{\bs \mu}
\no \\ &&
\approx
\sum_{j}x_{\mu j}^2(W_j+V_j\delta_{ab})
\equiv
W_{\mu}+V_{\mu}\delta_{ab}.
\nee
The correlation between $\beta_{j}$ and $\beta_k$ for $j\neq k$ is neglected in the second sum because it is not taken into account in the present framework, as explained above. Moreover, the addition of the $i$-th term in the same sum does not affect the following discussion because it is sufficiently small in the summation. Based on these observations, $Z_{\mu i}^{a}$ is decomposed as follows:
\nbe
Z_{\mu i}^{a}=D_{\mu i}+\Delta_{\mu i}^{a},
\nee
where $D_{\mu i}$ and $\Delta_{\mu i}^{a}$ are zero-mean Gaussian variables whose covariances are $\Cov{D_{\mu i}}{D_{\mu i}}{\bs \mu}=W_{\mu},~\Cov{D_{\mu i}}{\Delta_{\mu i}^{a}}{\bs \mu}=0,~\Cov{\Delta_{\mu i}^{a}}{\Delta_{\mu i}^{b}}{\bs \mu}=V_{\mu}\delta_{ab}$.

$\BPC{\phi}_{\mu \to i}$ can now be computed. The integration with respect to $\{\V{\beta}_{j} \}_{j(\neq i)}$ in \Req{BP1} is replaced by that over $D_{i}$ and $\Delta_{i}^{a}$. The result is
\nbe
&&
\BPC{\phi}_{\mu \to i}(\V{\beta}_{i})
\approx 
\int dD_{\mu i} P(D_{\mu i})\int \prod_{a}d\Delta^a_{\mu i} P(\Delta^a_{\mu i})
\lsb e^{-\frac{1}{2}\gamma c\sum_{a}(r_{\mu i}-x_{\mu i}\beta^{a}_i+D_{\mu i}+\Delta_{\mu i}^{a} )^2}   
\rsb_{c}
\no \\ &&
=
\lsb
U(c)^{-\frac{1}{2}}\lb 1+c\gamma V_{\mu} \rb^{-\frac{n}{2}}
e^{-\frac{1}{2}S(c)\sum_{a}\lb r_{\mu i}-x_{\mu i} \beta_{i}^{a} \rb^2
+\frac{1}{2}T(c)\lb \sum_{a} (r_{\mu i}-x_{\mu i} \beta_{i}^{a}) \rb^2}
\rsb_c,
\nee
where
\nbe
&&
S(c)=\frac{c\gamma}{1+c\gamma V_{\mu}},
\\ &&
T(c)=\frac{c^2\gamma^2 W_{\mu} }{(1+c\gamma V_{\mu})(1+c\gamma (V_{\mu}+nW_{\mu}))},
\\ &&
U(c)=\frac{1+c\gamma (V_{\mu}+nW_{\mu})}{1+c\gamma V_{\mu}}.
\nee
$\ln \BPC{\phi}_{\mu \to i}(\V{\beta}_{i})$ can be expanded with respect to $x_{\mu i} \beta_{i}^{a}$, and the expansion up to the second order leads to an effective Gaussian approximation of the message, yielding
\nbe
\BPC{\phi}_{\mu \to i}(\V{\beta}_{i})
\approx 
e^{
-\frac{1}{2}\gamma \tilde{A}_{\mu \to i} \sum_{a}\lb \beta_{i}^a \rb^2 
+\frac{1}{2}\gamma^2 \tilde{C}_{\mu \to i}\lb \sum_{a} \beta_{i}^a \rb^2
+\gamma \tilde{B}_{\mu \to i}\sum_{a} \beta_{i}^a
},
\nee
where
\subbe
\Leq{horizontal-BP}
\be
&&
\hspace{-10mm}
\tilde{A}_{\mu \to i}=\lsb\frac{c}{1+c\gamma V_{\mu}}\rsb_c x_{\mu i}^2,
\\ &&
\hspace{-10mm}
\tilde{B}_{\mu \to i}=\lsb\frac{c}{1+c\gamma V_{\mu}}\rsb_c r_{\mu i} x_{\mu i},
\\ &&
\hspace{-10mm}
\tilde{C}_{\mu \to i}=
\lbb
\lsb \lb \frac{c}{1+c\gamma V_{\mu}} \rb^2\rsb_c W_{\mu}
+
\lb \lsb \lb \frac{c}{1+c\gamma V_{\mu}} \rb^2\rsb_c-\lsb\frac{c}{1+c\gamma V_{\mu}} \rsb_c^2 \rb \lb r_{\mu i} \rb^2
\rbb x_{\mu i}^2.
\ee
\subee
It should be noted that the $n\to 0$ limit was already taken in these coefficients. 

Owing to the Gaussian approximation of $\BPC{\phi}_{\mu \to i}(\V{\beta}_{i})$, the marginal distribution of the replicated Boltzmann distribution can be simply written as
\be
&&
P_i(\V{\beta}_i|D)
\equiv 
\int \prod_{j(\neq i)}d\V{\beta}_j P(\{\V{\beta}_i\}_i|D)
\propto 
\Psi(\V{\beta}_i)
\prod_{\mu}\BPC{\phi}_{\mu \to i}(\V{\beta}_{i})
\no \\ &&
\approx
\lsb
\int Dz~
e^{
\gamma \lb 
-\frac{1}{2}A_{i} \sum_{a}\lb \beta_{i}^a \rb^2 
+ (B_{i}+ \sqrt{C_{i}}z)\sum_{a} \beta_{i}^a
-\lambda \sum_{a}|\beta_i^a|
\rb
}
\rsb_{\lambda},
\Leq{fullmarginal}
\ee
where $Dz=dz e^{-\frac{1}{2}z^2}/\sqrt{2\pi}$, and the following identity was used:
\nbe
e^{\frac{1}{2}Cx^2}=\int Dz~e^{\sqrt{C}zx}.
\nee
Moreover, $A_{i}=\sum_{\mu}\tilde{A}_{\mu \to i},~B_{i}=\sum_{\mu}\tilde{B}_{\mu \to i},$ and $C_{i}=\sum_{\mu}\tilde{C}_{\mu \to i}$. All replicas are now factorized, and the average can be taken for each replica independently, allowing passing to the $n\to 0$ limit by analytic continuation from $\mN$ to $\mR$ with respect to $n$. For example, the mean of $\beta^a_i$ is computed as
\nbe
&&
\overline{\beta^{a}_{i}}
=
\lsb
\int Dz
\lb 
\int d\beta\, e^{
\gamma \lb 
-\frac{1}{2}A_{i} \beta^2 
+ (B_{i}+ \sqrt{C_{i}}z)\beta_{i}
-\lambda|\beta|
\rb
}
\rb^n
\rsb_{\lambda}^{-1}
\no \\ && \times
\lsb
\int Dz
\int d\beta\, \beta e^{
\gamma \lb 
-\frac{1}{2}A_{i} \beta^2 
+ (B_{i}+ \sqrt{C_{i}}z)\beta_{i}
-\lambda|\beta|
\rb
} 
\lb 
\int d\beta\, e^{
\gamma \lb 
-\frac{1}{2}A_{i} \beta^2 
+ (B_{i}+ \sqrt{C_{i}}z)\beta_{i}
-\lambda|\beta|
\rb
}
\rb^{n-1}
\rsb_{\lambda}
\no \\ &&
\xrightarrow[]{n \to 0}
\lsb
\int Dz\frac{
\int d\beta\, \beta e^{
\gamma \lb 
-\frac{1}{2}A_{i} \beta^2 
+ (B_{i}+ \sqrt{C_{i}}z)\beta_{i}
-\lambda|\beta|
\rb
} 
}{
\int d\beta\, e^{
\gamma \lb 
-\frac{1}{2}A_{i} \beta^2 
+ (B_{i}+ \sqrt{C_{i}}z)\beta_{i}
-\lambda|\beta|
\rb
}
}
\rsb_{\lambda}
\xrightarrow[]{\gamma \to \infty}
\lsb
\int Dz
S_{\lambda}(B_i+\sqrt{C_i}z;A_i)
\rsb_{\lambda},
\nee
where $S_{\lambda}$ is the so-called soft thresholding function
\nbe
S_{\lambda}(x;A)=\frac{1}{A}(x-\lambda \sgn{x})\theta(|x|-\lambda),
\nee
and $\theta(x)$ is the step function that is equal to $1$ if $x>0$ and $0$ otherwise. Thus, the mean of the basic estimator takes a reasonable form. 

In the above average, it is assumed that the coefficients $A_i,B_i,C_i$ remain finite as $\gamma\to \infty$. This is the case if the following holds:
\nbe
\chi_{\mu}\equiv \gamma V_{\mu}=\sum_{i}x_{\mu i}^2 \gamma V_{i} \to O(1),~(\gamma \to \infty).
\nee
This scaling is consistent, which can be shown by
\nbe
&&
\hspace{-0.8cm}
\chi_i \equiv \gamma V_{i}
=\gamma \lbb \overline{ \lb \beta^{a}_i\rb^2 -\beta^{a}_i\beta^{b}_i }\rbb
\no \\ &&
\hspace{-0.8cm}
\xrightarrow[]{n \to 0}
\lsb
\int Dz\,
\gamma 
\lbb
\frac{
\int d\beta\, \beta^2 e^{
\gamma \lb 
-\frac{1}{2}A_{i} \beta^2 
+ (B_{i}+ \sqrt{C_{i}}z)\beta_{i}
-\lambda|\beta|
\rb
} 
}{
\int d\beta\, e^{
\gamma \lb 
-\frac{1}{2}A_{i} \beta^2 
+ (B_{i}+ \sqrt{C_{i}}z)\beta_{i}
-\lambda|\beta|
\rb
}
}
-
\lb
\frac{
\int d\beta\, \beta e^{
\gamma \lb 
-\frac{1}{2}A_{i} \beta 
+ (B_{i}+ \sqrt{C_{i}}z)\beta_{i}
-\lambda|\beta|
\rb
} 
}{
\int d\beta\, e^{
\gamma \lb 
-\frac{1}{2}A_{i} \beta^2 
+ (B_{i}+ \sqrt{C_{i}}z)\beta_{i}
-\lambda|\beta|
\rb
}
}
\rb^2
\rbb
\rsb_{\lambda}
\no \\ &&
\hspace{-0.8cm}
\xrightarrow[]{\gamma \to \infty}
\frac{1}{A_{i}}
\lsb
\int Dz 
\theta\lb |B_i+\sqrt{C_i}z|-\lambda \rb
\rsb_{\lambda}
=
\Part{\overline{\beta_{i}}}{B_{i}}{}.
\nee
In contrast to $V_i$, the inter-sample fluctuation $W_i$ takes a finite value even as $\gamma \to \infty$. Its explicit form is
\nbe
&&
W_i =\overline{ \beta^{a}_i\beta^{b}_i }-\overline{\beta^{a}_i}\overline{\beta^{b}_i}
\no \\ &&
\xrightarrow[]{n\to 0,~\gamma \to \infty}
\lsb
\int Dz 
S^2_{\lambda}\lb B_i+\sqrt{C_i}z; A_{i}\rb
\rsb_{\lambda}-
\lb
\lsb
\int Dz 
S_{\lambda}\lb B_i+\sqrt{C_i}z; A_{i}\rb
\rsb_{\lambda}
\rb^2.
\nee

It follows that any moment of the basic estimator can be computed from \Req{fullmarginal}, once the coefficients $\{ A_{i},B_{i},C_{i} \}_{i=1}^{N}$ are correctly estimated. Hence, the next task is to derive a set of self-consistent equations for the coefficients and to construct an algorithm for solving it. 

\subsection{Self-consistent equations and a message passing algorithm} \Lsec{Self-consistent equations}
Using \Req{BP2}, we obtain
\nbe
&&
\phi_{i \to \mu}(\V{\beta}_i)
\propto 
\Psi\lb \V{\beta}_i \rb \prod_{\nu(\neq \mu)}\BPC{\phi}_{\nu \to i}(\V{\beta}_{i})
\no \\ &&
\approx
\lsb
\int Dz~
e^{
\gamma \lb 
-\frac{1}{2}A_{i\to \mu} \sum_{a}\lb \beta_{i}^a \rb^2 
+ (B_{i\to \mu}+ \sqrt{C_{i\to \mu}}z)\sum_{a} \beta_{i}^a
-\lambda \sum_{a}|\beta_i^a|
\rb
}
\rsb_{\lambda}
,
\nee
where $A_{i\to \mu}=\sum_{\nu(\neq \mu)}\tilde{A}_{\nu \to i},~B_{i\to \mu}=\sum_{\nu(\neq \mu)}\tilde{B}_{\nu \to i},$ and $C_{i\to \mu}=\sum_{\nu(\neq \mu)}\tilde{C}_{\nu \to i}$. Inserting this into \Req{BP1}, we can derive a set of self-consistent equations determining all the cavity coefficients $\{ A_{i\to \mu},B_{i\to \mu},C_{i\to \mu},\tilde{A}_{\mu \to i},\tilde{B}_{\mu \to i},\tilde{C}_{\mu \to i} \}_{i,\mu}$. This procedure suggests an iterative algorithm, conventionally called BP algorithm, which is schematically described as follows:
\subbe
\Leq{BPschematic}
\be
&&
\{ \tilde{A}_{\mu \to i},\tilde{B}_{\mu \to i},\tilde{C}_{\mu \to i} \}^{(t)}
\leftarrow 
\{ \overline{\beta_i}^{\bs \mu}, V_{i},W_{i} \}^{(t)},
\\ &&
\{ A_{i \to \mu}, B_{i \to \mu},C_{i \to \mu} \}^{(t+1)} \leftarrow \{ \tilde{A}_{\mu \to i},\tilde{B}_{\mu \to i},\tilde{C}_{\mu \to i} \}^{(t)},
\\ &&
\{ \overline{\beta_i}^{\bs \mu}, V_{i},W_{i} \}^{(t+1)} \leftarrow 
\{ A_{i \to \mu}, B_{i \to \mu},C_{i \to \mu} \}^{(t+1)},
\ee
\subee
where $t=0,1,\cdots,$ denotes the algorithm time step. If the BP algorithm converges, the full coefficients $\{ A_{i},B_{i},C_{i} \}_{i=1}^{N}$ are given from the converged values of $\{ \tilde{A}_{\mu \to i},\tilde{B}_{\mu \to i},\tilde{C}_{\mu \to i} \}_{i,\mu}$. However, this algorithm is not particularly efficient because its computational cost is $O(NM^2)$. 

A more efficient algorithm is derived by approximately rewriting the cavity coefficients and the cavity mean $\overline{\V{\beta}}^{\bs \mu}$ using the full coefficients $\{ A_{i},B_{i},C_{i}\}_{i}$. To this end, $\phi_{i\to \mu}(\V{\beta}_i)$ is connected with $P_i(\V{\beta}_i|D)$ in a perturbative manner. Comparing the coefficients, it follows that the difference between $A_{i}$ and $A_{i \to \mu}$ is negligibly small, as it is proportional to $x_{\mu i}^2 =O(1/N)$. The same is true between $C_{i}$ and $C_{i \to \mu}$. Hence, the relevant difference is only $\Delta B^{(t)}_i=B^{(t)}_{i}-B^{(t)}_{i\to \mu}$ and is expressed as
\nbe
&&
\Delta B^{(t)}_i
=
\lsb \frac{c}{1+c\chi_{\mu}^{(t-1)}} \rsb_c r^{(t-1)}_{\mu i} x_{\mu i}
=
a^{(t-1)}_{\mu}x_{\mu i}
+
\lsb \frac{c}{1+c\chi_{\mu}^{(t-1)}} \rsb_c  \lb \overline{\beta_{i}}^{\bs \mu} \rb^{(t-1)} x^2_{\mu i}
\no \\ &&
\approx 
a^{(t-1)}_{\mu}x_{\mu i},
\nee
where
\be
a^{(t)}_{\mu}
\equiv \lsb \frac{c}{1+c\chi_{\mu}^{(t)}} \rsb_c  \lb y_{\mu}-\sum_{j}x_{\mu j}\lb \overline{\beta_{j}}^{\bs \mu} \rb^{(t)} \rb
= \lsb \frac{c}{1+c\chi_{\mu}^{(t)}} \rsb_c   \lb r_{\mu i}^{(t)}-x_{\mu i}\lb \overline{\beta_{i}}^{\bs \mu} \rb^{(t)} \rb.
\Leq{cavity residual}
\ee
Accordingly, the difference between $\overline{\beta_i}^{\bs \mu}$ and $\overline{\beta_{i}}$ is computed as
\be
\lb \overline{\beta_i}^{\bs \mu} \rb^{(t)}
\approx 
\overline{\beta_{i} }^{(t)} - \frac{ \partial \overline{\beta_{i}}^{(t)} }{ \partial B^{(t)}_{i} } \Delta B^{(t)}_i
=\overline{\beta_{i}}^{(t)}-\chi_i^{(t)} 
a_{\mu}^{(t-1)} x_{\mu i},
\Leq{linear response}
\ee
Inserting \Req{linear response} into \Req{cavity residual} yields 
\nbe
a_{\mu}^{(t)}= 
\lsb \frac{c}{1+c\chi_{\mu}^{(t)}} \rsb_c  \lb y_{\mu}-\sum_{j}x_{\mu j} \overline{\beta_{j}}^{(t)} +\chi_{\mu}^{(t)}a_{\mu}^{(t-1)} \rb,
\nee
where the last term is interpreted as the Onsager reaction term in physics. Rewriting $r_{\mu i}$ using $a_{\mu}$ in the right-hand sides of \Req{horizontal-BP} and collecting several factors, a simplified message passing algorithm corresponding to \Req{BPschematic} is obtained as follows:
\subbe
\Leq{AMPR}
\be
&&
\chi_{\mu}^{(t)}=\sum_{i}x_{\mu i}^2\chi_{i}^{(t)},
\\ &&
W_{\mu}^{(t)}=\sum_{i}x_{\mu i}^2W_{i}^{(t)},
\\ &&
\lb f^{(t)}_{1\mu}, f^{(t)}_{2\mu}\rb= 
\lb
 \lsb\frac{c}{1+c\chi^{(t)}_{\mu}}\rsb_c, 
 \lsb\lb \frac{c}{1+c\chi^{(t)}_{\mu}} \rb^2\rsb_c
\rb,
\\ &&
a_{\mu}^{(t)}= f_{1\mu}^{(t)}\lb y_{\mu}-\sum_{j}x_{\mu j} \overline{\beta_{j}}^{(t)} +\chi_{\mu}^{(t)}a_{\mu}^{(t-1)} \rb,
\Leq{AMPR-a}
\\ &&
A_{i}^{(t+1)}=\sum_{\mu}x_{\mu i}^2 f_{1\mu}^{(t)},
\\ &&
B_{i}^{(t+1)}=\sum_{\mu} x_{\mu i} a_{\mu}^{(t)}
+\lb \sum_{\mu}x_{\mu i}^2 f_{1\mu}^{(t)} \rb \overline{ \beta_{i} }^{(t)},
\\ &&
C_{i}^{(t+1)}=\sum_{\mu}x_{\mu i}^2 \lbb 
  f_{2\mu}^{(t)}W_{\mu}^{(t)}
  +
  \lb f_{2\mu}^{(t)}- \lb f_{1\mu}^{(t)} \rb^2 \rb \lb \frac{a_{\mu}^{(t)}}{f_{1\mu}^{(t)}} \rb^2 
\rbb 
,
\\ &&
\overline{\beta_i}^{(t+1)}=\lsb \int Dz S_{\lambda}\lb B^{(t+1)}_i+\sqrt{C^{(t+1)}_i}z;A^{(t+1)}_i \rb \rsb_{\lambda},
\\ &&
\chi^{(t+1)}_i
=\frac{1}{A^{(t+1)}_i}\lsb \int Dz~\theta \lb \left| B^{(t+1)}_i+\sqrt{C^{(t+1)}_i}z \right|-\lambda \rb \rsb_{\lambda},
\Leq{AMPR-chi}
\\ &&
W_i^{(t+1)}
=\lsb \int Dz S^2_{\lambda}\lb B^{(t+1)}_i+\sqrt{C^{(t+1)}_i}z;A^{(t+1)}_i \rb \rsb_{\lambda}
-
\lb \overline{\beta_i}^{(t+1)}\rb^2.
\ee
\subee
We call the algorithm \NReq{AMPR} AMPR (Approximate Message Passing with Resampling) because it can be regarded as an extension of the AMP in the usual Lasso to the resampling case. The computational cost is $O(NM)$ per iteration and is significantly reduced compared with the BP algorithm. This yields the main result of this study. 

There is an ambiguity in the initial condition for AMPR. Here we assume that we are given an initial estimate $\{ \overline{\beta_i}^{(0)},{\chi_i}^{(0)},{W_i}^{(0)}\}_i$ and conduct the iteration based on \Req{AMPR} until convergence. Still, there is an ambiguity in computing $a_{\mu}^{(0)}$, due to the presence of $a_{\mu}^{(-1)}$ in \Req{AMPR-a}. To resolve this, we assume $a_{\mu}^{(-1)}=0$, yielding
\be
a_{\mu}^{(0)}= \lsb\frac{c}{1+c\chi^{(0)}_{\mu}}\rsb_c \lb y_{\mu}-\sum_{j}x_{\mu j} \overline{\beta_{j}}^{(0)} \rb.
\ee
These completely determine the initial condition. 

As explained at the end of \Rsec{Gaussian approximation}, the convergent solution of AMPR, $\{A_i^{*},B_i^{*},C_i^{*}\}_i$, enables the computation of any moment of the basic estimator as follows:
\nbe
\lsb \Ave{\beta_{i}}^r \rsb_{\V{c},\V{\lambda}}
=
\lim_{n\to 0}\overline{\prod_{a=1}^{r}\beta^{a}_{i}}
=
\lsb \int Dz S^{r}_{\lambda}\lb B^*_i+\sqrt{C^*_i}z;A^*_i \rb \rsb_{\lambda},
\nee
which indicates that the marginal distribution of the basic estimator is obtained as
\nbe
P(\beta_i) =\lsb \int Dz~\delta\lb \beta_i- S_\lambda \lb B_i^*+\sqrt{C_i^*}z;A_i^*\rb \rb \rsb_{\lambda}.
\nee 
This yields the positive probability, which is important for variable selection techniques, as follows:
\be
\Pi_i\equiv \mathrm{Prob}\lb | \hat{\beta}_i| \neq 0 \rb
=\lsb \int Dz~\theta \lb \left| B^{*}_i+\sqrt{C^{*}_i}z \right|-\lambda \rb \rsb_{\lambda}.
\Leq{Pi}
\ee
Applications using these relations are provided in \Rsec{Numerical}.
\subsection{State evolution for AMPR} \Lsec{State evolution}
A benefit of the AMP type algorithms is that it is possible to track the macroscopic dynamical behavior of the algorithm. This can be done by using the so-called state evolution (SE) equations. Here we derive the SE equations associated with AMPR. The derivation relies on the  i.i.d. assumption of the covariates, and hence we assume each $x_{\mu i}$ is i.i.d. from the zero-mean Gaussian as
\be
x_{\mu i} \sim \mc{N}(0,\sigma_x^2).
\Leq{iidness}
\ee
The variance value is arbitrary in general, but for notational simplicity it is chosen as $\sigma_{x}^2=1/N$ in this section. Furthermore, we assume the data is generated from the following linear process:
\nbe
\V{y}=X\V{\beta}_0+\V{\xi},
\nee
where $\V{\xi}$ is the noise vector whose component is i.i.d. from $\mathcal{N}(0,\sigma_{\xi}^2)$ and $\V{\beta}_0$ is the true parameters whose component is also i.i.d. from a certain distribution $P_{\beta_{0}}(\cdot)$.

Under the assumption \NReq{iidness} with $\sigma_{x}^2=1/N$, the intra- and inter-sample variances can be simplified as
\be
&&
\chi_{\mu}=\sum_{i}x_{\mu i}^2 \chi_{i} 
\approx \sum_{i} \mathrm{E}\lsb  x_{\mu i}^2 \chi_{i} \rsb 
\approx
 \frac{1}{N}\sum_{i} \chi_{i}
 \equiv \tilde{\chi}
, \\ &&
W_{\mu}
=\sum_{i}x_{\mu i}^2 W_{i} 
\approx \sum_{i} \mathrm{E}\lsb  x_{\mu i}^2 W_{i} \rsb 
\approx \frac{1}{N}\sum_{i} W_{i} 
\equiv \tilde{W},
\ee
where we have neglected the correlations between $x_{\mu i}$ and the variances\footnote{
Remembering the discussion in \Rsec{Self-consistent equations}, these variances are actually the ones computed in the absence of the $\mu$-th potential function, $\chi_{i}^{\bs \mu}$ and $W_i^{\bs \mu}$, and hence this neglect can be justified. The same discussion can be applied to the following.}. 
Accordingly, many quantities appearing in \Req{AMPR} become independent of the subscripts $\mu$ and $i$. Terms retaining the dependence are only the linear terms with respect to $x_{\mu i}$ such as $\overline{\beta}_i,B_i$ and $a_{\mu}$. To derive the SE equations, we need to handle those terms.

We start from the following form of $B^{(t+1)}_{i}$:
\be
&&
B_{i}^{(t+1)}
=
\sum_{\nu}
\lsb \frac{c}{1+c\chi_\nu} \rsb_{c}r^{(t)}_{\nu i}x_{\nu i}
\approx f_{1}^{(t)}
\sum_{\nu} x_{\nu i}
\lb y_{\nu}-\sum_{j (\neq i)}x_{\nu j}\lb \bar{\beta}^{\bs \nu}_{j} \rb^{(t)}\rb.
\ee
The righthand side is the sum of a large number of random variables, and hence we can treat it as a Gaussian variable with appropriate mean and variance. The mean is 
\be
&&
\mathrm{E}
\lsb 
f_{1}^{t} \sum_{\nu} x_{\nu i}
\lb y_{\nu}-\sum_{j (\neq i)}x_{\nu j}\lb \bar{\beta}^{\bs \nu}_{j} \rb^{(t)}\rb
\rsb
\no \\ && 
=
f_{1}^{t}
\sum_{\nu} 
\lb 
\mathrm{E}\lsb x^2_{\nu i}\rsb \beta_{0i}
+
\sum_{j (\neq i)}
\mathrm{E}\lsb x_{\nu i}x_{\nu j} \rsb \lb {\beta}_{0j} - \lb \bar{\beta}^{\bs \mu}_{j} \rb^{(t)}   \rb
+
\mathrm{E}\lsb x_{\nu i} \xi_{\nu} \rsb
\rb
\no \\ && 
=
f_{1}^{t}
\sum_{\nu} 
\frac{1}{N} \beta_{0i}
=
\alpha f_{1}^{t}\beta_{0i},
\Leq{E_derivation}
\ee
where $\alpha=M/N$ is the ratio of the dataset size to the dimensionality. In the same way, the variance becomes
\be
\mathrm{V}\lsb 
f_{1}^{t} \sum_{\nu} x_{\nu i}
\lb y_{\nu}-\sum_{j (\neq i)}x_{\nu i}\lb \bar{\beta}^{\bs \nu}_{j} \rb^{(t)}\rb
\rsb
\approx \alpha \lb f_{1}^{t} \rb^2 \lb \MSE^{(t)}+\sigma_{\xi}^2 \rb
\equiv 
v_0^{(t+1)}
,
\Leq{V_derivation}
\ee
where $\MSE^{(t)}$ denotes the mean-squared error (MSE) between the true and averaged parameters:
\be
\MSE^{(t)}\equiv 
\frac{1}{N}\sum_{i=1}^{N}\lb \beta_{0i}-\overline{\beta}_{i}^{(t)}\rb^2
\approx
\frac{1}{N}\sum_{i=1}^{N}\lb \beta_{0i}-\lb \overline{\beta}^{\bs \mu}_{i}\rb^{(t)} \rb^2.
\ee
Hence, we may write
\be
B_{i}^{(t+1)}=
\alpha f_{1}^{t}\beta_{0i}+\sqrt{v_0^{(t+1)} }u_{i},
\Leq{SE-B}
\ee
where $u_{i}\sim \mc{N}(0,1)$. Besides, in the computation of $C_{i}^{(t+1)}$, we have 
\be
&&
\sum_{\nu}x_{\nu i}^2 f_{2 \nu}^{(t)}W_{\nu}^{(t)}
\approx 
\alpha f_{2}^{(t)}\tilde{W}^{(t)},
\\ &&
\sum_{\nu}x_{\nu i}^2 \lb r_{\nu i}^{(t)}\rb^{2}
\approx 
\alpha \lb \MSE^{(t)}+\sigma_{\xi}^2 \rb.
\ee
This yields
\be
C^{(t+1)}_{i}
\approx
C^{(t+1)}
=
\alpha f_{2}^{(t)}\tilde{W}^{(t)}
+
\alpha \lb f_{2}^{(t)}- \lb f_{1}^{(t)} \rb^2 \rb \lb \MSE^{(t)}+\sigma_{\xi}^2 \rb.
\ee
In the same level of approximation, we get $A_{i}^{(t+1)} \approx A^{(t+1)}=\alpha f_{1}^{(t)}$. 

To derive a closed set of equations, we have to compute $\tilde{\chi}^{(t+1)},\tilde{W}^{(t+1)}$ and $\MSE^{(t+1)}$ from $\{ A^{(t+1)},\{ B^{(t+1)}_{i}\}_i,C^{(t+1)} \}$. As an example, we show the derivation of $\tilde{\chi}^{(t+1)}$ from \Req{AMPR-chi} in the following:
\be
&&
\hspace{-10mm}
\tilde{\chi}^{(t+1)}
=
\frac{1}{N} \sum_{i=1}^{N}\chi_i^{(t+1)}
\approx 
\frac{1}{N} \sum_{i=1}^{N}
\frac{1}{ A^{(t+1)} }
\lsb \int Dz~\theta \lb \left| B^{(t+1)}_i+\sqrt{C^{(t+1)}}z \right|-\lambda \rb
\rsb_{\lambda} 
\no \\ &&
\hspace{-10mm}
\approx 
\frac{1}{ A^{(t+1)} }
\int d\beta P_{\beta_{0}}( \beta )
\int Du
\lsb \int Dz~
\theta \lb \left|  
\alpha f_1^{(t)}\beta+\sqrt{v_0^{(t+1)}}u+ \sqrt{C^{(t+1)}}z \right|-\lambda \rb
\rsb_{\lambda},
\ee
where we have applied the law of large numbers. The other quantities $\tilde{W},\MSE$ are computed in the same manner. Overall, we reach the following set of equations:
\subbe
\Leq{SE}
\be
&&
\lb f^{(t)}_{1}, f^{(t)}_{2}\rb= 
\lb
 \lsb\frac{c}{1+c\tilde{\chi}^{(t)}}\rsb_c, 
 \lsb\lb \frac{c}{1+c\tilde{\chi}^{(t)}} \rb^2\rsb_c
\rb,
\\ &&
A^{(t+1)}=\alpha f_{1}^{(t)},
\\ &&
C^{(t+1)}=\alpha f_{2}^{(t)}\tilde{W}^{(t)}+\alpha \lb f_{2}^{(t)}- \lb f_{1}^{(t)} \rb^2 \rb \lb \MSE^{(t)}+\sigma_{\xi}^2 \rb,
\\ &&
v_{0}^{(t+1)}=\alpha \lb f_{1}^{(t)} \rb^2 \lb \MSE^{(t)}+\sigma_{\xi}^2 \rb,
\\ &&
\tilde{\chi}^{(t+1)}
=\frac{1}{ A^{(t+1)} }
\int d\beta P_{\beta_{0}}( \beta )
\int Du
\no \\ && \hspace{5mm} \times 
\lsb \int Dz~
\theta \lb \left|  
A^{(t+1)}\beta+\sqrt{v_0^{(t+1)}}u+ \sqrt{C^{(t+1)}}z \right|-\lambda \rb
\rsb_{\lambda},
\\ &&
\tilde{W}^{(t+1)}
=
\int d\beta P_{\beta_{0}}( \beta )
\int Du \Biggl\{
\lsb \int Dz S^2_{\lambda}\lb A^{(t+1)}\beta+\sqrt{v_0^{(t+1)}}u+\sqrt{C^{(t+1)}}z;A^{(t+1)} \rb \rsb_{\lambda}
\no \\ && \hspace{5mm}
-\lsb \int Dz S_{\lambda}\lb A^{(t+1)}\beta+\sqrt{v_0^{(t+1)}}u+\sqrt{C^{(t+1)}}z;A^{(t+1)} \rb \rsb_{\lambda}^2
\Biggr\}
,
\\ &&
\MSE^{(t+1)}
=\int d\beta P_{\beta_{0}}( \beta )
\int Du 
\no \\ && \hspace{5mm} \times
\Biggl\{
\beta-
\lsb \int Dz S_{\lambda}\lb A^{(t+1)}\beta+\sqrt{v_0^{(t+1)}}u+\sqrt{C^{(t+1)}}z;A^{(t+1)} \rb \rsb_{\lambda}
\Biggr\}^2.
\ee
\subee
Given an initial condition $\{ \tilde{\chi}^{(0)},\tilde{W}^{(0)},\MSE^{(0)} \}$, we can track the dynamical evolution of those quantities according to \Req{SE}. This is the SE equations for AMPR.

A direct consequence of the SE equations is the convergence property of AMPR: Its convergence depends on neither the dataset size $M$ nor the model dimensionality $N$. Hence, we can assume the iteration steps required for convergence is $O(1)$ and the total computational cost of AMPR is thus guaranteed to be $O(NM)$. This reinforces the superiority of the present approach.

We, however, warn that the discussion based on the SE equations strongly relies on the i.i.d. assumption among covariates, which is not necessarily satisfied in real-world datasets. For covariates with non-trivial correlations and heterogeneity, it is known that AMP type algorithms tend to show slow convergence, or even not to converge in particular cases~\citep{caltagirone2014convergence}. A common prescription to overcome this difficulty is to introduce a damping factor in the update of the messages~\citep{rangan2014convergence}, which is also employed in our implementation~\citep{obuchi:AMPR}. In \Rsecs{Correlated covariates}{Real-world dataset}, we see how this prescription works for datasets with nontrivial covariates in numerical simulations.

\section{Numerical experiments}\Lsec{Numerical}
In this section, the accuracy and the computational time of the proposed semi-analytic method based on AMPR is examined by a comparison with direct numerical resampling. We also check how nontrivial correlations among covariates affect the performance of AMPR. Both simulated and real-world datasets~\citep[from UCI machine learning repository, ][]{Lichman:13} are used. 

For all experiments involving numerical resampling, {\it Glmnet}~\citep{friedman2010regularization}, implemented as an {\it MEX} subroutine in MATLAB\textsuperscript{\textregistered}, was employed for solving \Req{Lasso-random}, given a sample $\{\V{\lambda},D_{\V{c}}\}$. Moreover, the proposed AMPR algorithm was implemented as raw code in MATLAB. This is not the most optimized approach because AMPR uses a number of {\it for} and {\it while} loops which are slow in MATLAB; hence, the comparison of computational time is not necessarily fair. However, even in this comparison, there is a significant difference in the computational time between the proposed semi-analytic method and the numerical resampling approach. For reference, it should be noted that all experiments below were conducted in a single thread on a single CPU of Intel(R) Xeon(R) E5-2630 v3 2.4GHz.  

For actual computations, the distribution $P(\V{\lambda})$ should be specified. In SS, the following distribution is used~\citep{meinshausen2010stability}:
\nbe
P(\V{\lambda})=\prod_{i=1}^{N}\lbb p_{w}\delta \lb \lambda_i-\lambda/w \rb+ (1-p_{w})\delta \lb \lambda_i-\lambda \rb  \rbb,
\nee
with $0 <w \leq 1$ and $0 < p_w < 1$. The case of the non-random penalty coefficient, in which Bolasso is included~\citep{bach2008bolasso}, is recovered at $w=1$, irrespective of the value of $p_w$. This distribution is adopted below.

\subsection{Simulated dataset}\Lsec{On simulated}
Here, simulated datasets are treated.  The data is supposed to be generated from the following linear model:
\nbe
\V{y}=X\V{\beta}_0+\V{\xi},
\nee
where each component of the design matrix $X=(\V{x}_1,\V{x}_2,\cdots,\V{x}_{N})$ is i.i.d. from $\mathcal{N}\lb 0,N^{-1} \rb$, and $\V{\xi}$ is the noise vector, whose component is i.i.d. from $\mathcal{N}(0,\sigma_{\xi}^2)$. The ratio of the dataset size $M$ to the model dimensionality $N$ is denoted as $\alpha \equiv M/N$ hereafter. These settings are identical to the ones assumed in \Rsec{State evolution}. The true signal $\V{\beta}_{0}\in \mR^{N}$ is assumed to be $K_{0}(=N\rho_{0})$-sparse vector, and the non-zero components are i.i.d. from $\mathcal{N}(0,1/\rho_0)$, setting the power of the signal unity. The index set of non-zero components is denoted by $S_0=\{i||\beta_{0i}|\neq 0 \}$ and is called true support. Any estimator of the true support is simply called support and is hereafter denoted by $S$.

For simplicity, the number of resampling times is fixed at $N_{\rm res}=1000$ in the experiments with numerical resampling. 

\subsubsection{Accuracy of the semi-analytic method}\Lsec{Accuracy of the}
Let us first check the consistency between the results of our semi-analytic method and of the direct numerical resampling. 

\Rfig{yyplot1} shows the plots of the experimental values of the quantities $\{\overline{\beta_{i}},W_i,\Pi_i\}_i$ against their semi-analytic counterparts for the non-random penalty case $w=1$ with the bootstrap resampling $\tau=1$, which is the situation considered in Bolasso. The same plots in the SS situation, $w=0.5(<1)$, $p_w=0.5$, and $\tau=0.5$, are shown in \Rfig{yyplot2}. Other detailed parameters are provided in the captions. 
\begin{figure}[htbp]
\begin{center}
 \includegraphics[width=0.32\columnwidth]{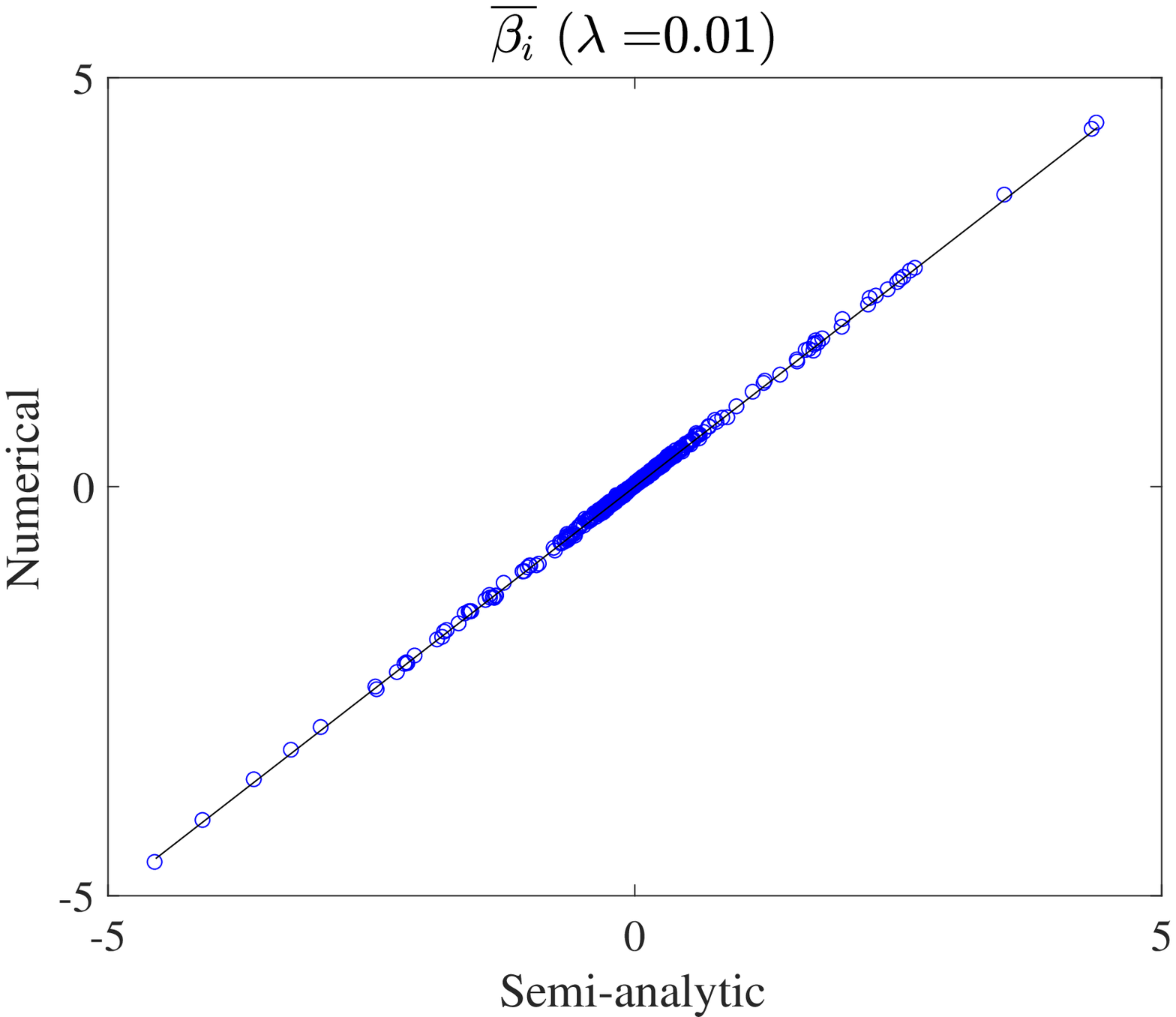}
 \vspace{1mm}
 \includegraphics[width=0.32\columnwidth]{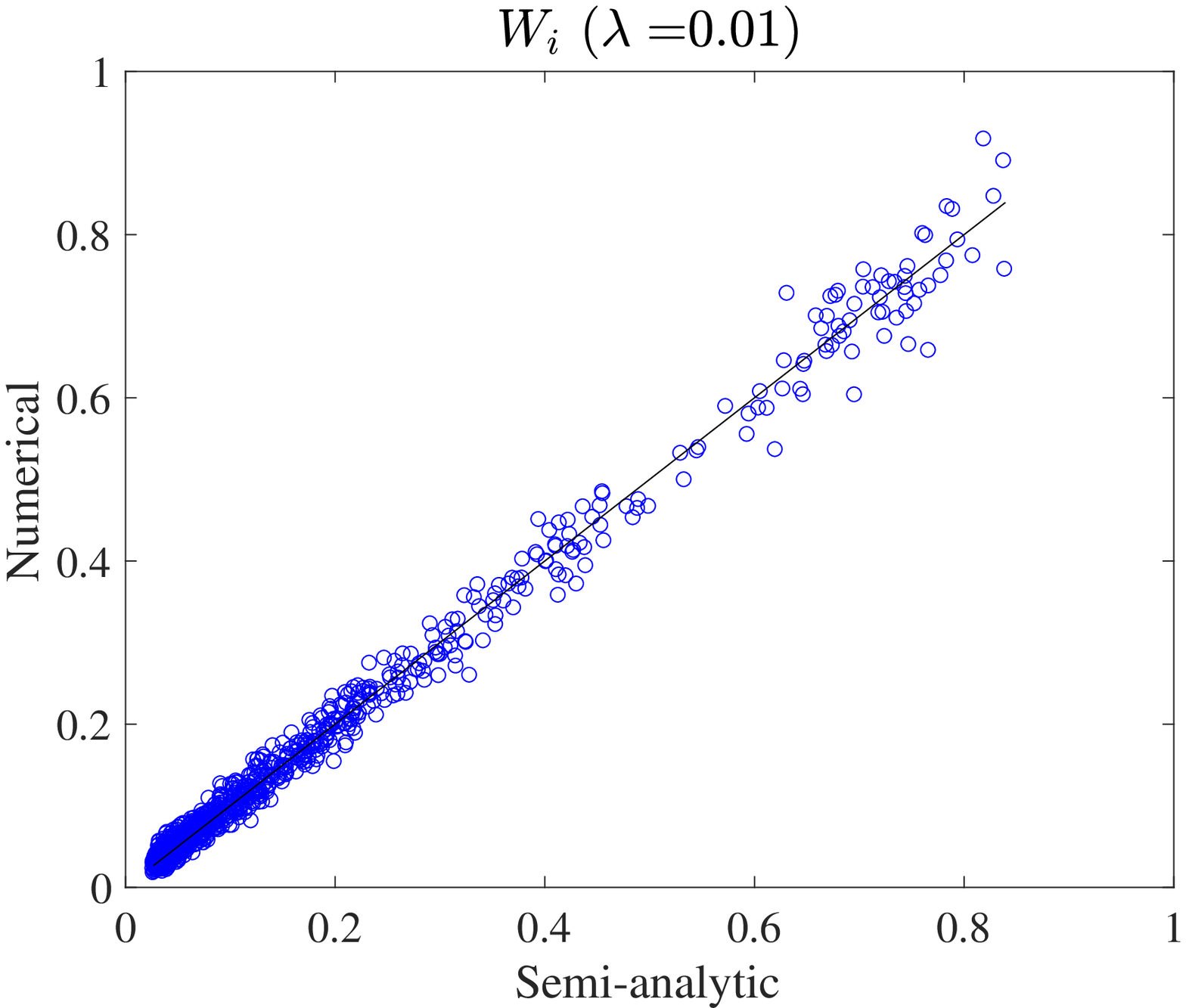}
 \vspace{1mm}
 \includegraphics[width=0.32\columnwidth]{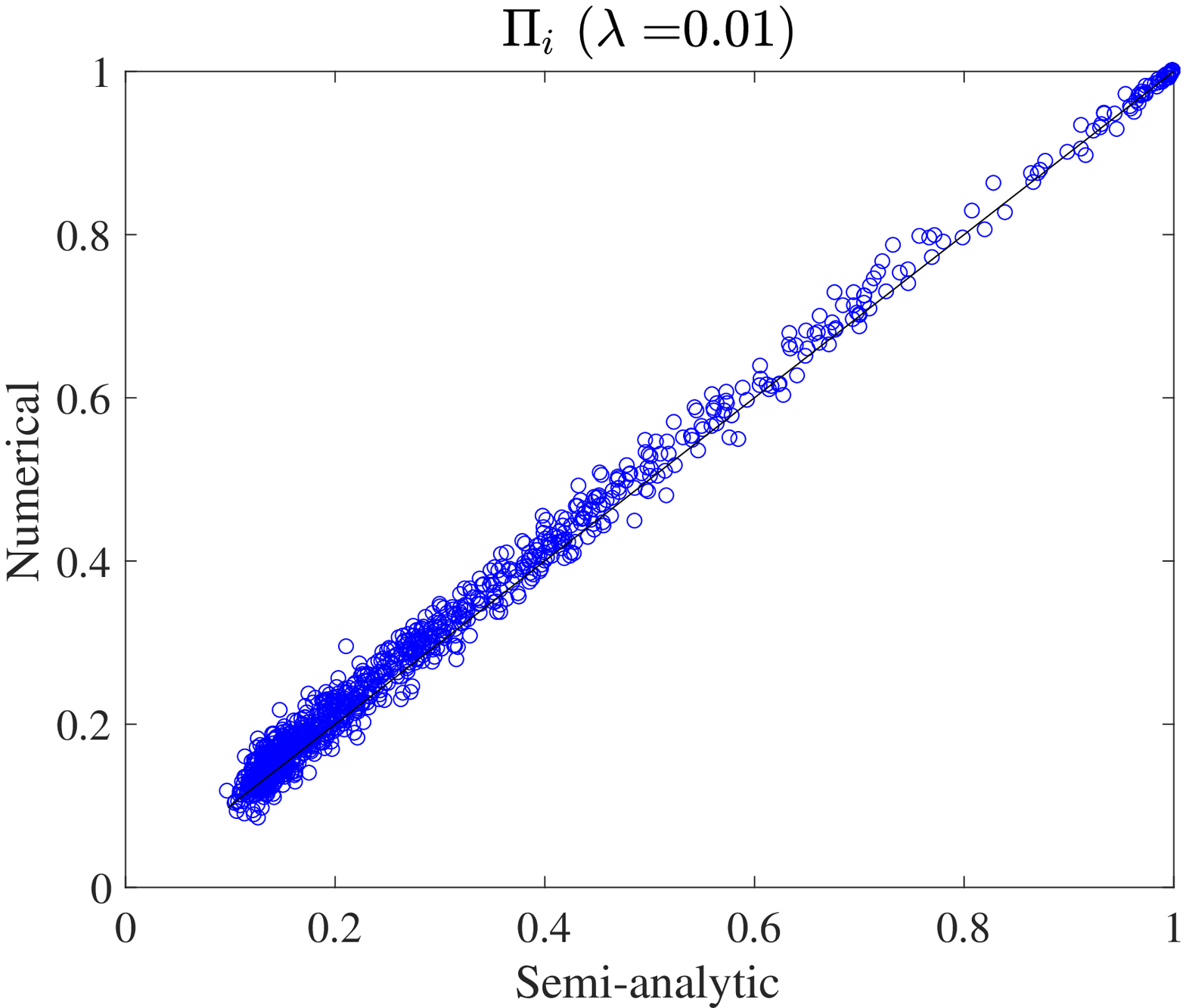}
 \vspace{1mm}
 \includegraphics[width=0.32\columnwidth]{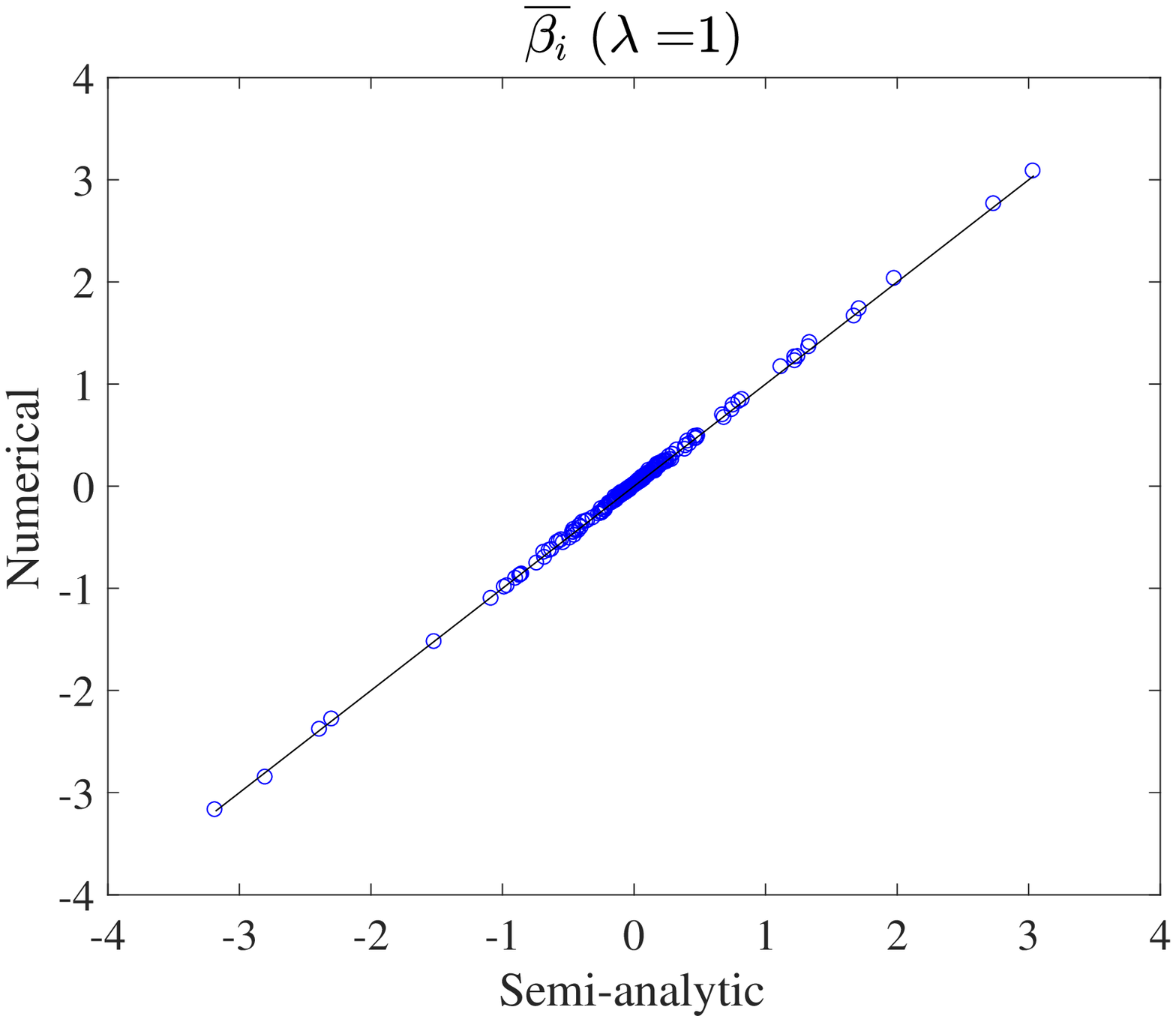}
 \includegraphics[width=0.32\columnwidth]{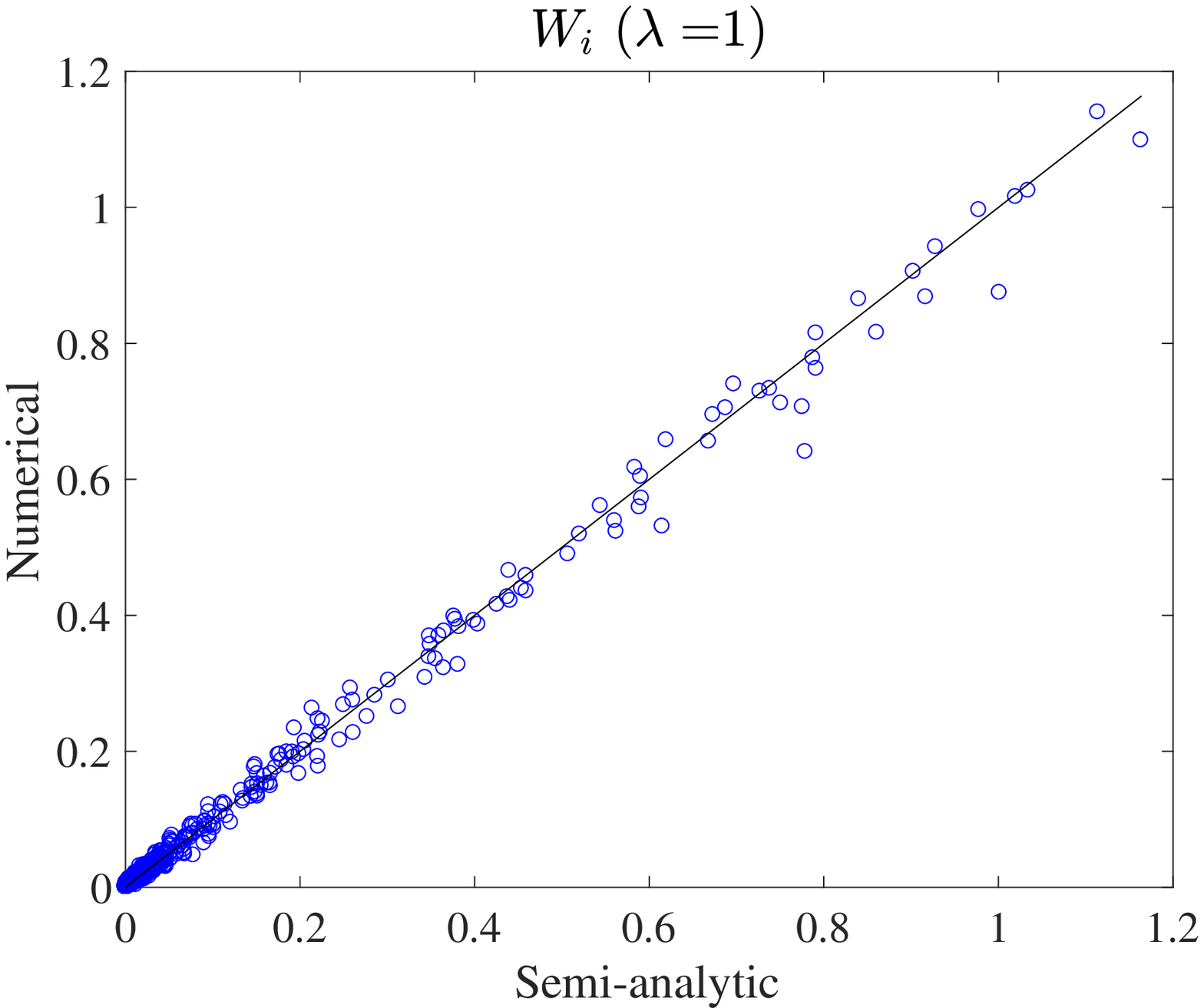}
 \includegraphics[width=0.32\columnwidth]{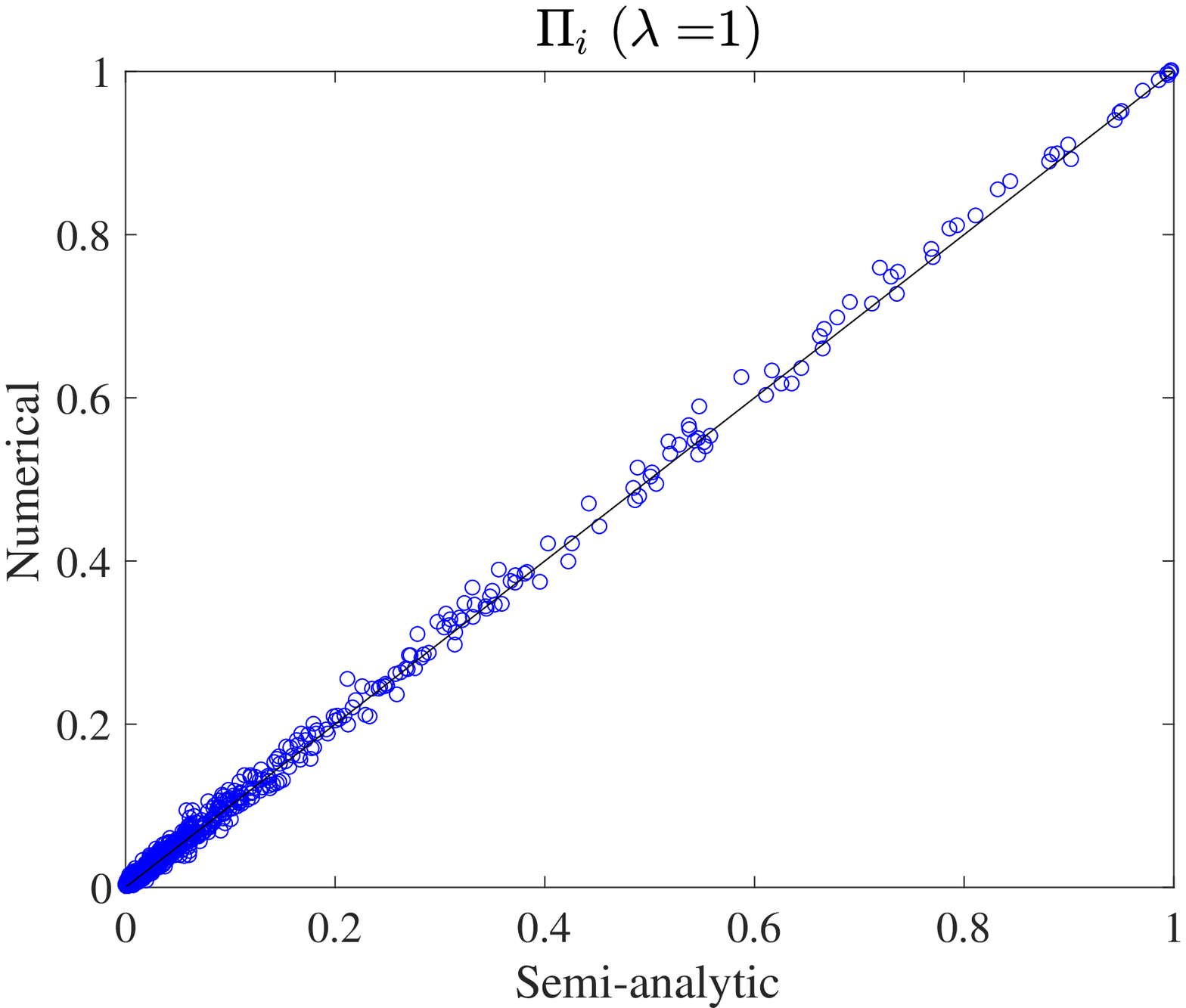}
\caption{Experimental values of $\overline{\beta_i}$ (left), $W_{i}$ (middle), and $\Pi_{i}$ (right) are plotted against those computed by the semi-analytic method in the non-random penalty $w=1$ and $\tau=1$ case. The upper panels are for $\lambda=0.01$ and the lower are for $\lambda=1$. The other parameters are set to be $(N,\alpha,\rho_{0},\sigma_{\xi}^2)=(1000,0.5,0.2,0.01)$. }
\Lfig{yyplot1}
\end{center}
\end{figure}
\begin{figure}[htbp]
\begin{center}
 \includegraphics[width=0.32\columnwidth]{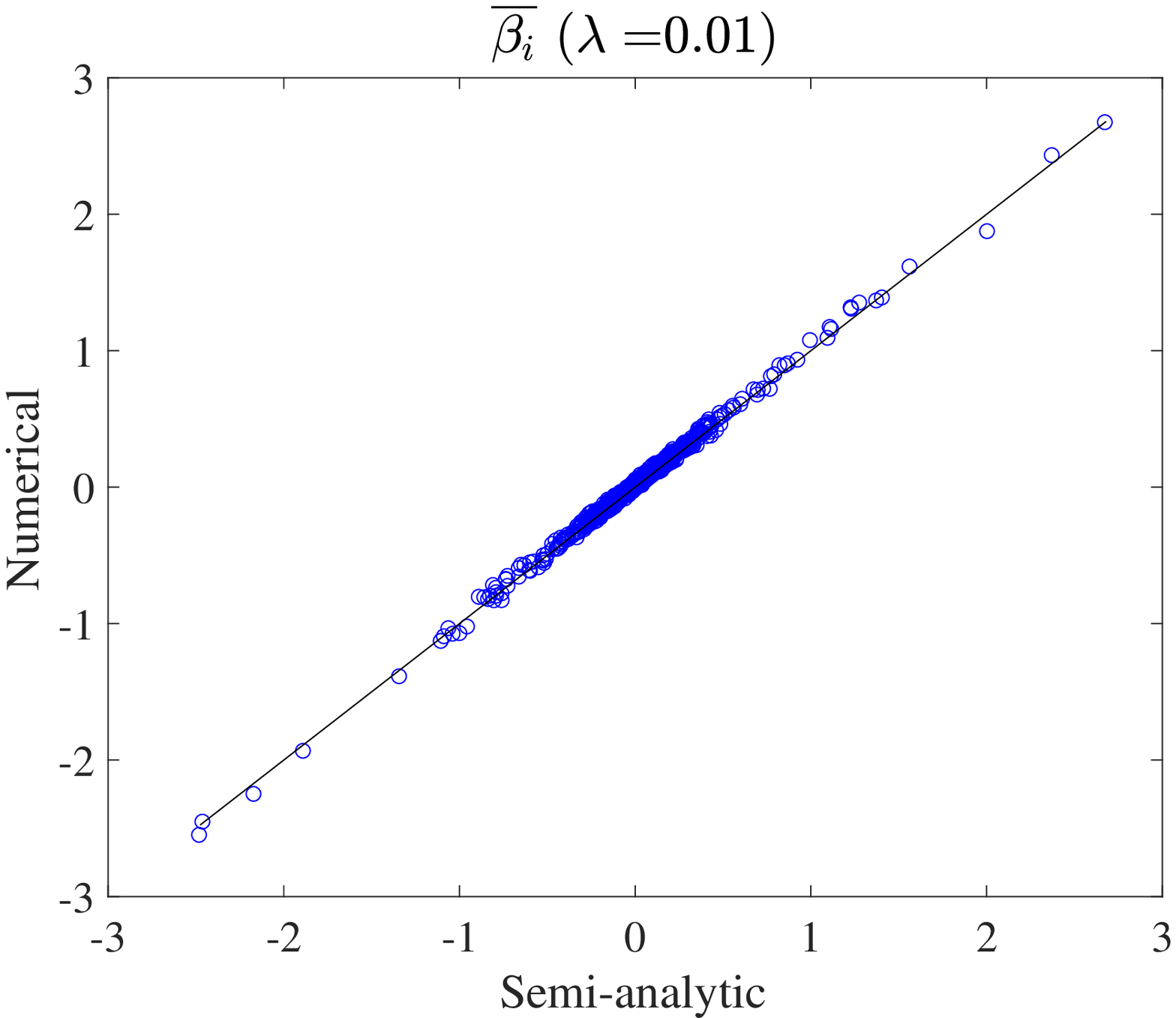}
 \vspace{1mm}
 \includegraphics[width=0.32\columnwidth]{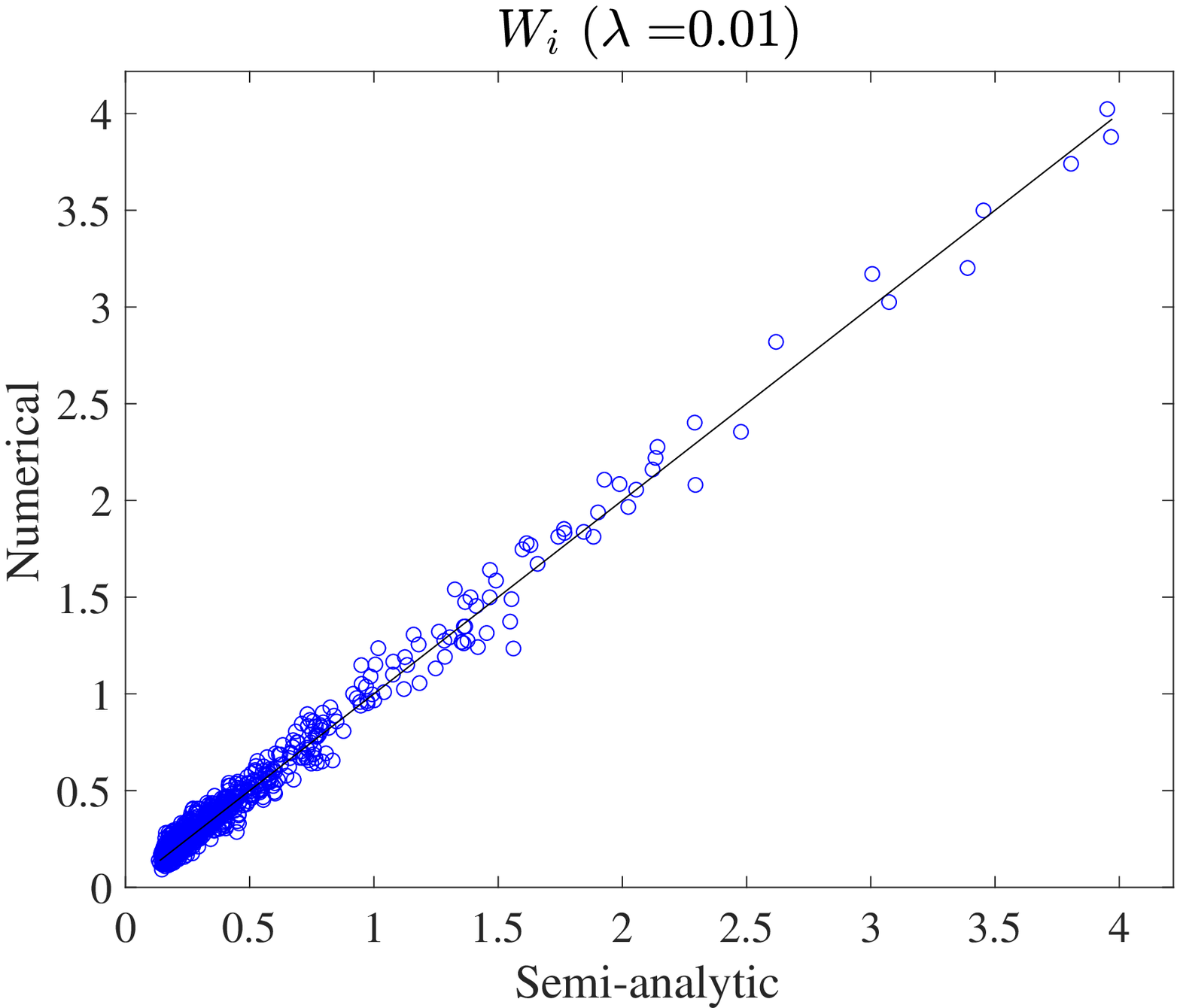}
 \vspace{1mm}
 \includegraphics[width=0.32\columnwidth]{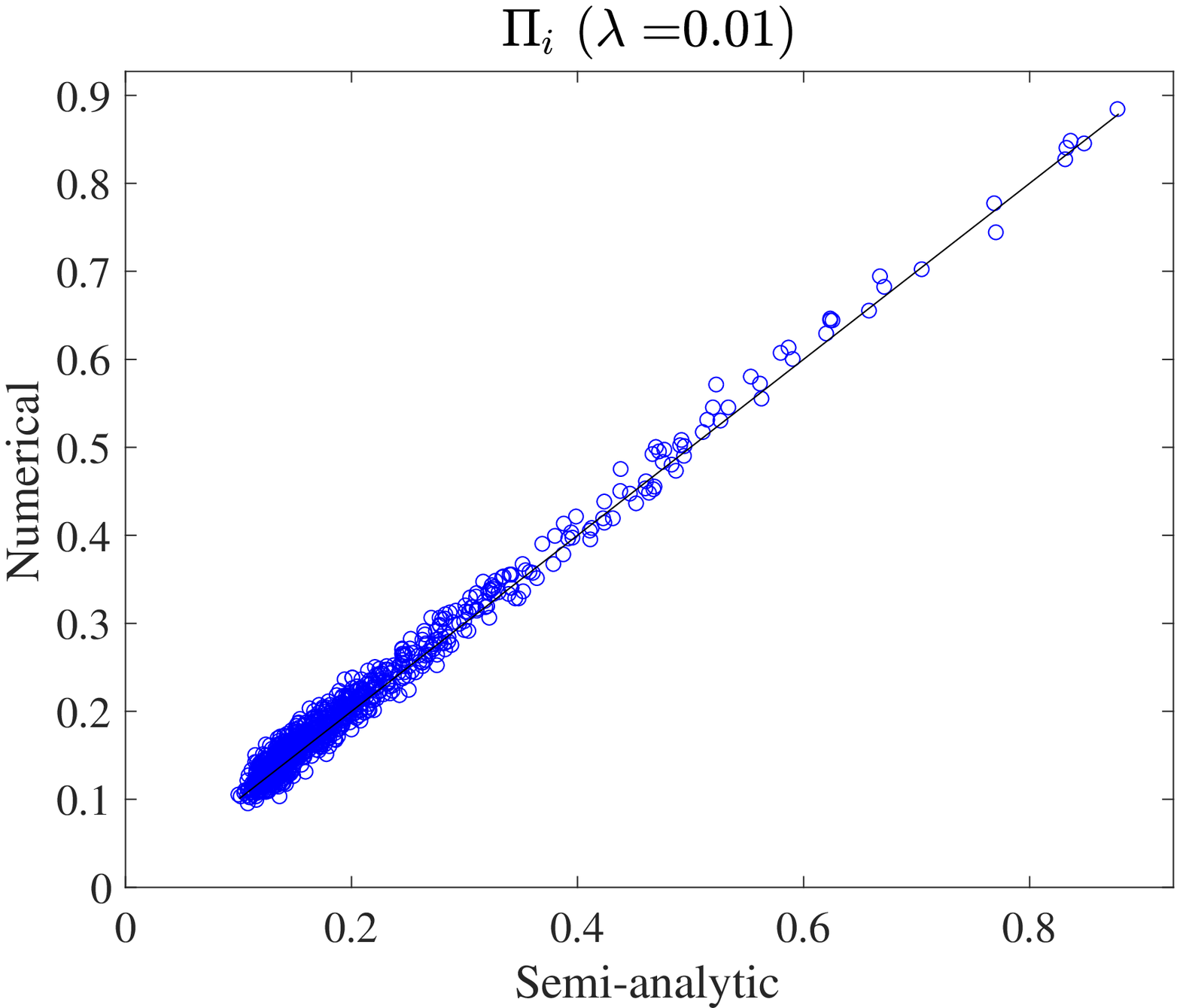}
 \vspace{1mm}
 \includegraphics[width=0.32\columnwidth]{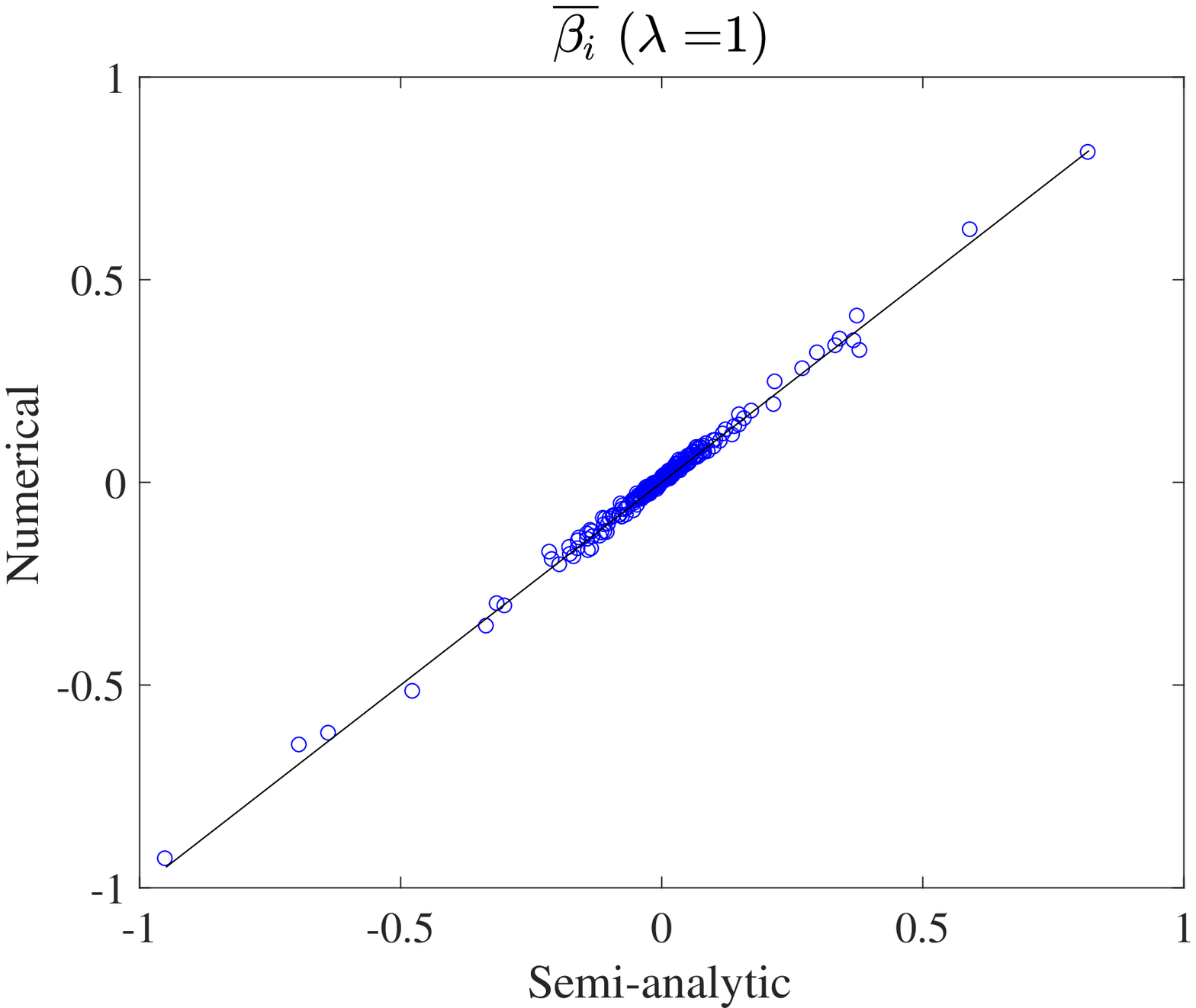}
 \includegraphics[width=0.32\columnwidth]{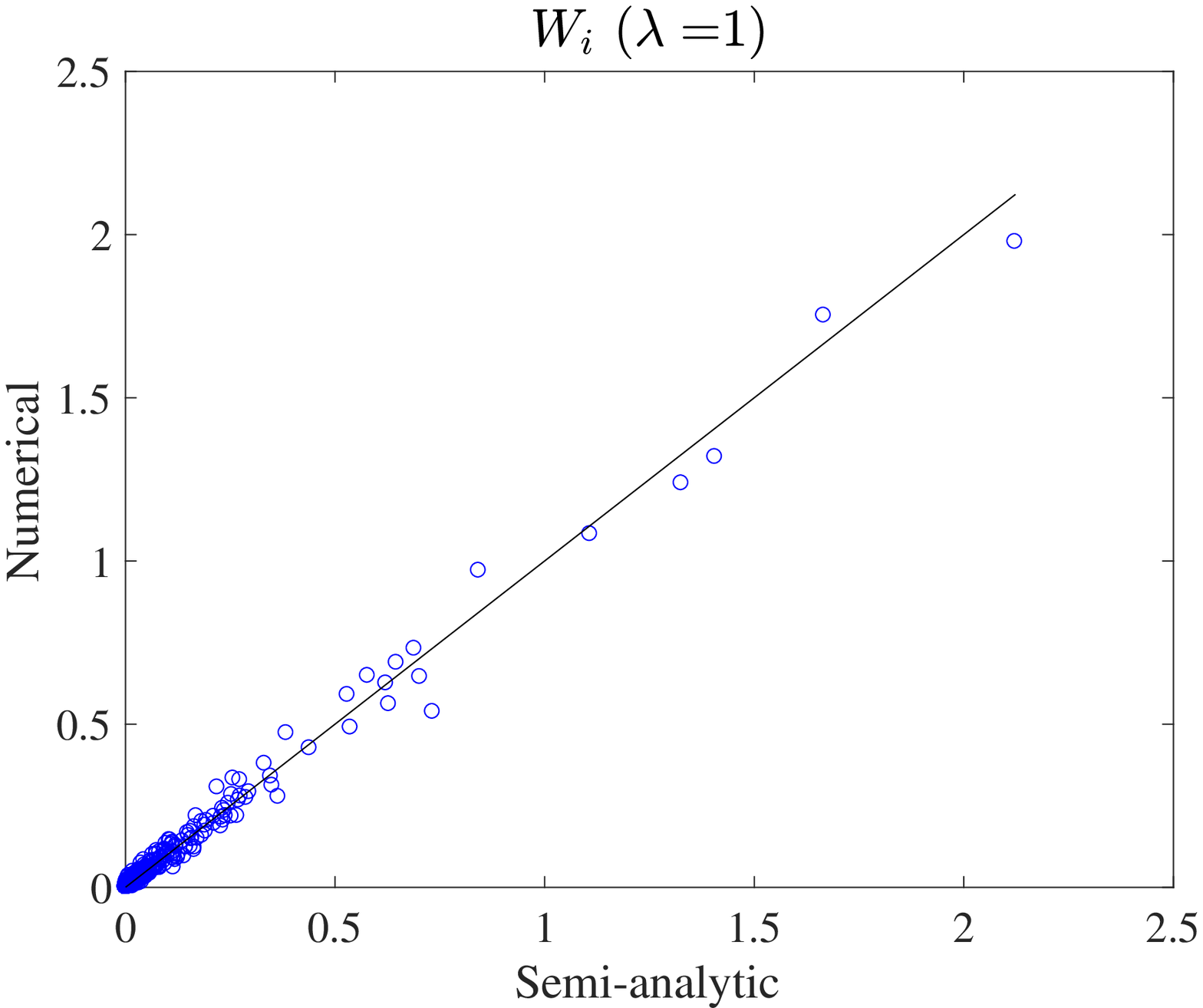}
 \includegraphics[width=0.32\columnwidth]{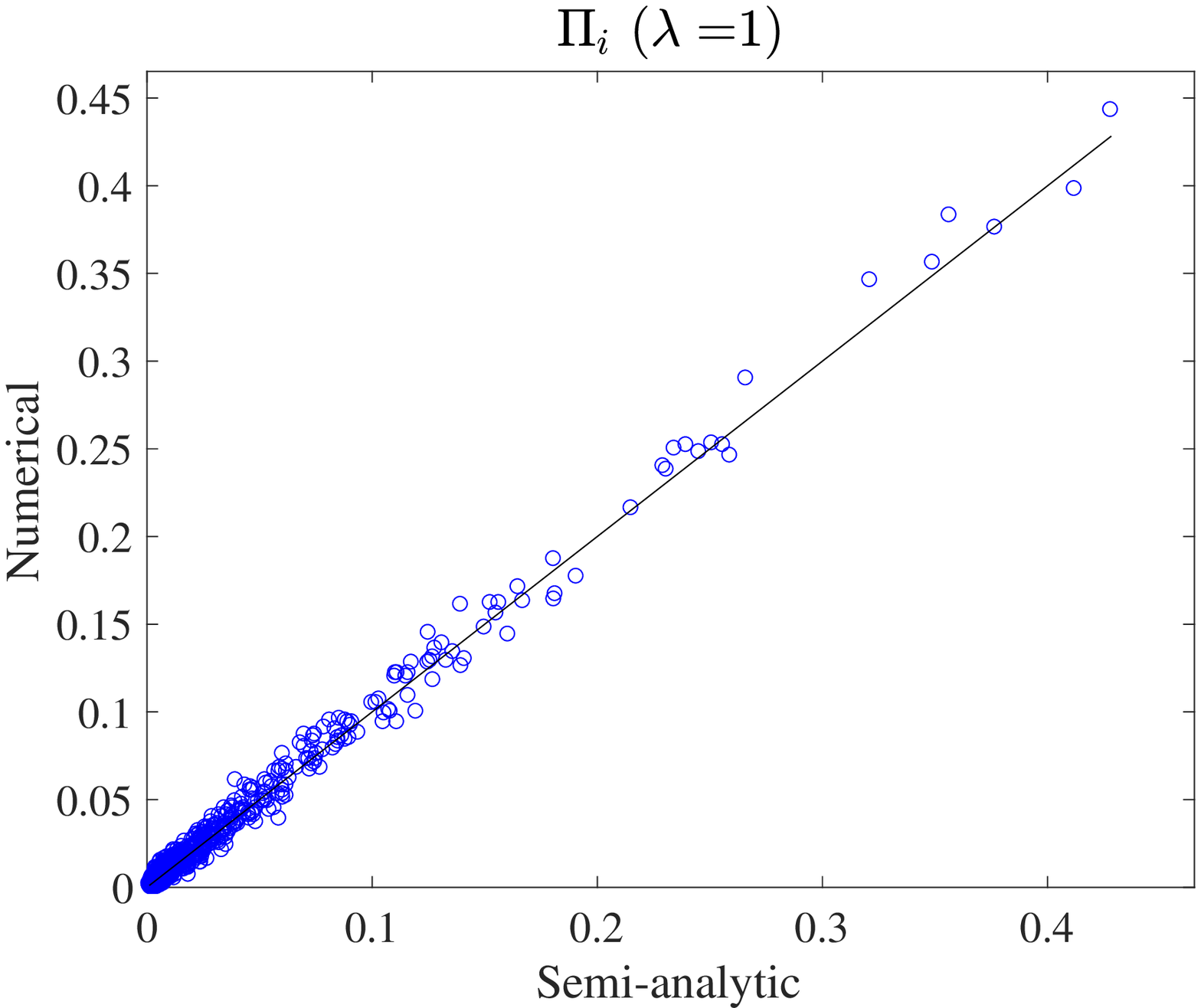}
\caption{Same plots as in \Rfig{yyplot1} in the random penalty case with $w=0.5$, $p_w=0.5$, and $\tau=0.5$. The parameters of each panel are identical to the corresponding parameters in \Rfig{yyplot1}. In comparison to \Rfig{yyplot1}, $|\overline{\beta}_i|$ and $\Pi_{i}$ tend to be smaller, whereas $W_i$ tends to be larger. This is probably due to the additional stochastic variation coming from the randomization on the penalty coefficient and the difference between $\tau=1$ and $1/2$.}
\Lfig{yyplot2}
\end{center}
\end{figure}
These results show that the proposed semi-analytic method reproduces the numerical results fairly accurately. As far as it was examined, results of similar accuracy were obtained for a very wide range of parameters. These validate the proposed semi-analytic method.

In the above experiments using Glmnet, we set a threshold $\epsilon$ to judge the algorithm convergence as $\epsilon=10^{-10}$, which is rather tighter than the default value. This is necessary for examining consistency with the proposed semi-analytic method. For example, for $\lambda=0.01$ (the upper panels in \Rfigs{yyplot1}{yyplot2}), a systematic deviation from the proposed semi-analytic method ($\overline{\beta}_i$ tends to be underestimated) emerges at the default value $\epsilon=10^{-7}$. This implies that a rather tight threshold is required for microscopic quantities such as $\overline{\beta_{i}}$ and $W_i$. Unless explicitly mentioned, this value $\epsilon=10^{-10}$ is used below.

\subsubsection{Comparison with state evolution}\Lsec{Comparison with SE}
To examine the convergence properties of AMPR, we next see the dynamical behavior of the macroscopic quantities $(\tilde{\chi}^{(t)},\tilde{W}^{(t)},\MSE^{(t)})$ as the algorithm step $t$ proceeds, in comparison with the SE equations \NReq{SE}. \Rfig{SE} shows the plots of them against $t$ for different parameters.
\begin{figure}[htbp]
\begin{center}
 \includegraphics[width=0.45\columnwidth]{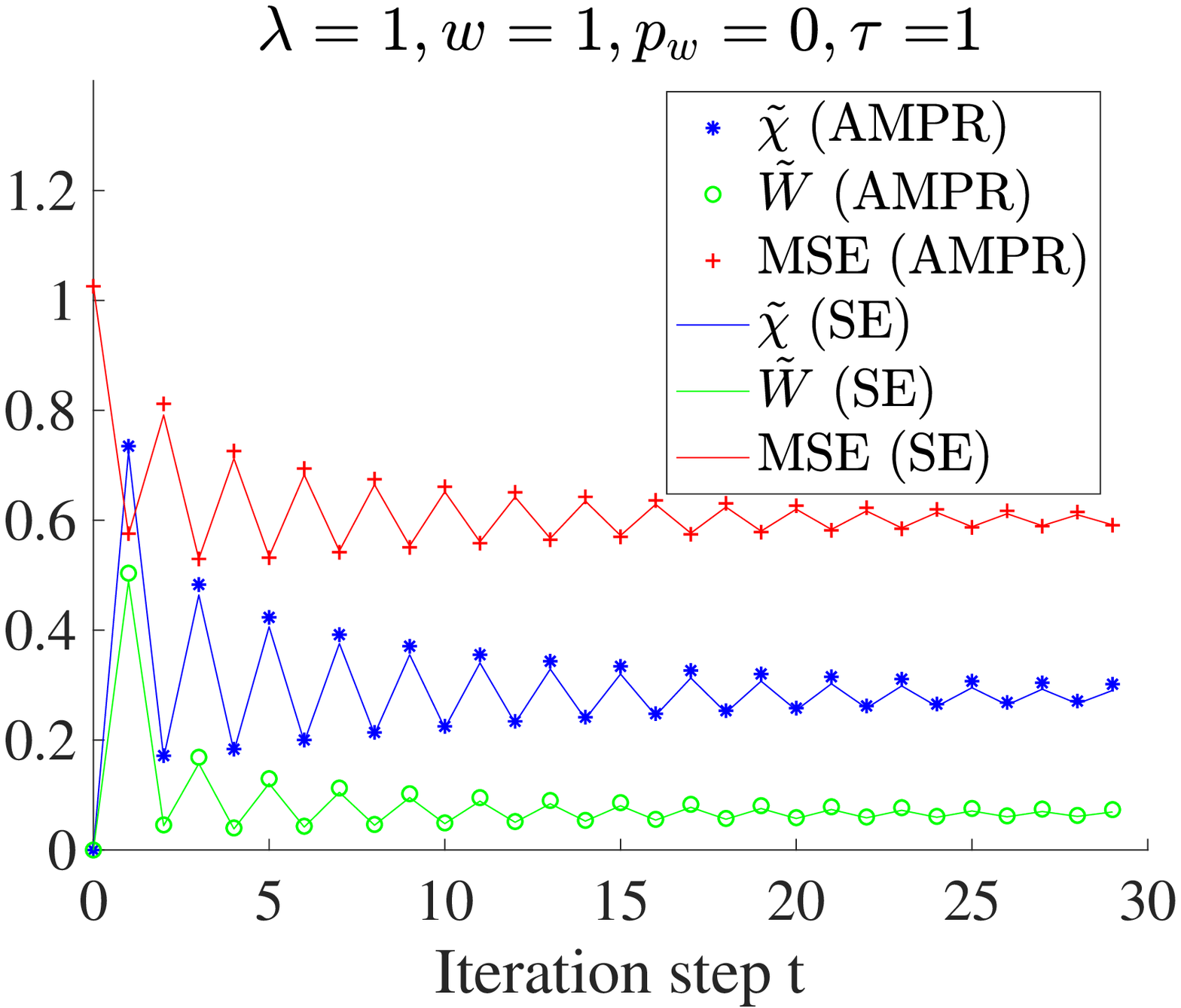}
 \vspace{1mm}
 \includegraphics[width=0.45\columnwidth]{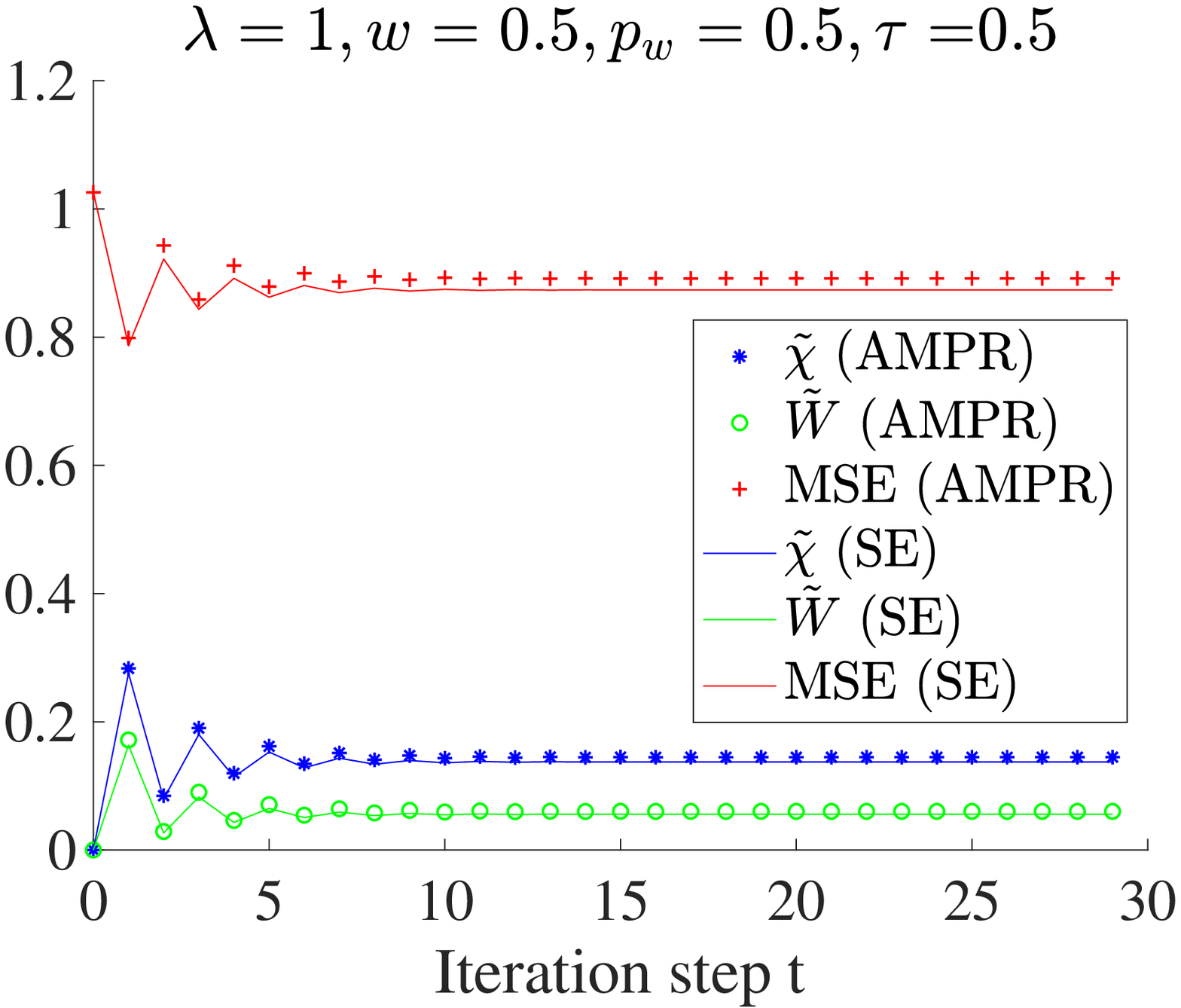}
 \vspace{1mm}
 \includegraphics[width=0.45\columnwidth]{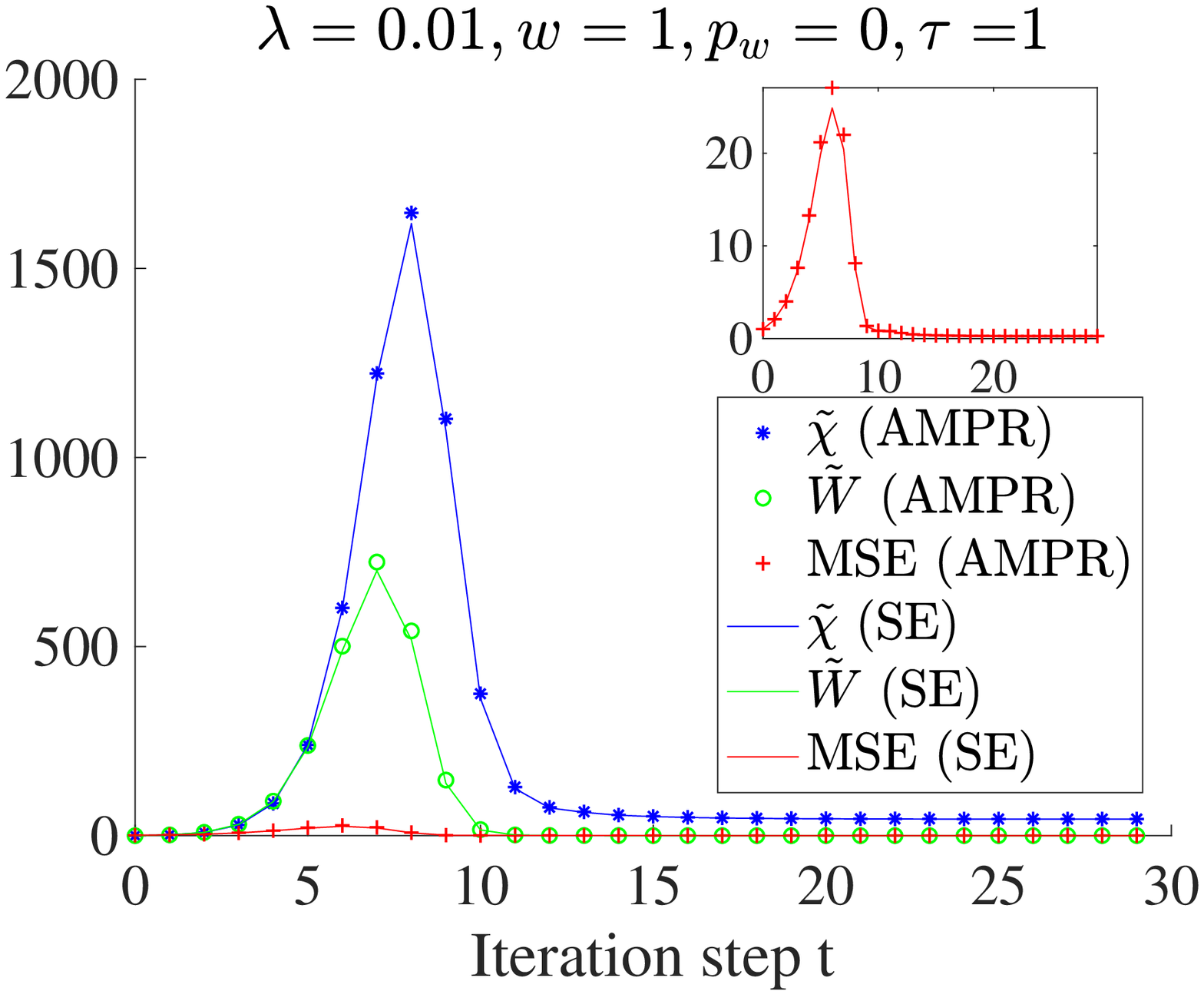}
 \includegraphics[width=0.45\columnwidth]{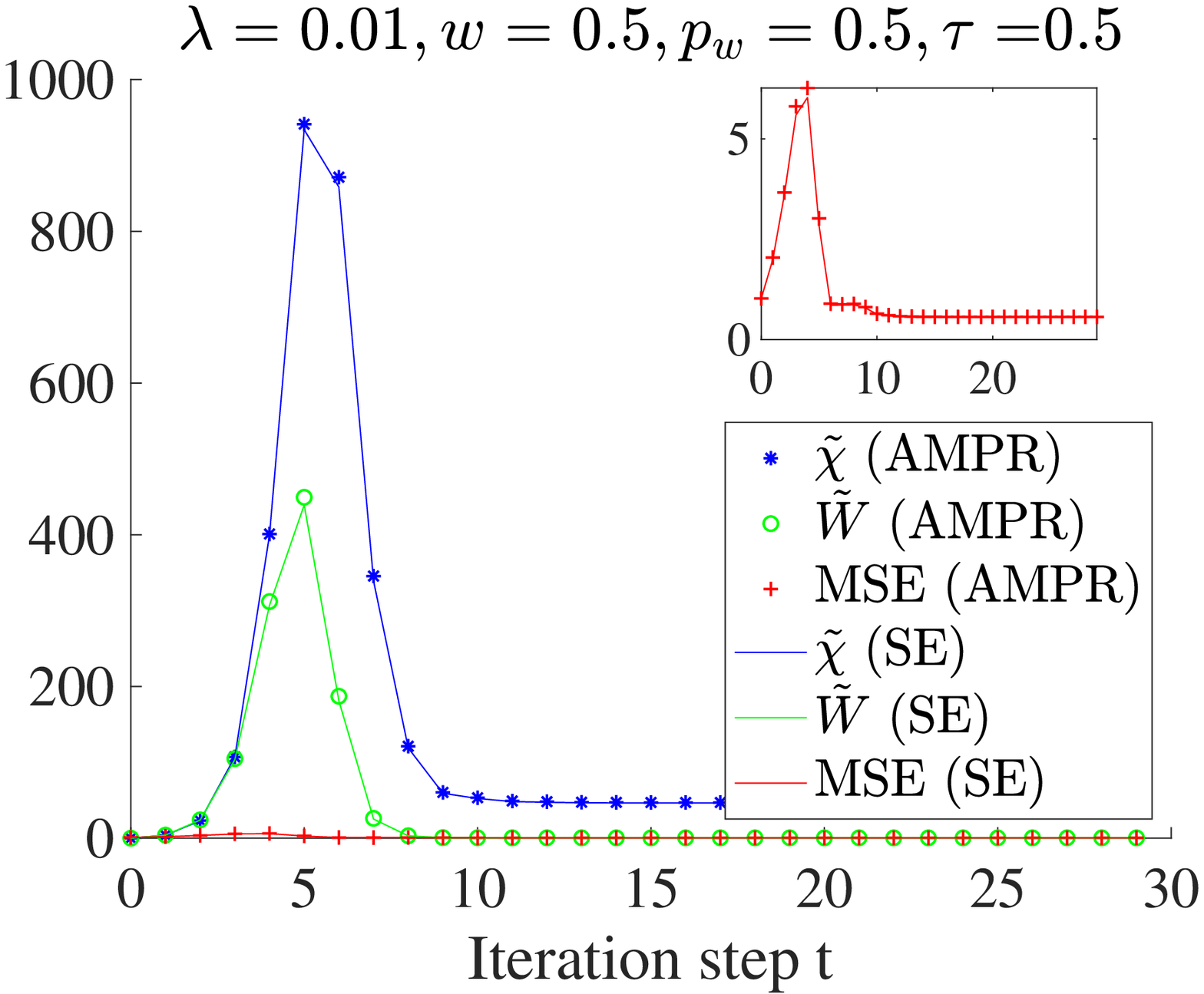}
\caption{Dynamical behavior of macroscopic parameters $(\tilde{\chi}^{(t)},\tilde{W}^{(t)},\MSE^{(t)})$ initialized at $(\tilde{\chi}^{(0)},\tilde{W}^{(0)})=(0,0)$ and $\MSE^{(0)}\approx 1$. The result computed by SE is denoted by lines while the one by AMPR is represented by markers, and the agreement is fairly good. (Left) The non-randomized penalty case ($w=1,p_w=0,\tau=1$). (Right) The randomized penalty case ($w=p_w=\tau=1/2$). The upper panels are for $\lambda=1$, while the lower ones are of $\lambda=0.01$ for which insets are given to show the MSE values in visible scales. The AMPR result is obtained at $N=20000$. The other parameters are fixed at $(\alpha,\rho_{0},\sigma_{\xi}^2)=(0.5,0.2,0.01)$.  
}
\Lfig{SE}
\end{center}
\end{figure}
In all the cases, these macroscopic quantities rapidly take stable values and the agreement between the AMPR and SE results is excellent. This demonstrates the fast convergence of AMPR, which does not depend on both $N$ and $M$. To see a good agreement with the SE result by just one sample of $\{\V{\beta}_0,X,\V{\xi}\}$, the model dimensionality $N$ is chosen as a rather large value of $N=20000$ in the AMPR experiment. Although in \Rfig{SE} the initial condition is fixed at $\V{\chi}^{(0)}=\V{W}^{(0)}=\overline{\V{\beta}}^{(0)}=\V{0}$ corresponding to $(\tilde{\chi}^{(0)},\tilde{W}^{(0)})=(0,0)$ and $\MSE^{(0)}\approx 1$, we also examined several other initial conditions and confirmed the good agreement between the AMPR and SE results in all the cases.

\subsubsection{Application to Bolasso}\Lsec{Application to Bolasso}
Bolasso is a variable selection method utilizing the positive probability~\NReq{Pi} evaluated by the bootstrap resampling $\tau=1$ with no penalty randomization $w=1$. Its soft version, abbreviated as Bolasso-S~\citep{bach2008bolasso}, selects variables with $\Pi_i \geq 0.9$ as active variables and other variables are rejected from the support. We here adopt this manner and see the performance of AMPR used for implementing this. 

The left panel of \Rfig{Bolasso} shows the plot of the true positive ratio (TP) against the false positive ratio (FP) as the values of $\alpha$ change. Bolasso requires scaling the regularization parameter as $\lambda \propto \sqrt{\alpha}$, and here, it is set to $\lambda=(1/2)\sqrt{\alpha}$. The other parameters are fixed at $(N,\rho_{0},\sigma_{\xi}^2)=(1000,0.2,0.01)$. As was theoretically shown~\citep{bach2008bolasso}, TP converges to unity, whereas FP tends to zero in this setup.  
\begin{figure}[htbp]
\begin{center}
 \includegraphics[width=0.45\columnwidth]{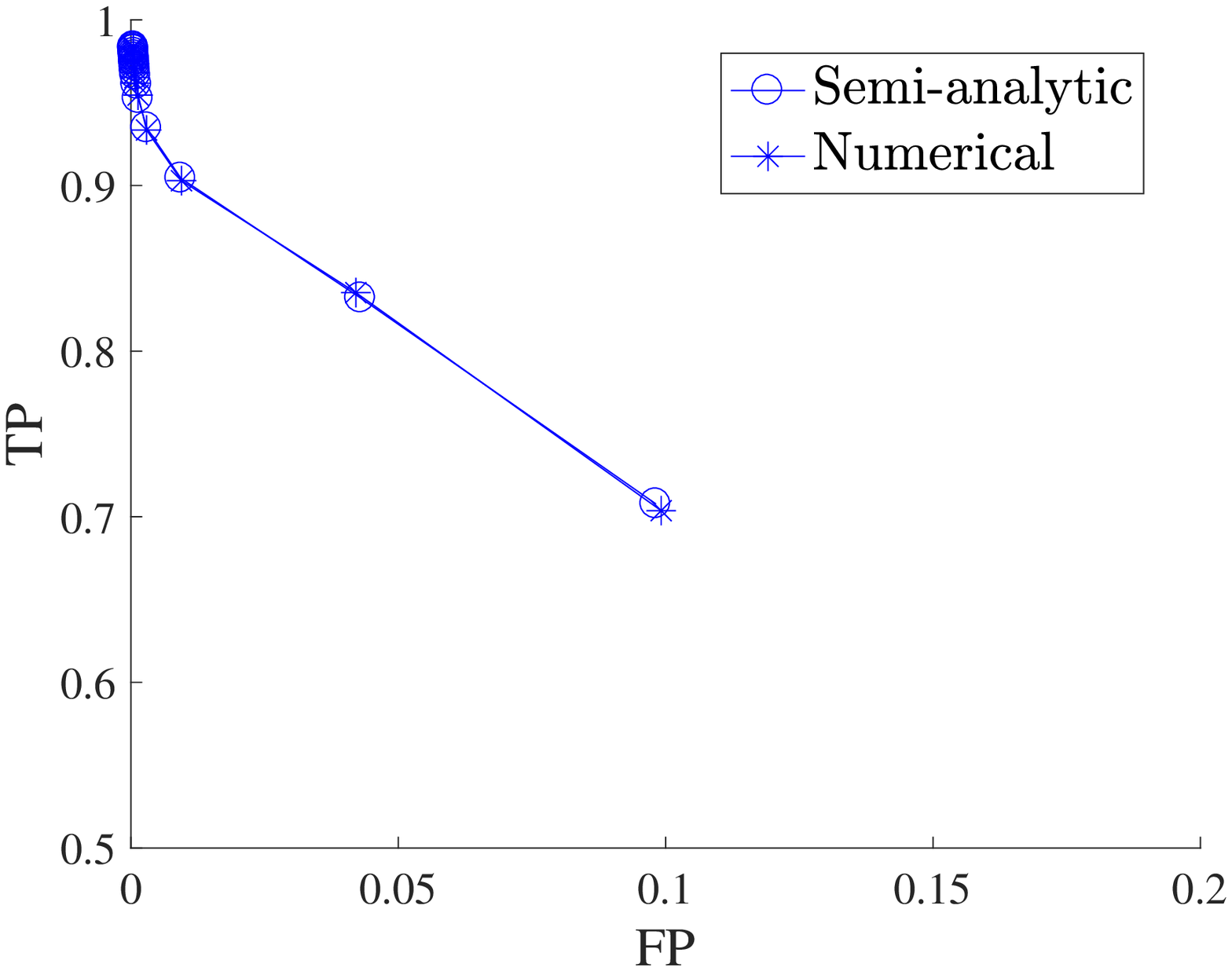}
 \includegraphics[width=0.45\columnwidth]{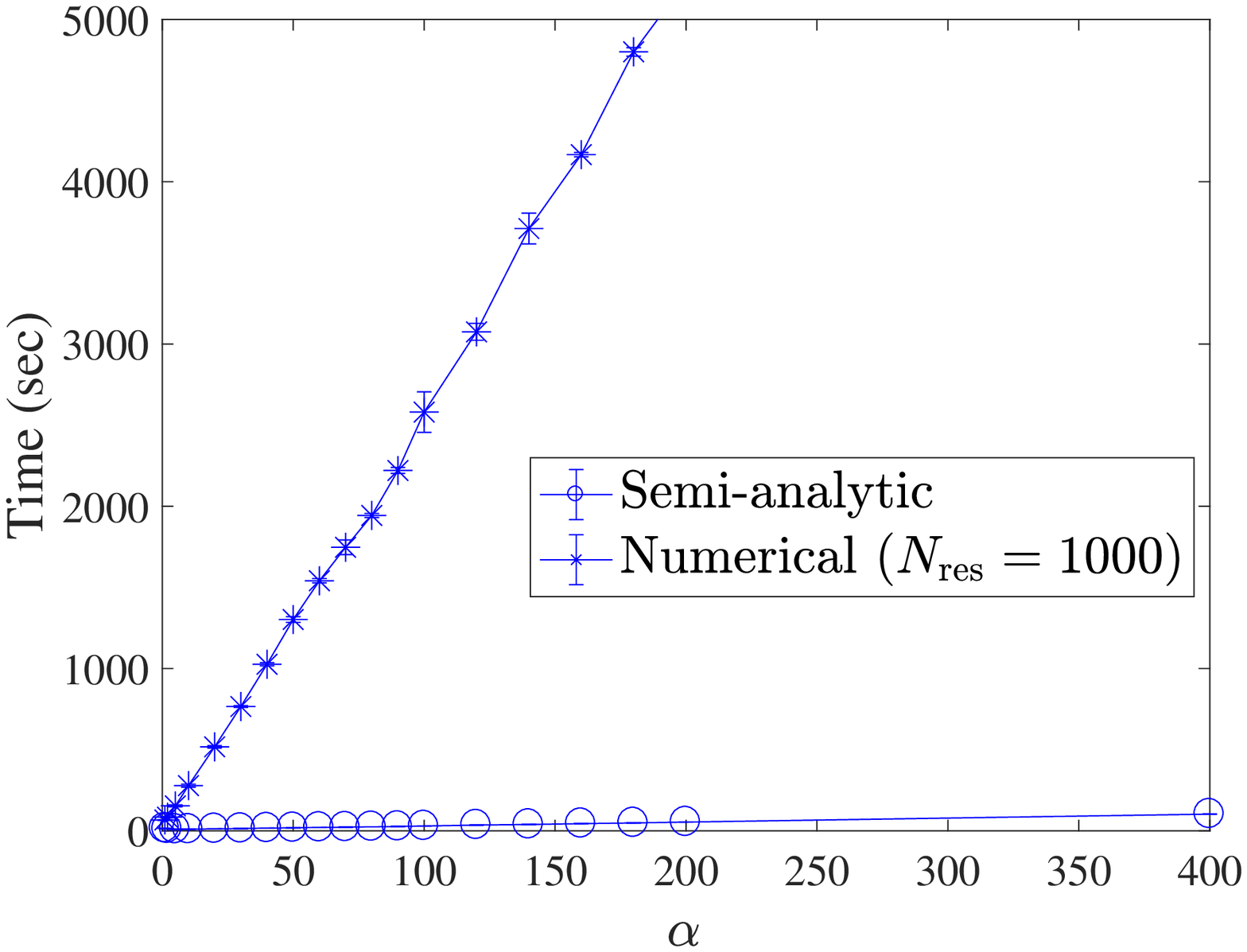}
\caption{Bolasso experiment at $(N,\rho_{0},\sigma_{\xi}^2)=(1000,0.2,0.01)$. The semi-analytic result (circle) is plotted with the numerical resampling result (asterisk). (Left) TP is plotted against FP as $\alpha$ increases: TP approaches unity, whereas FP tends to zero. The semi-analytic result completely overlaps that of the numerical resampling. Error bars are omitted for clarity. (Right) Computational time plotted against $\alpha$. The observed large difference is attributed to the numerical resampling cost.}
\Lfig{Bolasso}
\end{center}
\end{figure}
This demonstrates that the model consistency in the variable selection context is recovered. The contribution of this study is a significant reduction of the computational time: In the right panel, the actual computational time is compared between the semi-analytic method using AMPR and the direct numerical resampling, by plotting it against $\alpha$. The computational costs of AMPR and Glmnet are both scaled as $O(NM)$ and thus should be scaled linearly with respect to $\alpha$ for fixed $N$. \Rfig{Bolasso} clearly shows this linearity. The significant difference in computational time is fully attributed to the numerical resampling cost. This demonstrates the efficiency of AMPR. It should be noted that the average over $10$ different samples of $D$ is taken in \Rfig{Bolasso} to obtain smooth curves and error bars. 

\subsubsection{Application to stability selection}\Lsec{Application to stability}
SS is another variable selection method utilizing positive probability. The difference from Bolasso is the presence of the penalty coefficient randomization $(w<1)$ and that $\tau$ is set to $0.5$. These introduce a further stochastic variation in the method, and consequently they tend to show a clearer discrimination in the positive probabilities between the variables in and outside the true support. AMPR can easily implement this, and here, it is compared with numerical resampling.

\Rfig{stabsele} shows the plots of the positive probability values against $\lambda$, the so-called stability path~\citep{meinshausen2010stability}, and the computational time for obtaining the stability path against the covariate dimensionality $N$. The other parameters are fixed at $(\alpha,\rho_0,\sigma_{\xi}^2,w,p_w)=(2,0.2,0.01,0.5,0.5)$.  
\begin{figure}[htbp]
\begin{center}
 \includegraphics[width=0.45\columnwidth]{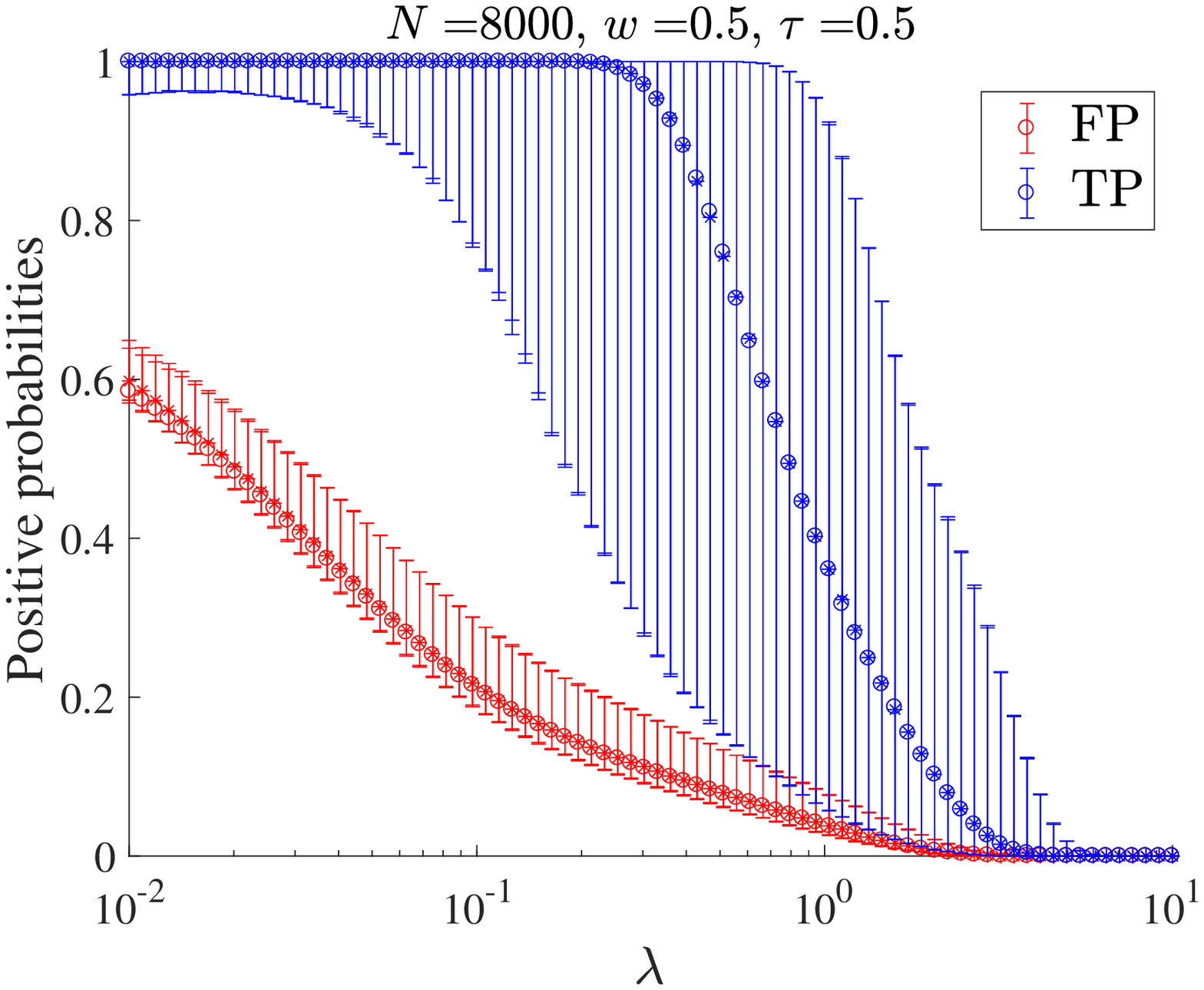}
 \includegraphics[width=0.45\columnwidth]{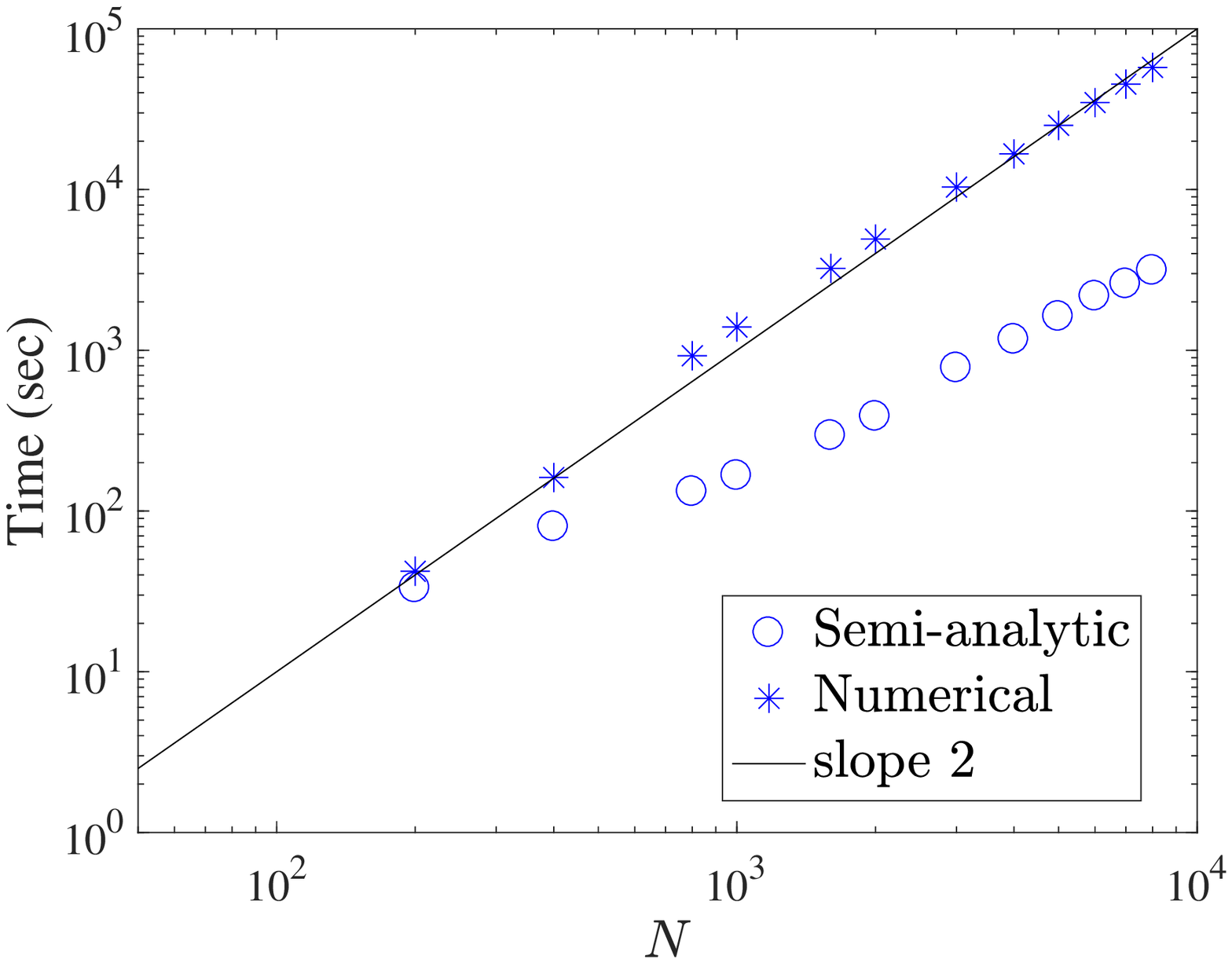}
\caption{SS experiment at $(\alpha,\rho_0,\sigma_{\xi}^2,w,p_w)=(2,0.2,0.01,0.5,0.5)$. The semi-analytic result (circle) is plotted with the numerical resampling result (asterisk). (Left) Plots of the medians (points) and $q$-percentiles (bars) of $\{\Pi_i\}_{i \in S_0}$ (TP) and $\{\Pi_i\}_{i \notin S_0}$ (FP) with $q=16$ and $84$ against $\lambda$ for $N=8000$. The semi-analytic result well overlaps that of the numerical resampling. There is a clear gap between TP and FP, suggesting that an accurate variable selection is possible. (Right) Computational time plotted against $N$ on double-logarithmic scale. The dominant reason for the difference is again the numerical resampling cost.}
\Lfig{stabsele}
\end{center}
\end{figure}
Drawing all stability paths $\{\Pi_i \}_i$ leads to an unclear plot; thus, only the median and $q$-percentiles of $\{\Pi_i\}_{i \in S_0}$ and those of $\{\Pi_i\}_{i \notin S_0}$ are plotted, which are denoted by TP and FP in the left panel of \Rfig{stabsele}, respectively. The medians are represented by points, and the percentiles of $q=16$ and $84$, approximately corresponding to one-sigma points in normal distributions, are represented by bars. There is a clear gap between TP and FP, suggesting that an accurate variable selection is possible by setting an appropriate threshold of probability value, as discussed by~\cite{meinshausen2010stability}. The right panel shows a plot of the computational time by AMPR and by the numerical resampling. Both are supposed to be scaled as $O(NM=N^2)$. Although such a scaling is not observed in the AMPR result, it is expected to appear for larger $N$. Again, a significant difference in computational time is present between the semi-analytic and the numerical resampling methods, demonstrating the effectiveness of the proposed method.

\subsubsection{Correlated covariates}\Lsec{Correlated covariates}
In the derivation of AMPR and the associated SE equations, the weakness of correlations between covariates is assumed, but this assumption does not necessarily hold in realistic situations. Hence, it is important to check the accuracy of results obtained by AMPR for datasets with correlated covariates. Here we numerically examine this point. 

To introduce correlations into the simulated dataset described above in a systematic way, we generate our covariates $\{ \V{x}_i \}_{i=1}^{N}$ in the following manner: As a common component we first generate a vector $\V{x}^{\rm com}\in \mR^{M}$ each component of which is i.i.d. from $\mc{N}(0,1/N)$; choose a number $0 \leq r^{\rm com}< 1$ controlling the ratio of the common component and generate a binary vector $\V{m}_i$ each component of which independently takes $1$ with probability $r^{\rm com}$; take another vector $\tilde{\V{x}}_i\in \mR^{M}$ each component of which is i.i.d. from $\mc{N}(0,1/N)$, and generate a covariate vector $\V{x}_i$ as a linear combination between $\V{x}^{\rm com}$ and $\tilde{\V{x}}_i$ with using $\V{m}_i$ as a mask. These operations can be summarized in the following equation: 
\be
\V{x}_i=\V{x}^{\rm com} \circ \V{m}_i(r^{\rm com})  +\tilde{\V{x}}_{i} \circ  (1-\V{m}_i(r^{\rm com})) ,
\ee
where $\circ$ denotes Hadamard (component-wise) product. The covariates' correlations monotonically increase as $r^{\rm com}$ grows. To quantify this, we compute an overlap between the covariates, ${\rm overlap}_{ij}=\V{x}_i^{\top}\V{x}_j / \lb \sqrt{ \V{x}_i^{\top}\V{x}_i } \sqrt{ \V{x}_j^{\top}\V{x}_j }  \rb$, and plot its mean value over $i$ and $j~ (\neq i)$ in \Rfig{overlap}.
\begin{figure}[htbp]
\begin{center}
 \includegraphics[width=0.45\columnwidth]{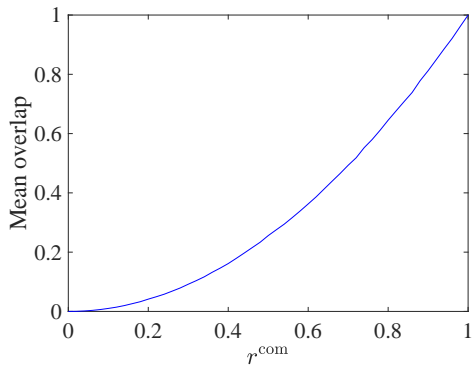}
\caption{ Plot of mean overlap of covariates against the common component ratio $r^{\rm com}$. This is a result of a single numerical simulation at $(N,\alpha)=(1000,0.5)$, but the invariability of the result against changing the parameters has been also checked.  }
\Lfig{overlap}
\end{center}
\end{figure}
Keeping in mind this quantitative information, we check the accuracy of AMPR below.

To directly see the accuracy of AMPR for correlated covariates, we plot the experimental values of the quantities $\{\overline{\beta_{i}},W_i,\Pi_i\}_i$ against those of AMPR in \Rfig{yyplot-correlated} for two different values of the common component ratio, $r^{\rm com}=0.4$ and $r^{\rm com}=0.8$. 
\begin{figure}[htbp]
\begin{center}
 \includegraphics[width=0.32\columnwidth]{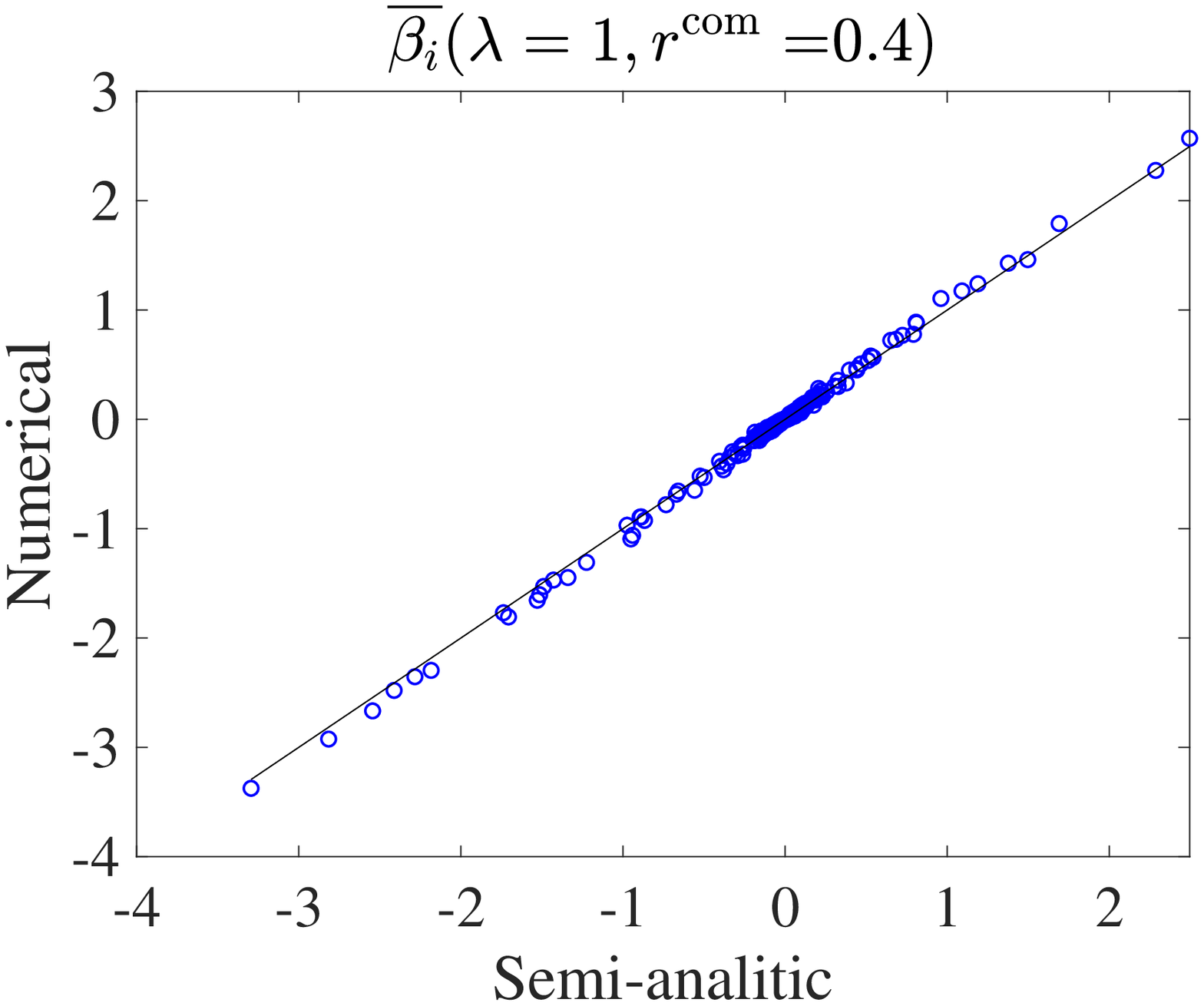}
 \vspace{1mm}
 \includegraphics[width=0.32\columnwidth]{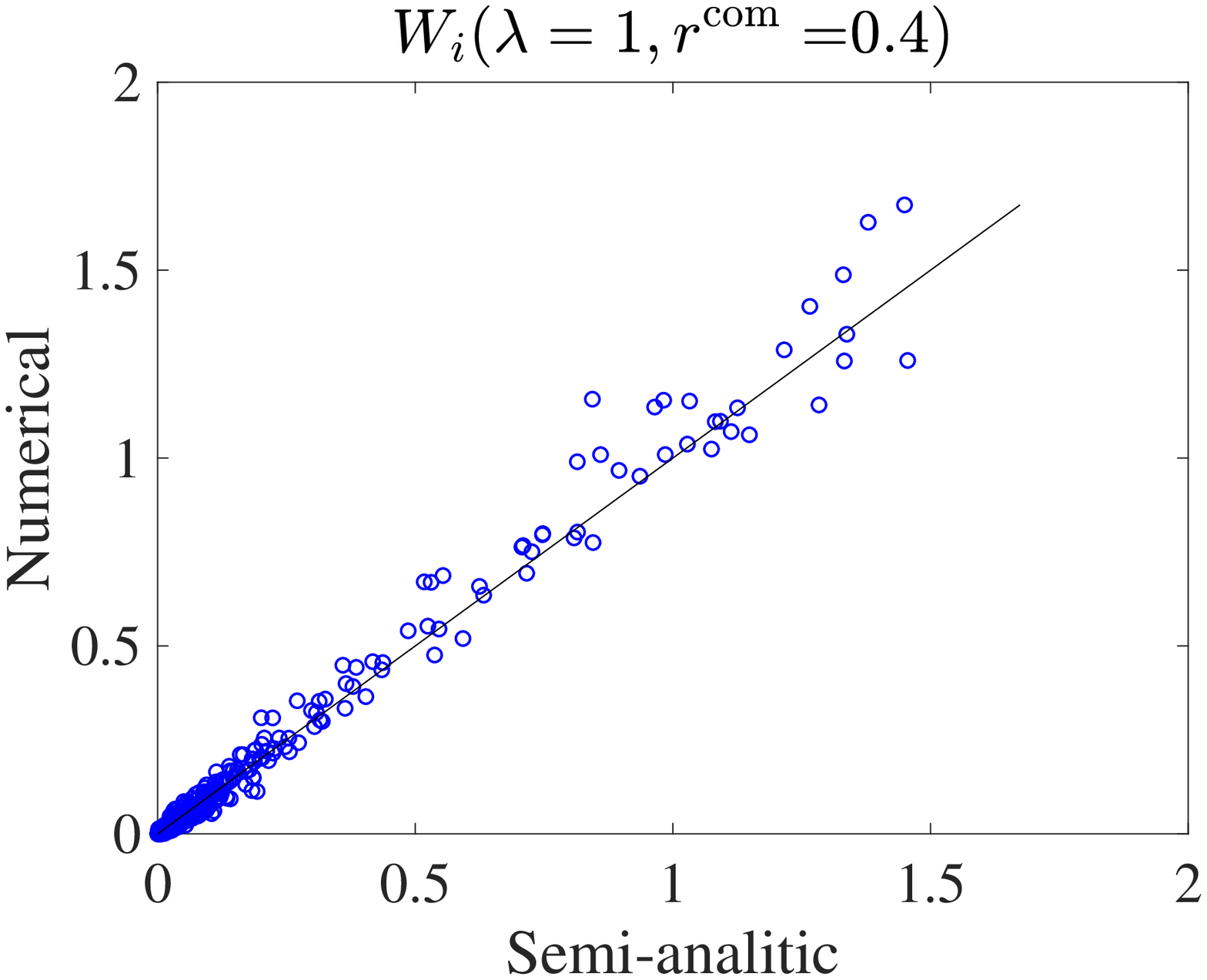}
 \vspace{1mm}
 \includegraphics[width=0.32\columnwidth]{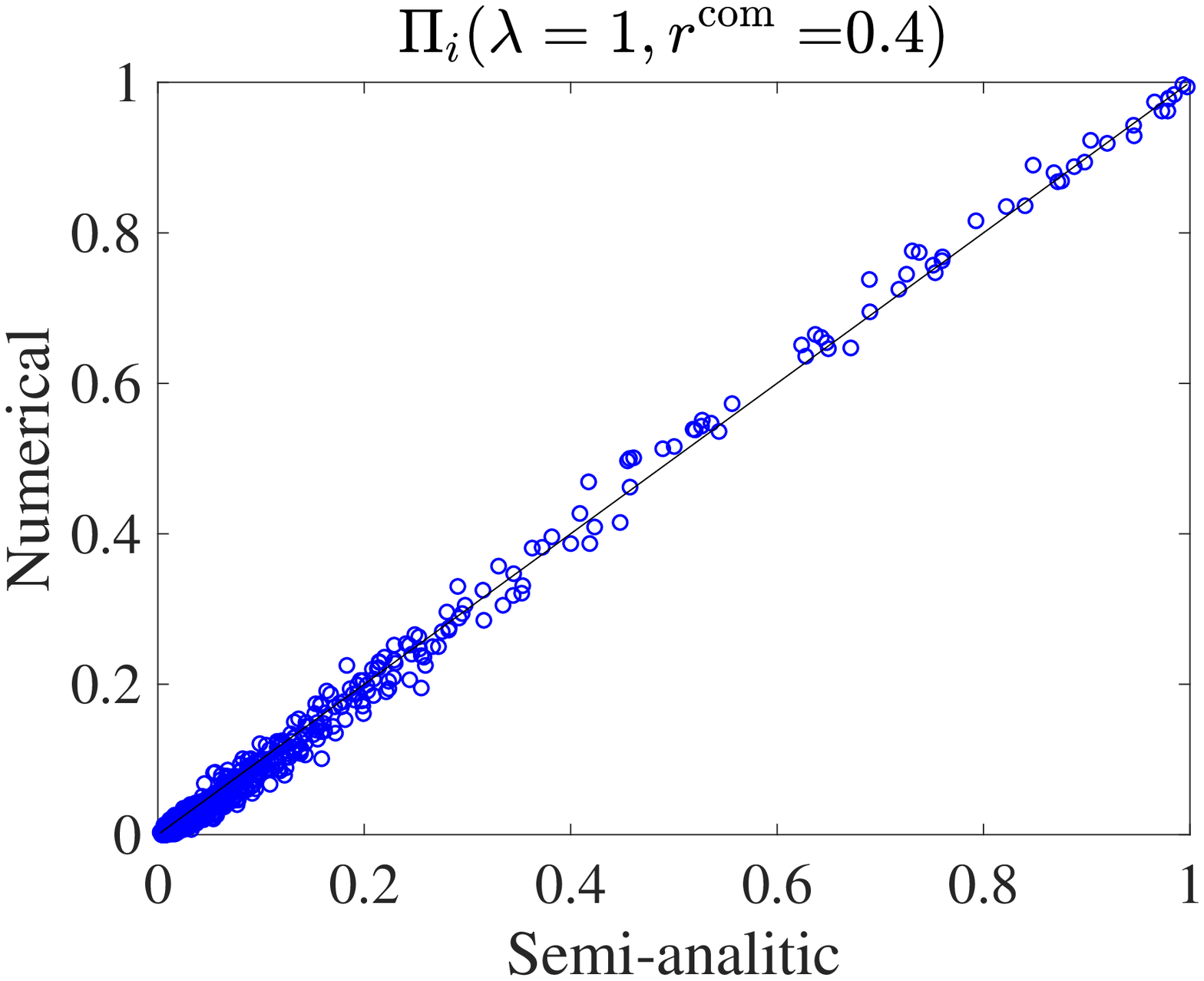}
 \vspace{1mm}
 \includegraphics[width=0.32\columnwidth]{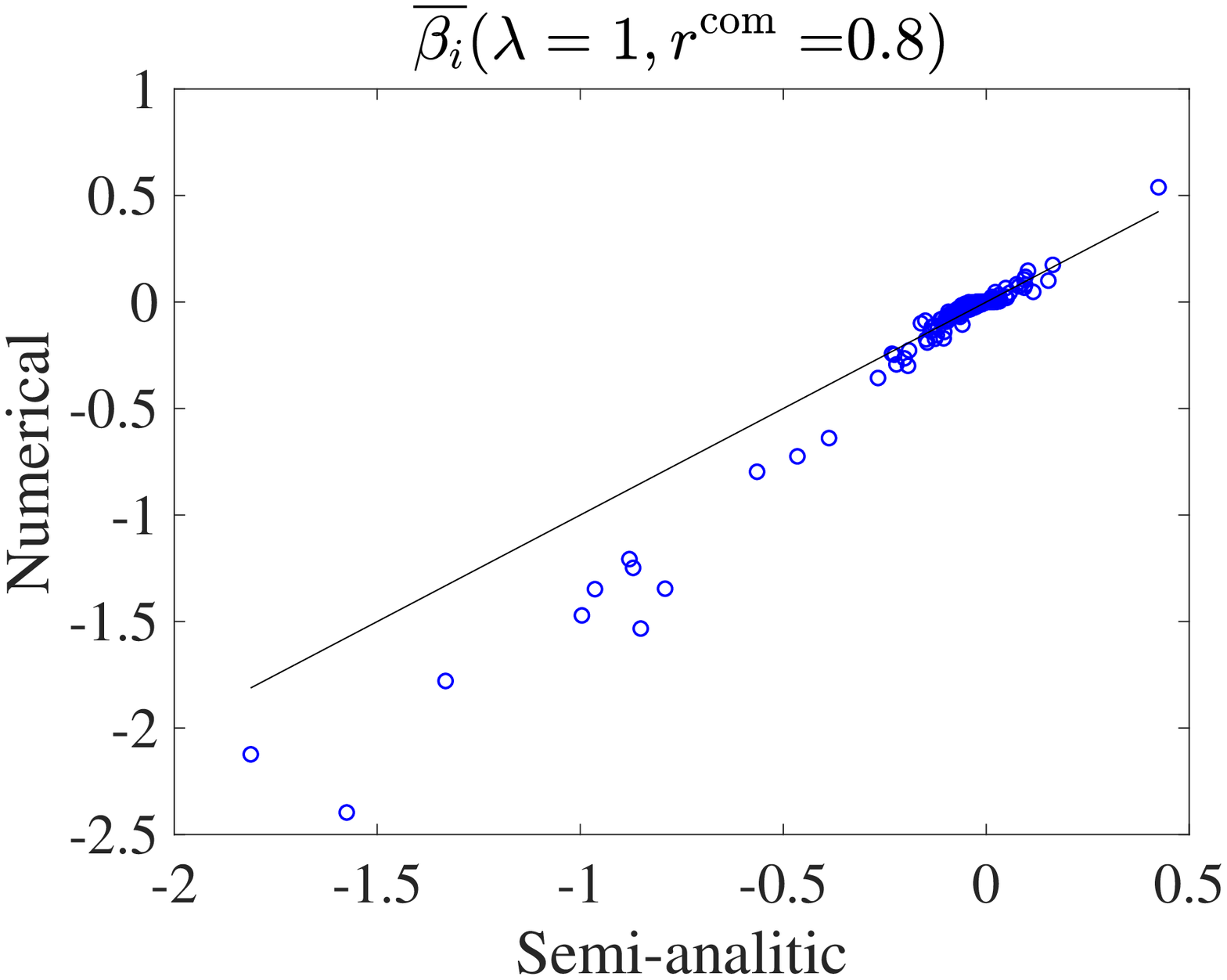}
 \includegraphics[width=0.32\columnwidth]{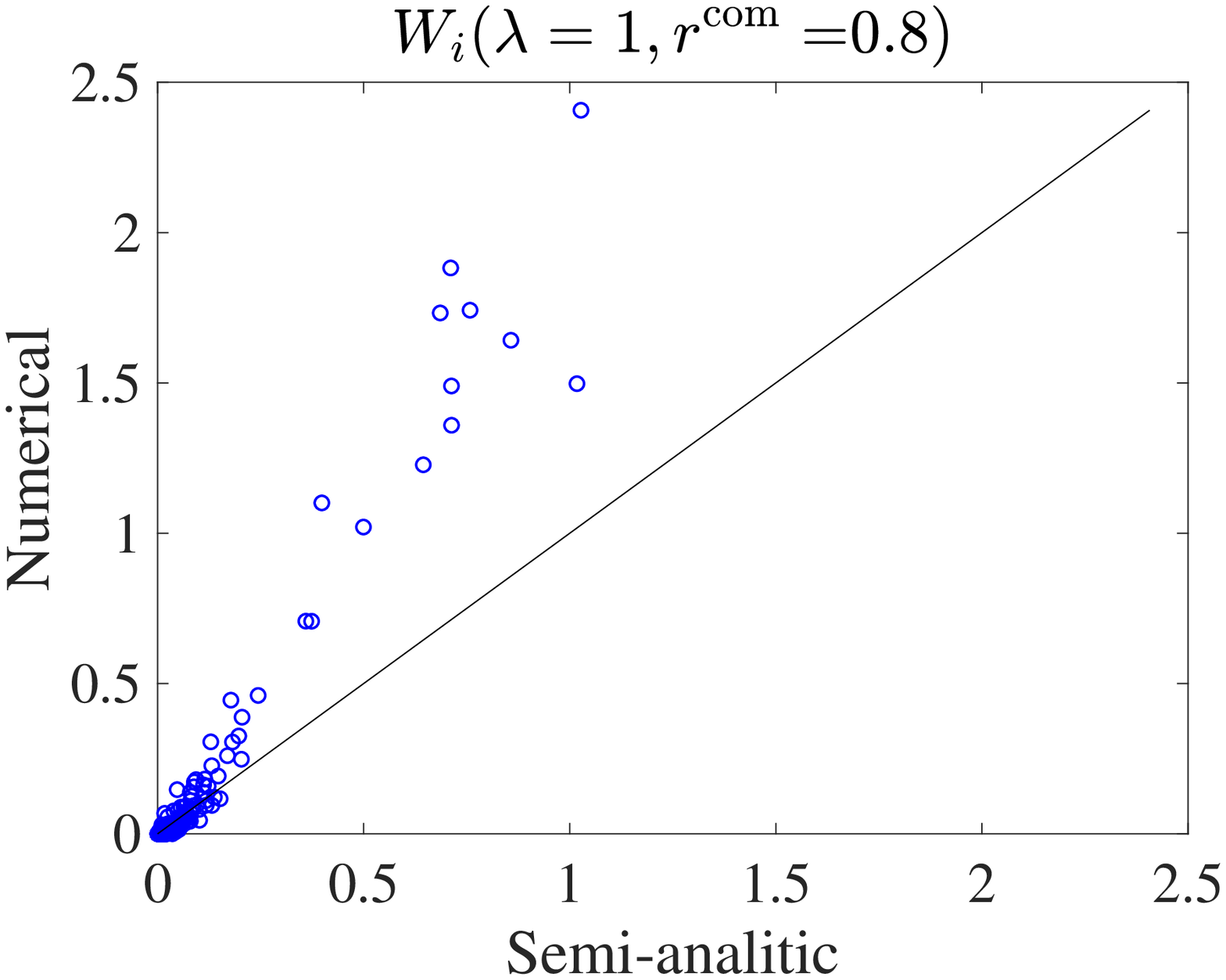}
 \includegraphics[width=0.32\columnwidth]{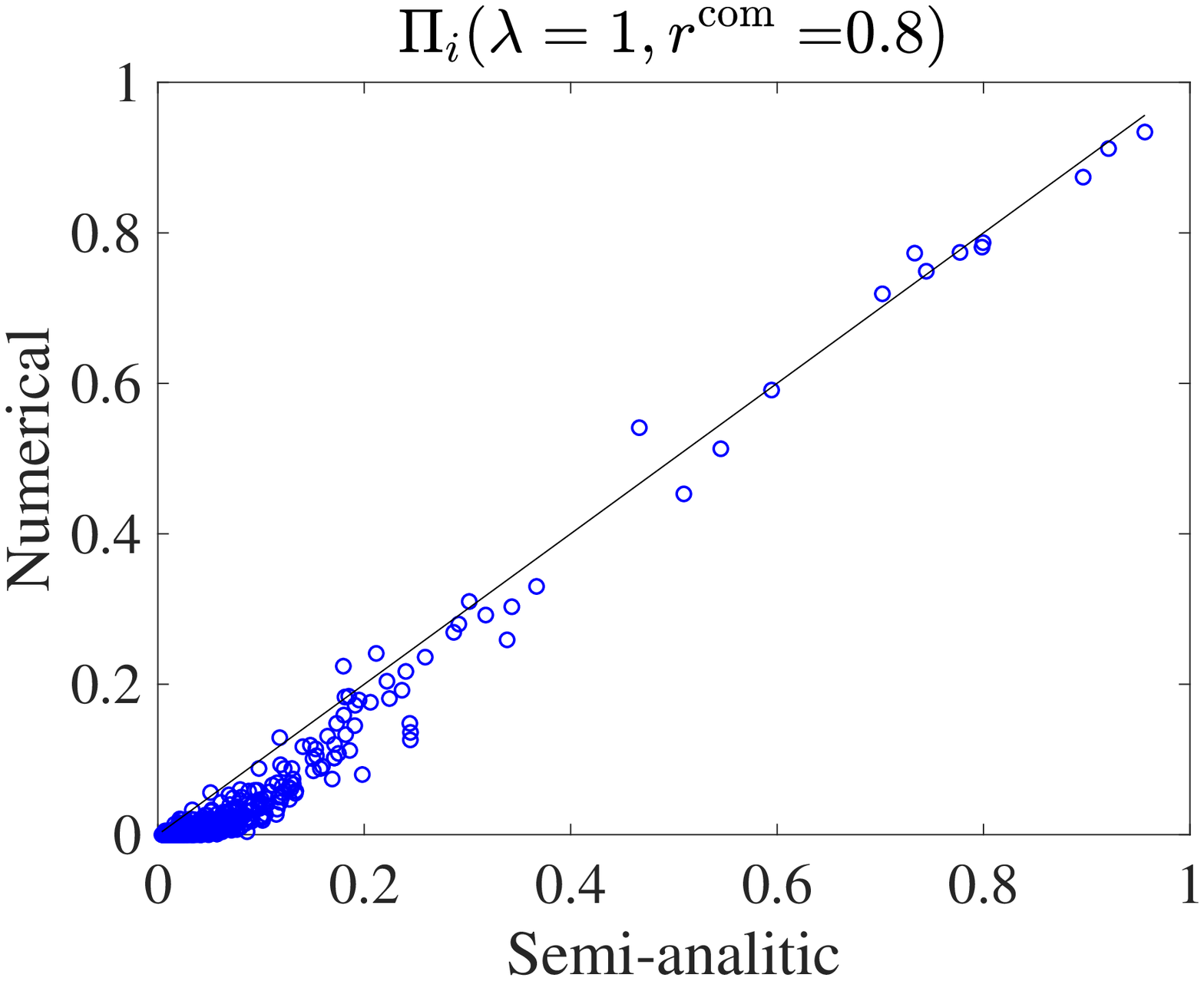}
\caption{Experimental values of $\overline{\beta_i}$ (left), $W_{i}$ (middle), and $\Pi_{i}$ (right) are plotted against those computed by AMPR for correlated covariates with $r^{\rm com}=0.4$ (upper) and $r^{\rm com}=0.8$ (lower). For the lower panels, there exists a systematic deviation in the AMPR result. Here the non-random penalty case ($w=1, \tau=1)$ is treated and the other parameters are $(N,\alpha,\rho_{0},\sigma_{\xi}^2,\lambda)=(1000,0.5,0.2,0.01,1)$.}
\Lfig{yyplot-correlated}
\end{center}
\end{figure}
For $r^{\rm com}=0.8$, the AMPR result shows a clear deviation from the experimental one, while for $r^{\rm com}=0.4$ it still exhibits a good agreement with the experiment. Although \Rfig{yyplot-correlated} is the specific result to the non-random penalty case ($w=1, \tau=1)$ with parameters $(N,\alpha,\rho_{0},\sigma_{\xi}^2,\lambda)=(1000,0.5,0.2,0.01,1)$, we have tested several different cases and observed similar tendency. To capture more global and quantitative information, we introduce a normalized MSE of $\overline{\beta}$ between the experimental $\overline{\beta}^{\rm exp}$ and AMPR $\overline{\beta}^{\rm AMPR}$ values as
\be
\frac{\sum_{i=1}^{N}\lb \overline{\beta}^{\rm exp}_i - \overline{\beta}^{\rm AMPR}_i \rb^2}{\sum_{i=1}^{N}\lb \overline{\beta}^{\rm AMPR}_i \rb^2}.
\ee
The counterparts for $W$ and $\Pi$ are defined in the same way. Plots of the normalized MSEs for these quantities against $r^{\rm com}$ are given in \Rfig{r-sweep}.
\begin{figure}[htbp]
\begin{center}
 \includegraphics[width=0.45\columnwidth]{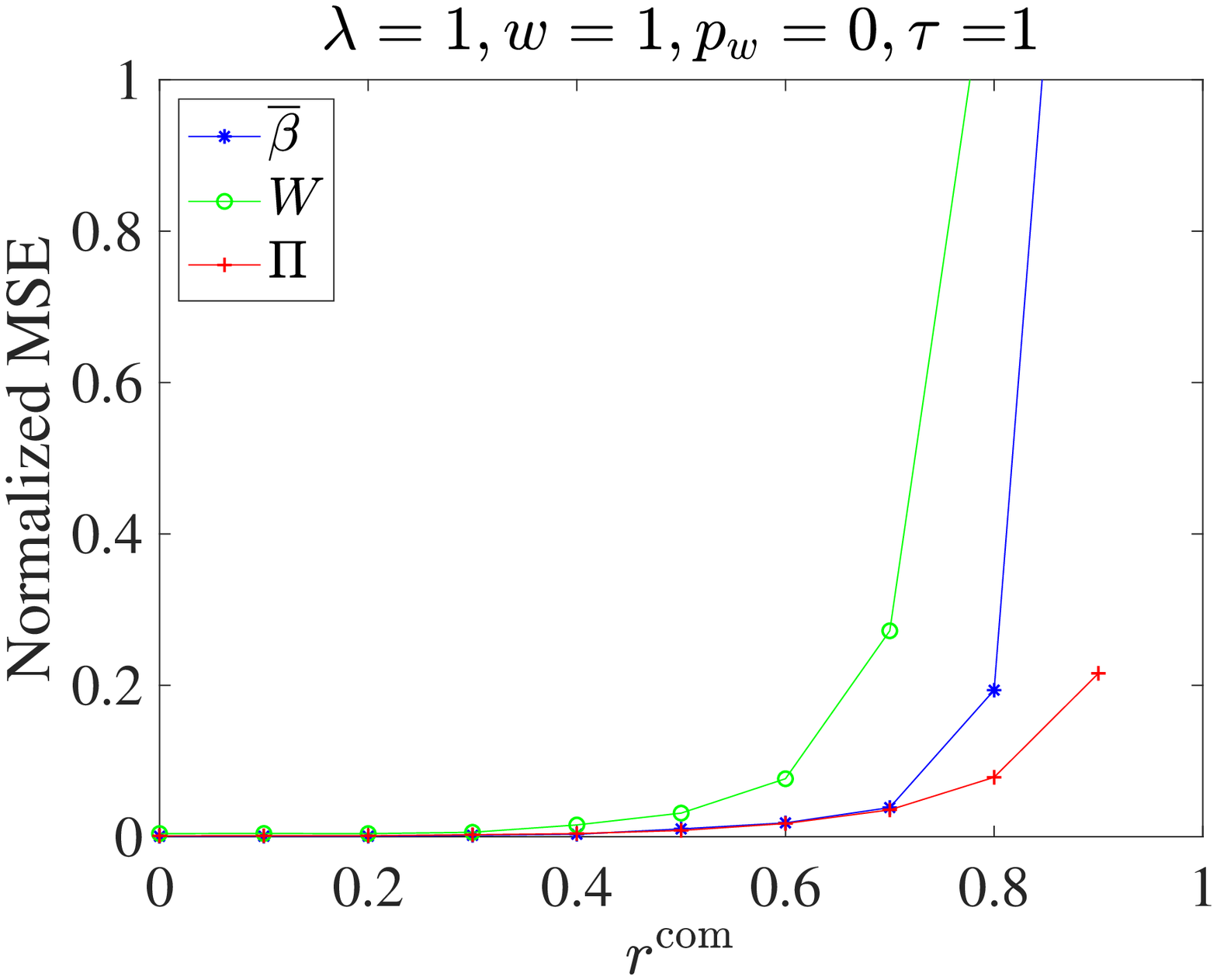}
 \vspace{1mm}
 \includegraphics[width=0.45\columnwidth]{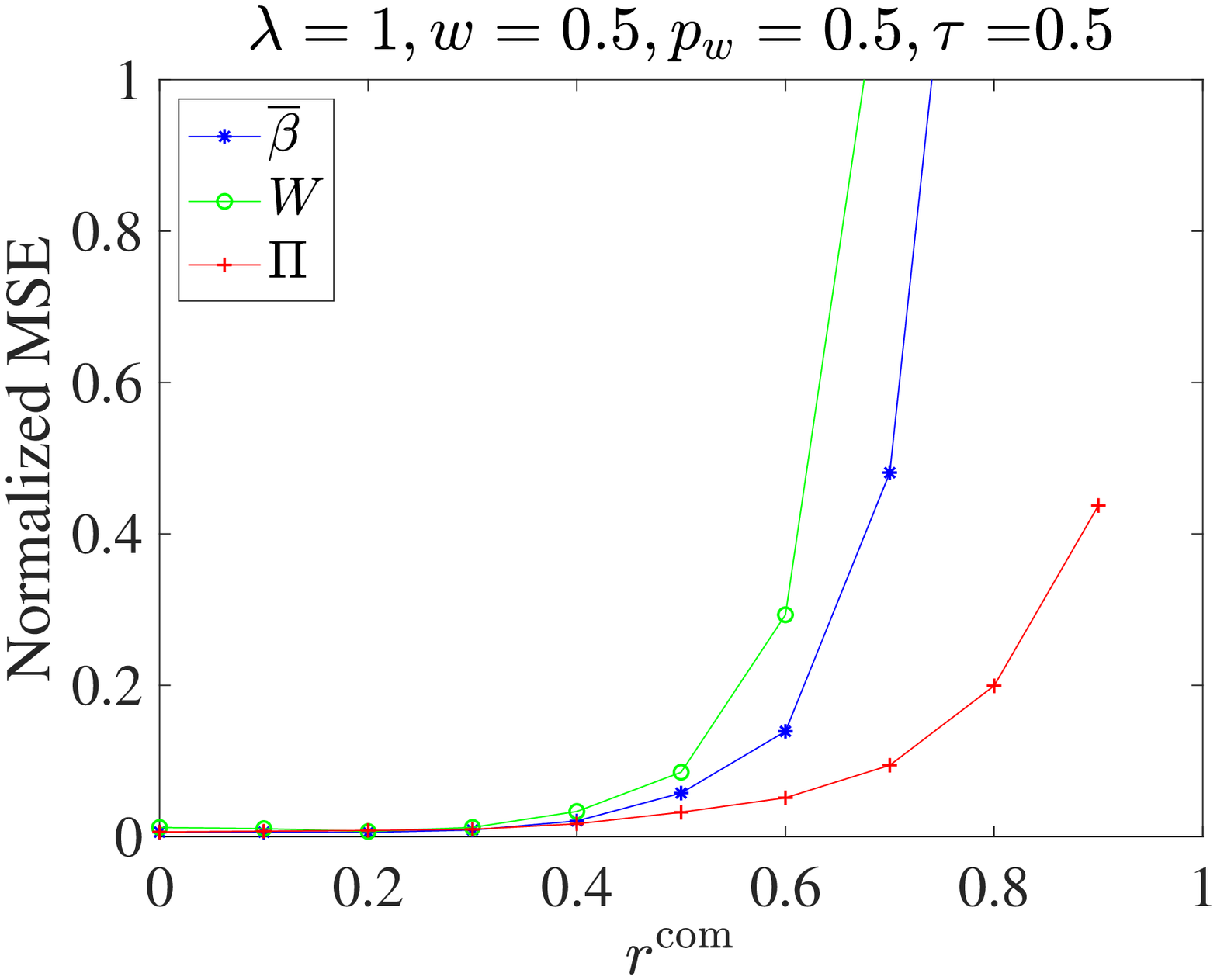}
  \vspace{1mm}
 \includegraphics[width=0.45\columnwidth]{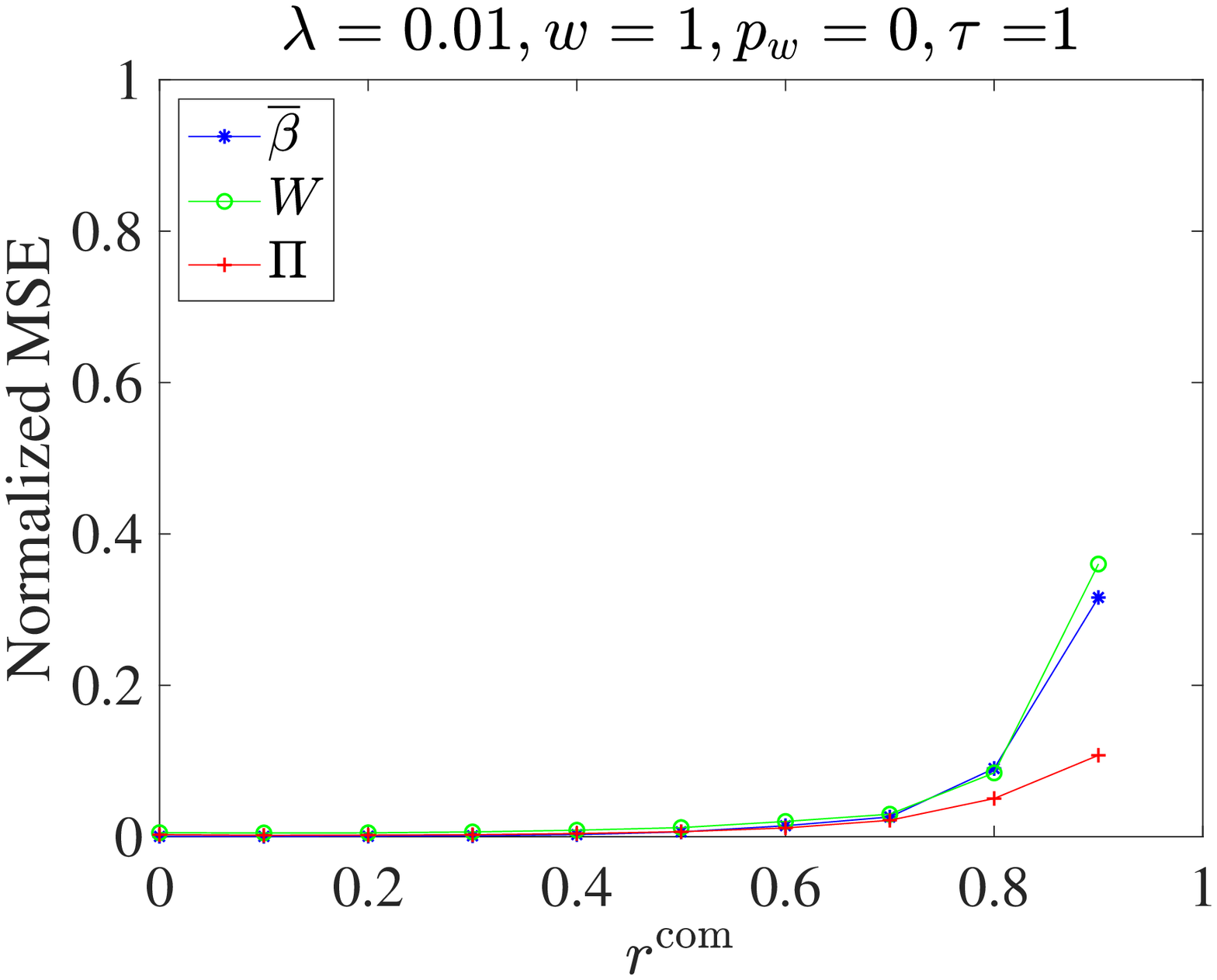}
 \includegraphics[width=0.45\columnwidth]{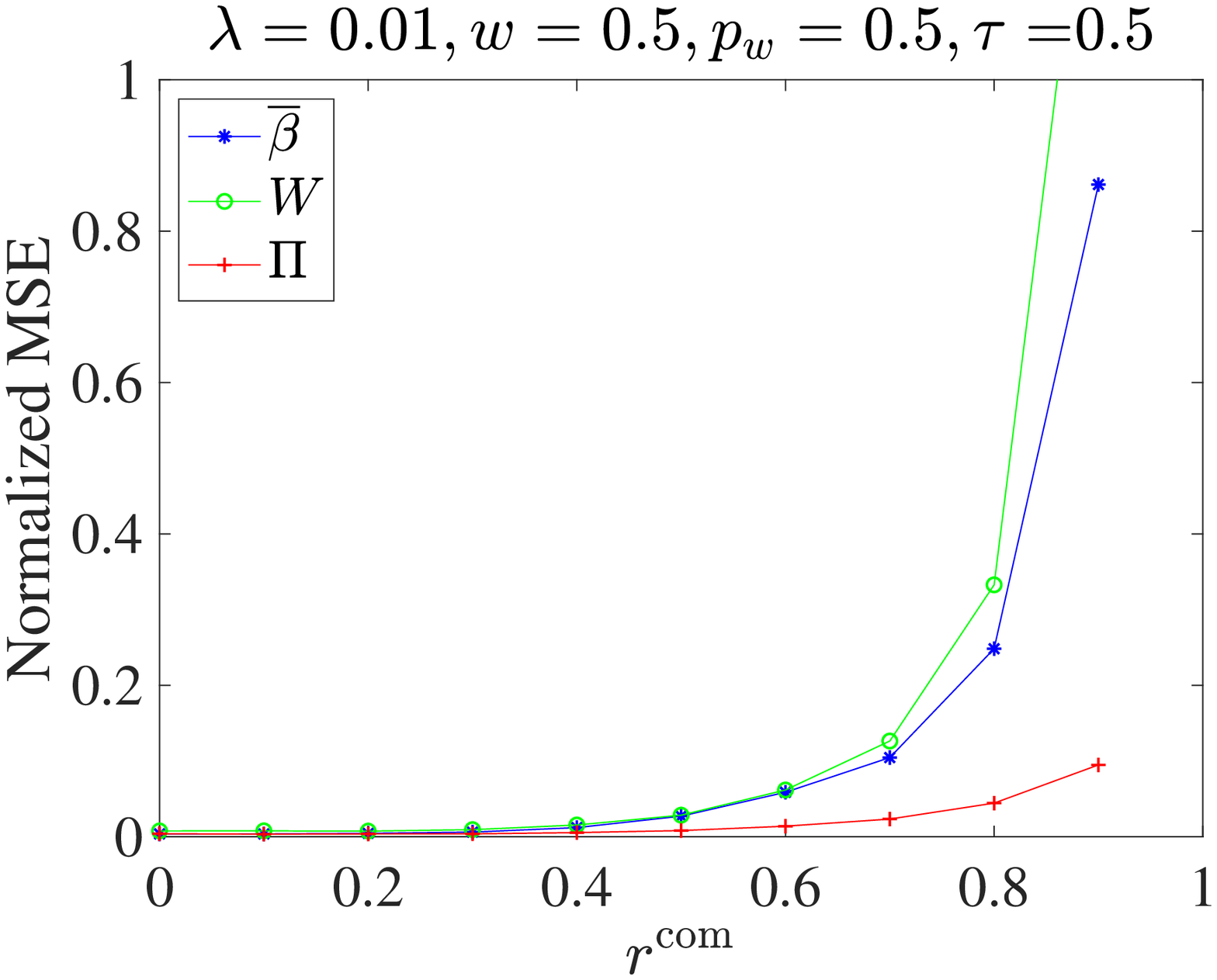}
\caption{Plots of normalized MSEs of $\overline{\beta},W,\Pi$ against $r^{\rm com}$ of the non-randomized (left, $w=1,p_w=0,\tau=1$) and randomized (right, $w=p_w=\tau=1/2$) penalty cases. The upper panels are for $\lambda=1$ while the lower ones are of $\lambda=0.01$. The other parameters are $(N,\alpha,\rho_{0},\sigma_{\xi}^2)=(1000,0.5,0.2,0.01)$.  }
\Lfig{r-sweep}
\end{center}
\end{figure}
Here the results for both the randomized ($w=p_w=\tau=0.5$) and non-randomized ($w=\tau=1,p_w=0$) penalty cases are presented, with two different values of $\lambda$. This suggests that AMPR is practically reliable up to a certain level of correlations: For example if $r^{\rm com}\leq 0.6$, the normalized MSE of $\overline{\beta}$ is well suppressed and is commonly less than $0.2$. According to \Rfig{overlap}, $r^{\rm com}=0.6$ corresponds to the mean overlap value of about $0.36$ which is not small. This speaks for the effectiveness of AMPR even for real-world datasets, as long as the correlations are not too large.

The above discussion clarifies the accuracy of AMPR for correlated covariates, but a more crucial issue is in its convergent property. For the non-correlated case of $r^{\rm com}=0$, AMPR converges very rapidly as shown in \Rfig{SE}, but for correlated cases it tends to badly converge and can even diverge. A common way to overcome this is to introduce a damping factor $\gamma$ in the update~\citep{caltagirone2014convergence,rangan2014convergence}, and we adopted this in the above experiments. The update with the damping factor can be symbolized as
\be
\V{\theta}^{(t+1)}=(1-\gamma)\V{\theta}^{(t)}+\gamma G(\V{\theta}^{(t)}),
\ee
where $\V{\theta}$ is a variable summarizing $(\overline{\V{\beta}},\V{\chi},\V{W})$ and $G$ represents the AMPR operations. The original message passing update corresponds to $\gamma=1$. Smaller values of $\gamma$ are better for the stability of the algorithm but takes a longer time until the convergence. Our experiments on the simulated dataset seemingly show that a smaller value of $\gamma$ is needed for larger $r^{\rm com}$ and smaller $\lambda$. For example for obtaining \Rfig{yyplot-correlated}, we needed to set $\gamma < 0.1$ for avoiding divergence. This value is found experimentally, and it is desired to find a more principled way to choose an appropriate value of $\gamma$ or a more effective way to better control the convergence of AMP type algorithms. Some earlier studies tackled this problem~\citep{caltagirone2014convergence,rangan2014convergence}, but the complete understanding of the convergent property is still missing. Investigation along this direction is an important topic but is beyond the purpose of the present paper.

\subsection{Real-world dataset}\Lsec{Real-world dataset}
In this subsection, AMPR is applied to a real-world dataset, and the stability path is computed for examining the relevant covariates or variables. The dataset treated here is the wine quality dataset~\citep{Lichman:13}: The data size is $M=4898$ (only white wine is treated), and the number of covariates, which represent physicochemical aspects of wine, is $N_{0}=11$. The basic aim of this dataset is to model wine quality in terms of the physicochemical aspects only, and the response is an integer-valued quality score from zero to ten obtained by human expert evaluation. This dataset can be used in both classification and regression, and the linear model was also tested in earlier studies~\citep{cortez2009modeling,melkumova2017comparing}. Lasso and SS are applied to this dataset.

As preprocessing, $N_{\rm noise}$ noise variables are added into the dataset. Each component of the noise variables is i.i.d. from $\mathcal{N}(0,1/N)$. The usual standardization, zeroing the mean of all variables and responses and normalizing the variables to be of unit norm, is also conducted. 

The reason of the introduction of the noise variables is to make an objective criterion for judging relevant stability paths. The noise variables are by definition irrelevant for describing the responses and hence if some stability paths of original variables behave similarly to the ones of the noise variables then those variables can be regarded as irrelevant. A distribution of stability paths of the noise variables thus defines a kind of ``rejection region'', resolving the arbitrariness of judging criterion of the positive probability both in Bolasso and SS. As far as we have searched, this kind of active construction of rejection regions has never been observed in literatures, and hence this proposition itself can be regarded as a part of new results of this paper.

This sort of manipulation to dataset is usually unfavored because it increases the computational cost and tends to contaminate the dataset. However thanks to AMPR, the first computational issue becomes less serious since the computational time of AMPR increases just linearly with respect to the number of added noise variables $N_{\rm noise}$. The second issue does not seem to be serious either for the wine quality dataset, because the dataset size $M=4898$ is very large compared to the number of original variables $N_0=11$. Below, we experimentally check if this strategy works well or not.

Let us start from checking how the noise variables influence on the dataset. To this end, we plot the solution paths and the generalization errors estimated by 10-fold cross validation (CV) in \Rfig{CV-wine} in the cases with and without the noise variables. 
\begin{figure}[htbp]
\begin{center}
 \includegraphics[width=0.45\columnwidth]{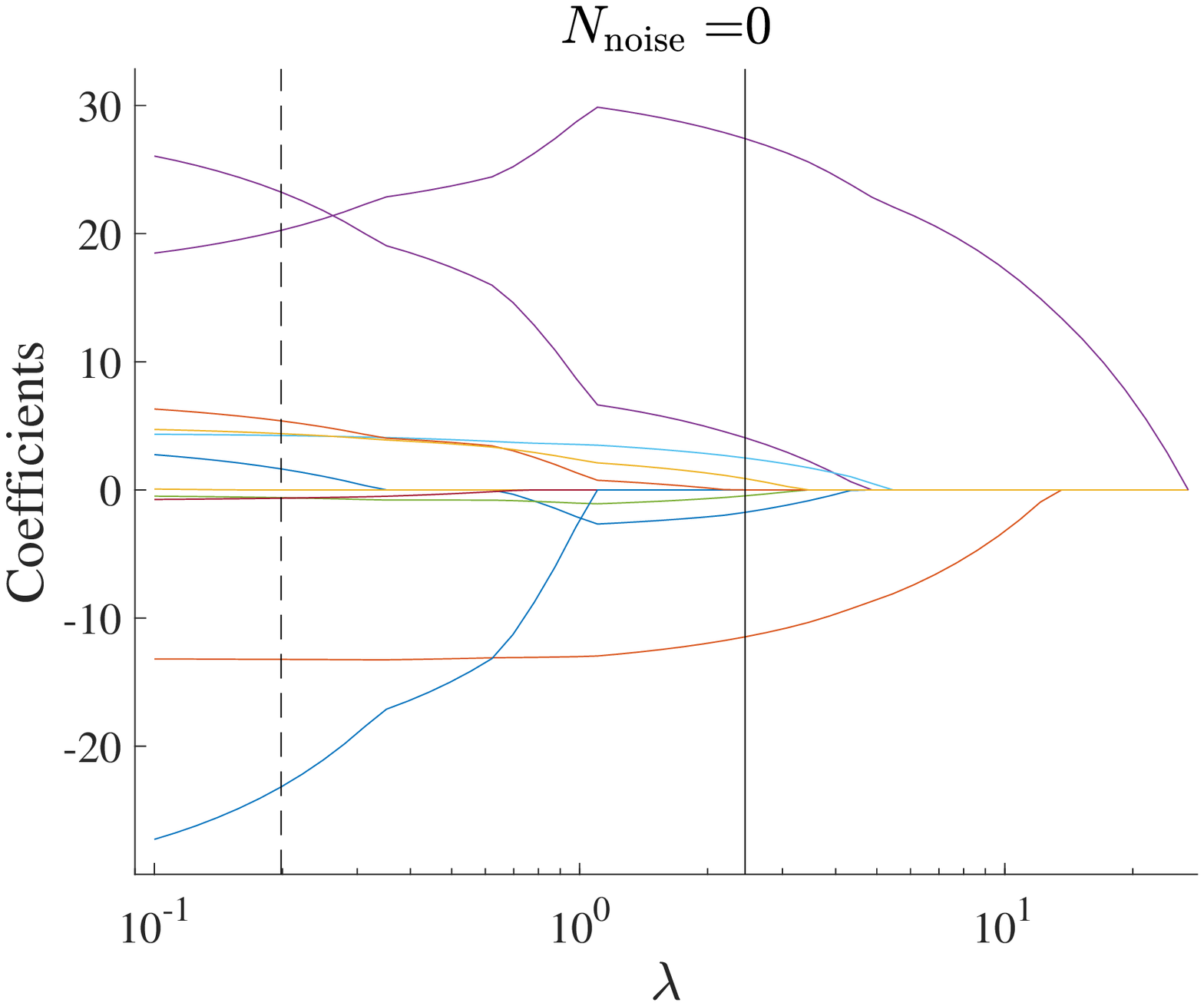}
 \includegraphics[width=0.45\columnwidth]{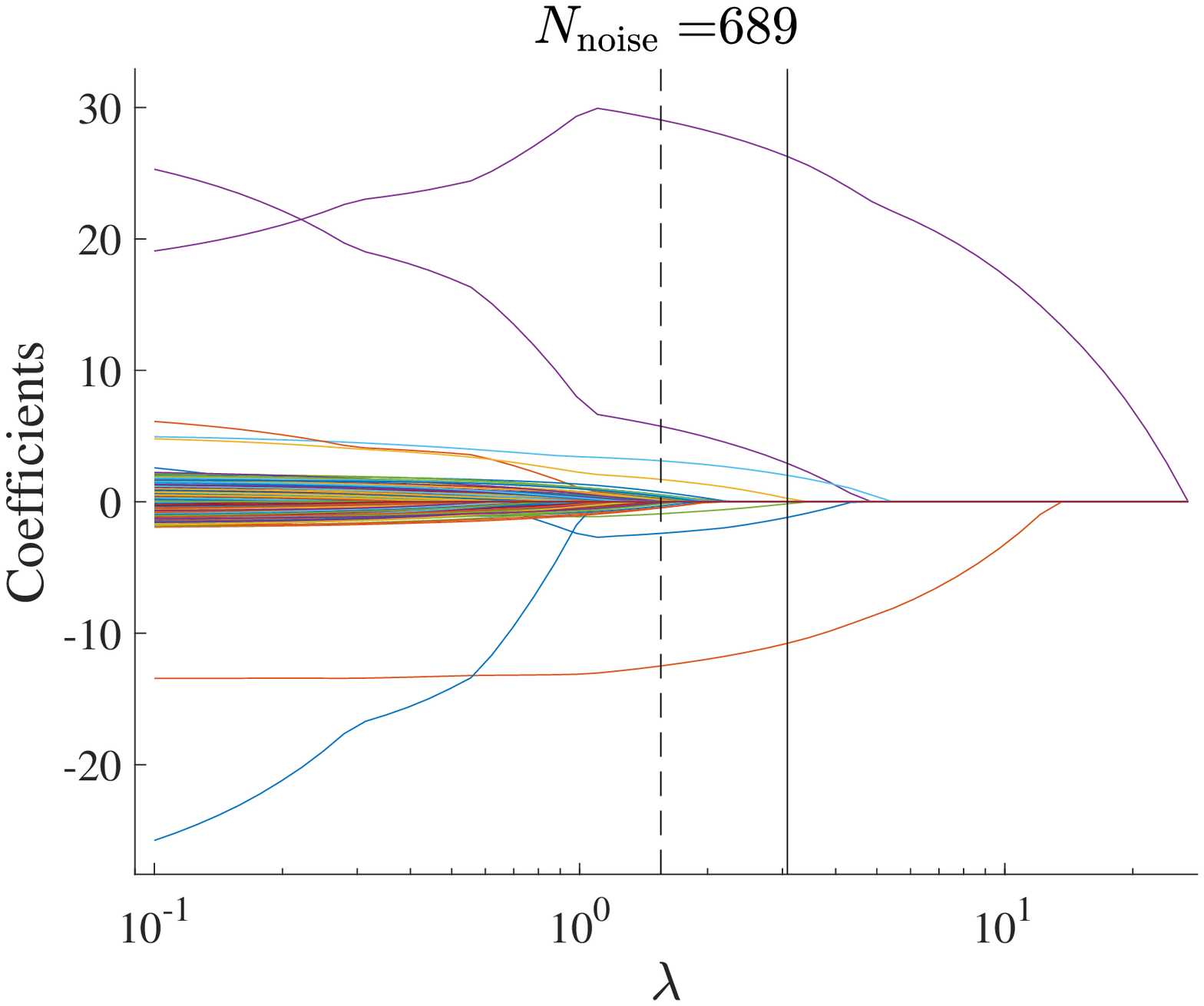}
 \includegraphics[width=0.45\columnwidth]{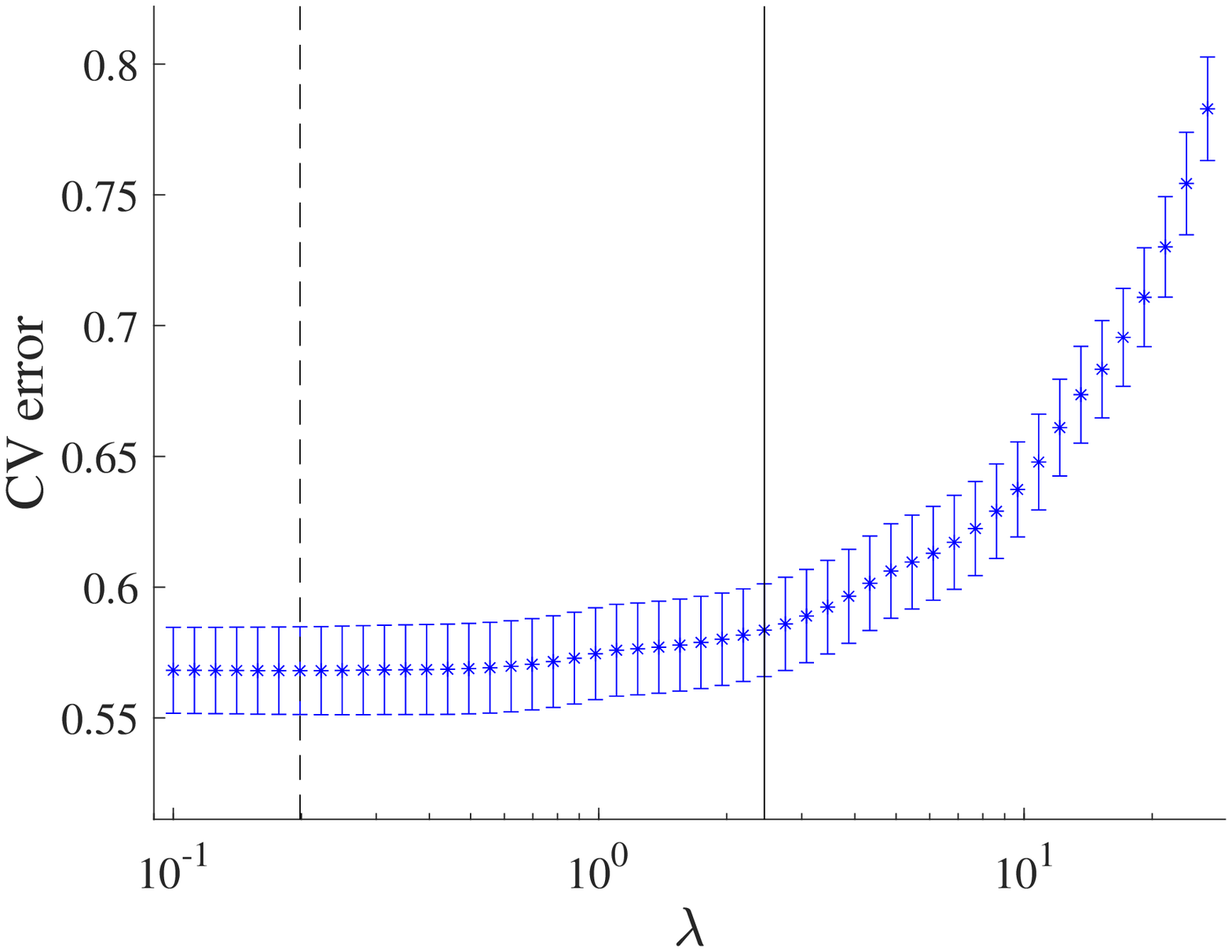}
 \includegraphics[width=0.45\columnwidth]{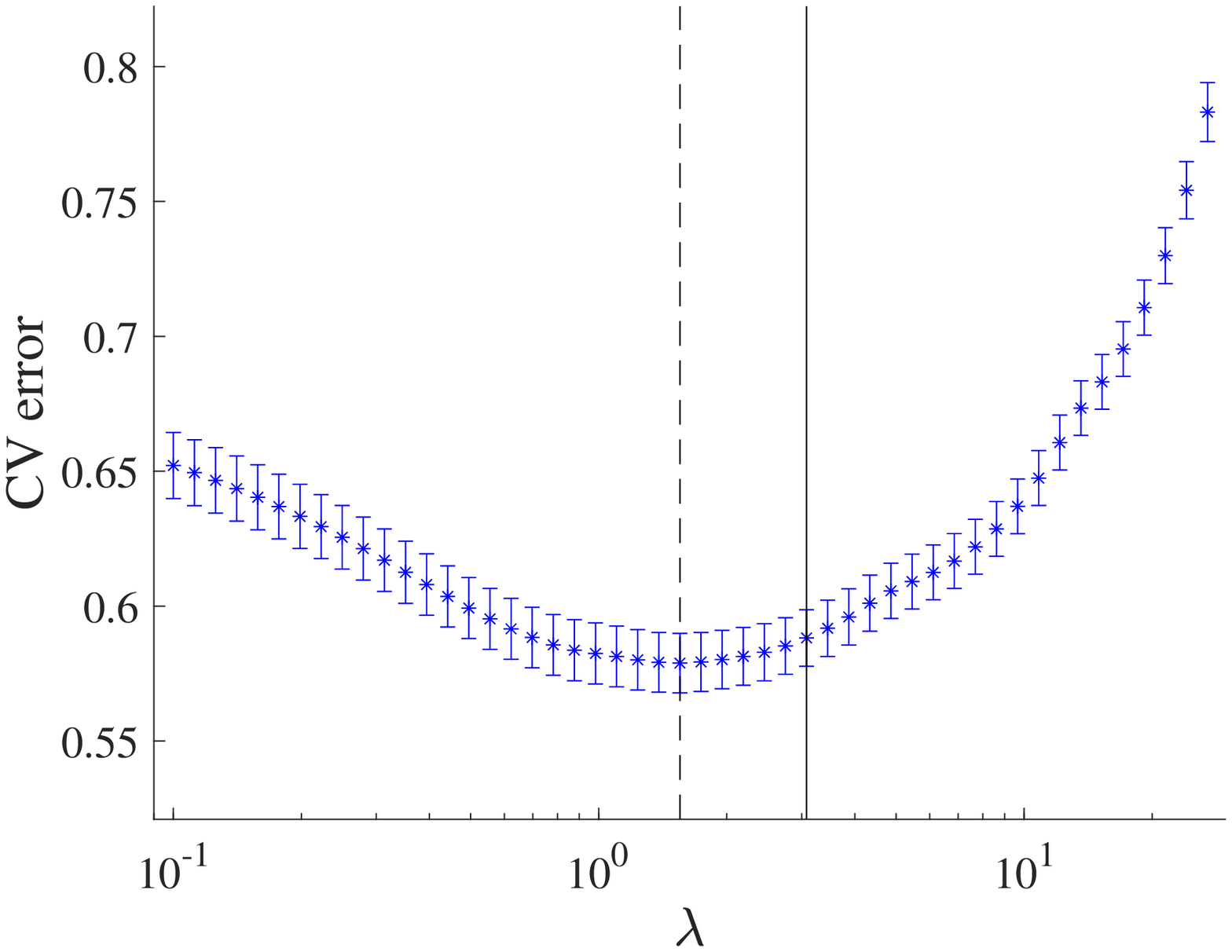}
\caption{Solution paths (top) and CV errors plotted against $\lambda$ (bottom), for $N{\rm noise}=0$ (left) and $N{\rm noise}=689$ (right). The vertical dotted line denotes the location of the CV error minimum, $\lambda_{\rm min}$, whereas the vertical straight line represents the location selected by the one-standard-error rule that ``defines'' the optimal $\lambda$ value, $\lambda_{\rm opt}$. The effect of the noise variables addition is weak around $\lambda_{\rm opt}$.
 }
\Lfig{CV-wine}
\end{center}
\end{figure}
This figure indicates that the solution paths of the original variables and the CV error value are stable against the introduction of noise variables, particularly in the relevant range of $\lambda$. At the optimal $\lambda$ ($\lambda_{\rm opt}$), chosen by the one-standard-error rule~\citep{Book:Hastie:09}, the variables in the support are common in both cases: They are indexed as $1,2,4,5,6,10$, and $11$, which represent fixed acidity, volatile acidity, residual sugar, chlorides, free sulfur dioxide, sulphates, and alcohol, respectively. Hence, we can conclude that the introduction of the noise variables hardly affects the estimates of the original variables, and we can effectively use the noise variables to judge the significance of each original variable.

Subsequently, the result of applying SS with $(\tau,w,p_w)=(0.5,0.5,0.5)$ to this dataset is examined. \Rfig{SS-wine} shows the plots of the stability paths according to the three categories of variables: The ``important'' variables are defined as being in the support, at $\lambda_{\rm opt}$ in the Lasso analysis, and thus their indices are $1,2,4,5,6,10,11$; the ``neutral'' variables are the other variables in the original dataset, and the corresponding indices are $3,7,8,9$; the remaining are the noise variables.  
\begin{figure}[htbp]
\begin{center}
 \includegraphics[width=0.45\columnwidth]{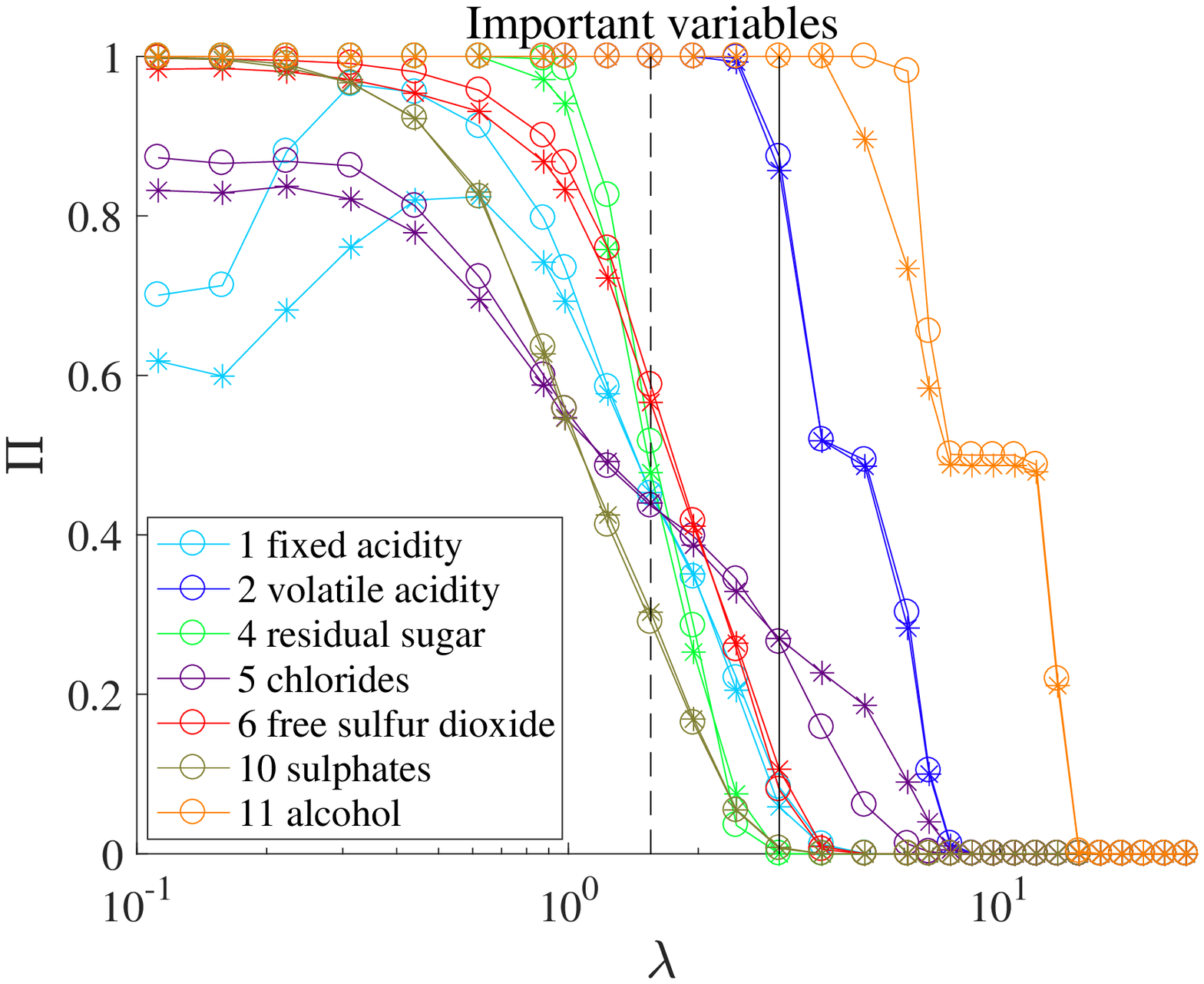}
 \includegraphics[width=0.45\columnwidth]{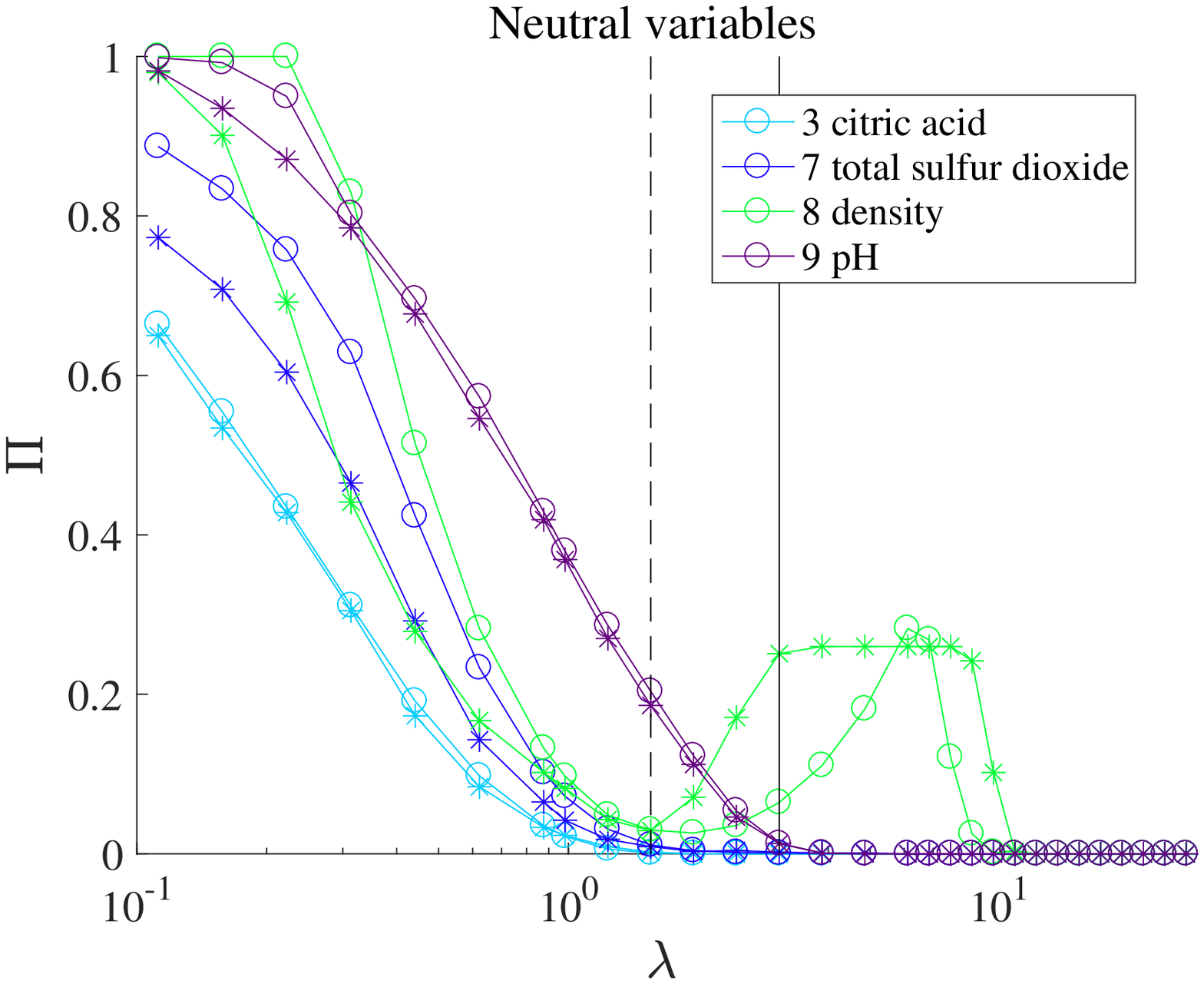}
 \includegraphics[width=0.45\columnwidth]{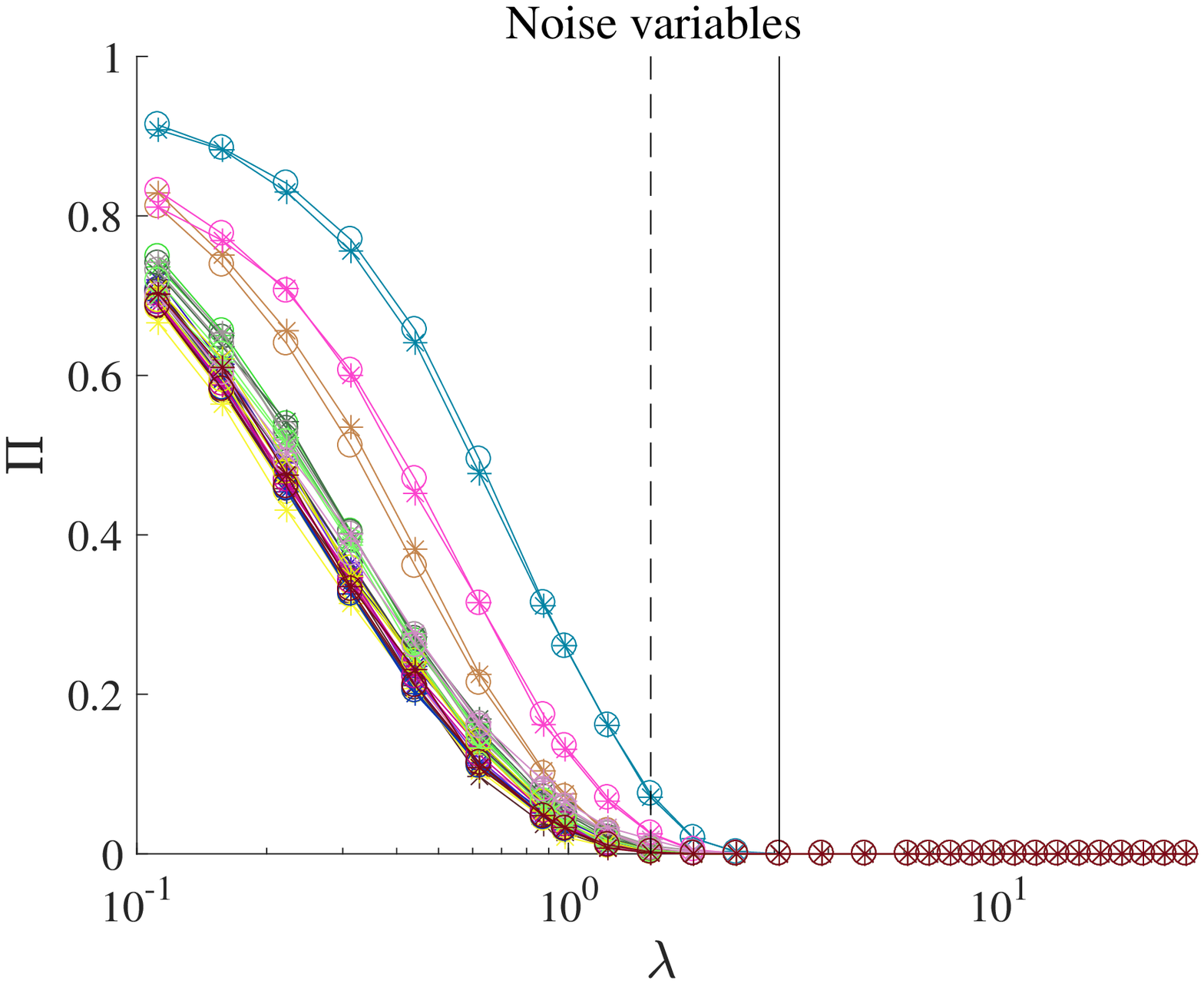}
 \includegraphics[width=0.45\columnwidth]{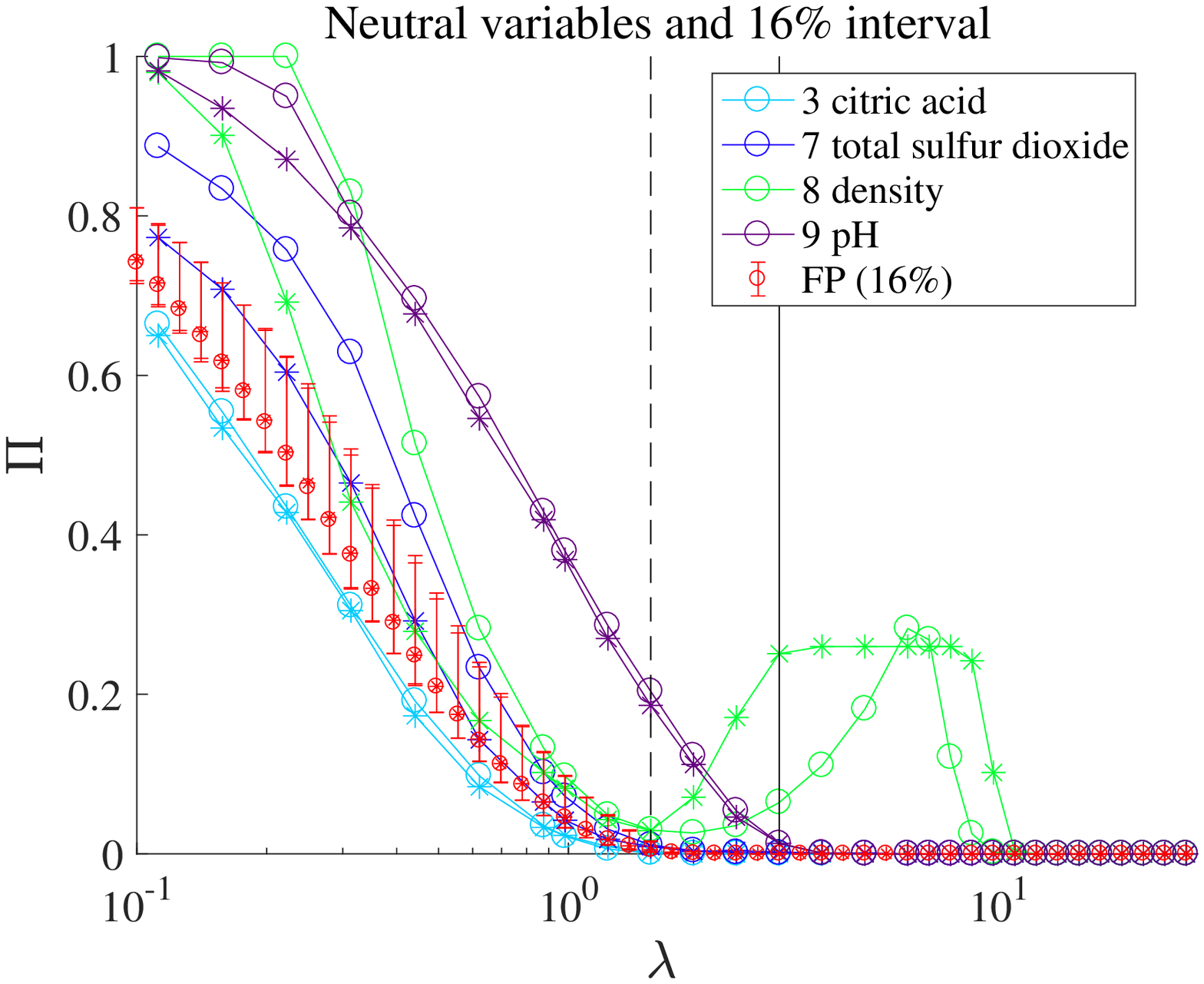}
\caption{Stability paths for the wine quality data with $N_{\rm noise}=689$. The paths of the important variables are in the upper left, those of the neutral variables are in the upper right, and those of the noise variables are in the lower left panels. For the lower left panel, the paths of only a part of the noise variables are shown. The vertical dotted and straight lines denote $\lambda_{\rm min}$ and $\lambda_{\rm opt}$, respectively, as in \Rfig{CV-wine}. Circles denote the result by AMPR and asterisks represent that by the direct numerical resampling, and the same variable is depicted in the same color; the semi-analytic and numerical results show a consistent agreement for all cases. The lower right panel is the simultaneous plot of the stability paths of the neutral variables and the FP interval computed from paths of the noise variables.  
}
\Lfig{SS-wine}
\end{center}
\end{figure}
Both the results of AMPR (circle) and the direct numerical resampling (asterisk) are shown (the same variable is depicted in the same color).

Consequences of \Rfig{SS-wine} are three-fold.
The first one is consistent agreement between the results of AMPR and the direct numerical resampling. For all categories, the two curves of each stability path $\Pi_i(\lambda)$ by AMPR and the direct numerical resampling are very close to each other. This is an additional evidence supporting the accuracy of the proposed semi-analytic method in real-world datasets with correlated covariates. The computational time for obtaining these results in an experiment is $1077$ s by AMPR and $2859$ s by the numerical resampling, demonstrating the efficiency of AMPR. It should be stressed that this efficiency can be enhanced by optimizing the implementation of AMPR. 
The second consequence is the different behaviors of stability paths in different categories. The positive probabilities of the important variables (upper left) are growing largely even for $\lambda>\lambda_{\rm min}$, while those of the noise variables (lower left) do not grow well unless $\lambda$ drops below $\lambda_{\rm min}$. 
The behavior of the neutral variables (upper right) is somewhat elusive, and a better interpretation is provided by utilizing the noise variables, which is the third consequence: As discussed above, we can define a rejection region from the distribution of stability paths of the noise variables; an actual definition here is the $q$-percentiles with $q=16$ and $84$ of the distribution, which is in the lower right panel depicted by red bars with the median (red markers) of the distribution, the legend of which is given as FP, in the same manner as \Rfig{stabsele}. This analysis shows that citric acid and total sulfur dioxide of the neutral variables tend to be in the rejection region, implying that they are irrelevant for modeling wine quality. Moreover, density and pH are well beyond the interval of the rejection region, and they can be regarded as relevant, even though the behavior of density's path is rather tricky. 

The above conclusion of citric acid's irrelevance contradicts that in~\citet{cortez2009modeling}, where citric acid was concluded to be the fourth most important variable. This may be explained by the collinearity of the covariates. In Table \ref{tab:overlap},  the covariates' overlap between citric acid and the others is summarized. 
\begin{table}[]
\centering
\caption{Overlap of the 3rd covariate (citric acid) and the others}
\label{tab:overlap}
\begin{tabular}{|c|c|c|c|c|c|c|c|c|c|c|c|}
\hline
\small index $i$   & 1    & 2     & 3    & 4    & 5    & 6    & 7    & 8    & 9     & 10   & 11    \\ \hline
\small overlap $\bm{x}_i^{\top}\bm{x}_3$ & 0.29 & -0.15 & 1.00 & 0.09 & 0.11 & 0.09 & 0.12 & 0.15 & -0.16 & 0.06 & -0.07 \\ 
\hline
\end{tabular}
\end{table}
This shows a strong collinearity of the 3rd variable with several others, particularly with the 1st variable. This implies that citric acid can be replaced by the 1st variable or fixed acidity. This is a plausible explanation because the proposed model puts considerably more weight on fixed acidity than on citric acid, whereas the opposite is the case in~\citet{cortez2009modeling}. Further thorough comparison will be required to determine which model is better. 

Table \ref{tab:overlap} also implies why AMPR works well for the wine quality dataset: The maximum value of the covariates's overlap is about 0.29 which corresponds $r^{\rm com}\approx 0.5$ in \Rfig{overlap}; \Rfig{r-sweep} shows that the accuracy of AMPR is commonly good around that value of $r^{\rm com}$, explaining the accuracy of AMPR. This consideration indicates that given a new dataset we can judge whether AMPR will give an accurate result or not for the dataset from the covariates' overlap and \Rfigs{overlap}{r-sweep}. 

Overall, the analysis using stability paths provides richer information that cannot be obtained by solely using Lasso. A new objective criterion could be proposed for determining the relevance of variables by utilizing the distribution of stability paths of the added noise variables. These facts highlight the effectiveness of the resampling strategy in variable selection, and the proposed semi-analytic method can implement this strategy in a computationally efficient manner.

\section{Conclusion}\Lsec{Conclusion}
An approximate method was developed for performing resampling in Lasso in a semi-analytic manner. The replica method enables us to analytically take the resampling average over the given data and the average over the penalty coefficient randomness, and the resultant replicated model is approximately handled by a Gaussian approximation using the cavity method. A message passing algorithm named AMPR is thus derived, the computational cost of which is $O(NM)$ per iteration and is sufficiently reasonable. Its convergence in $O(1)$ iterations is guaranteed by a state evolution analysis when covariates are given as zero-mean i.i.d. Gaussian random variables. We demonstrated how it actually works through numerical experiments using simulated and real-world datasets. Comparison with direct numerical resampling has evidenced its approximation accuracy and efficiency in terms of computational cost, even for covariates with correlations of a moderate level. AMPR was also employed to approximately perform Bolasso and SS, and it was applied to the wine quality dataset~\citep{Lichman:13,cortez2009modeling}. To provide a finer quantitative analysis of the dataset, an objective criterion was proposed for determining the relevance of the stability paths by processing the added noise variables, yielding reasonable results in satisfactory computational time. 
 
An advantage of the present framework is its generality. For example, its extension to a generalized linear model is straightforward. This is an immediate future research direction. Extensions to other resampling techniques, such as the multiscale bootstrap method~\citep{shimodaira2004approximately}, would also be interesting. 
 
An unsatisfactory aspect of the present AMPR is that the correlations between covariates are neglected by the approximation. This is a clear drawback, and certain issues arise when the present AMPR is applied to problems involving significantly correlated covariates, although numerical experiments showed that the accuracy of AMPR is still good in the presence of correlations of a moderate level. This drawback may be overcome using more sophisticated approximations, such as the expectation propagation or the adaptive TAP method~\citep{opper2001adaptive,opper2001tractable,opper2005expectation,kabashima2014signal,cakmak2014s,cespedes2014expectation,rangan2016vector,ma2017orthogonal,takeuchi2017rigorous}. Applying those approximations with retaining the benefit of message passing algorithms, the low computational cost, is still a nontrivial challenge and promising future work. 

Resampling is a very versatile framework applicable to various contexts and models in statistics and machine learning. Reducing its computational cost by extending the present method will thus be beneficial in various fields, and can even be imperative, as available data in society will continue to increase rapidly.

\section*{Acknowledgement}
This work was supported by JSPS KAKENHI Nos. 25120013 and 17H00764. TO is also supported by a Grant for Basic Science Research Projects from the Sumitomo Foundation. The authors thank 
Hideitsu Hino for useful comments.


\bibliography{obuchi}

\end{document}